\documentclass{article}

\usepackage{microtype}
\usepackage{graphicx}
\usepackage{subfigure}
\usepackage{booktabs} % for professional tables

\usepackage{times}
\usepackage{graphicx}
\usepackage{multirow}
\usepackage{amsmath}
\usepackage{makecell}
\usepackage{bm}
\usepackage[utf8]{inputenc} % allow utf-8 input
\usepackage[T1]{fontenc}    % use 8-bit T1 fonts
\usepackage{hyperref}       % hyperlinks
\usepackage{url}            % simple URL typesetting
\usepackage{booktabs}       % professional-quality tables
\usepackage{amsfonts}       % blackboard math symbols
\usepackage{nicefrac}       % compact symbols for 1/2, etc.
\usepackage{microtype}      % microtypography
\usepackage{xcolor}         % colors
\usepackage{wrapfig}
\usepackage{enumitem}

\usepackage{hyperref}

\usepackage[accepted]{icml2025}

\usepackage{amsmath}
\usepackage{amssymb}
\usepackage{mathtools}
\usepackage{amsthm}

\usepackage[capitalize,noabbrev]{cleveref}

\theoremstyle{plain}

\theoremstyle{definition}

\theoremstyle{remark}

\usepackage[textsize=tiny]{todonotes}

\begin{document}

\twocolumn[
\icmltitle{Towards a General Time Series Forecasting Model with
Unified Representation and Adaptive Transfer}
% Submission and Formatting Instructions for \\
%            International Conference on Machine Learning (ICML 2025)}



\icmlsetsymbol{equal}{*}

\begin{icmlauthorlist}
\icmlauthor{Yihang Wang}{equal,ecnu}
\icmlauthor{Yuying Qiu}{equal,ecnu}
\icmlauthor{Peng Chen}{ecnu}
\icmlauthor{Kai Zhao}{aalborg}
\icmlauthor{Yang Shu}{ecnu}
\icmlauthor{Zhongwen Rao}{hw}
\icmlauthor{Lujia Pan}{hw}
\icmlauthor{Bin Yang}{ecnu}
\icmlauthor{Chenjuan Guo}{ecnu}
\end{icmlauthorlist}

\icmlaffiliation{ecnu}{East China Normal University, Shanghai, China}
\icmlaffiliation{aalborg}{Aalborg University, Aalborg, Denmark}
\icmlaffiliation{hw}{Huawei Noah's Ark Lab, Shenzhen, China}

\icmlcorrespondingauthor{Zhongwen Rao}{raozhongwen@huawei.com}
\icmlcorrespondingauthor{Chenjuan Guo}{cjguo@dase.ecnu.edu.cn}

\icmlkeywords{Machine Learning, ICML}

\vskip 0.3in
]



\printAffiliationsAndNotice{\icmlEqualContribution} % otherwise use the standard text.

\begin{abstract}
With the growing availability of multi-domain time series data, there is an increasing demand for general forecasting models pre-trained on multi-source datasets to support diverse downstream prediction scenarios. Existing time series foundation models primarily focus on scaling up pre-training datasets and model sizes to enhance generalization performance. 
% However, these approaches often come with significant computational costs. 
In this paper, we take a different approach by addressing two critical aspects of general forecasting models: (1) how to derive unified representations from heterogeneous multi-domain time series data, and (2) how to effectively capture domain-specific features to enable adaptive transfer across various downstream scenarios. To address the first aspect, we propose Decomposed Frequency Learning as the pre-training task, which leverages frequency-based masking and reconstruction to decompose coupled semantic information in time series, resulting in unified representations across domains. For the second aspect, we introduce the Time Series Register, which captures domain-specific representations during pre-training and enhances adaptive transferability to downstream tasks. Our model achieves the state-of-the-art forecasting performance on seven real-world benchmarks, demonstrating remarkable few-shot and zero-shot capabilities.

\end{abstract}

\section{Introduction}
Time series forecasting plays a crucial role in various domains, including energy, smart transportation, weather, and economics~\citep{tfb, wu2025k2vae, qiu2025duet}. However, training deep learning models for each specific dataset is resource-intensive and requires tailored parameter tuning. This approach often suffers from limited prediction accuracy due to data scarcity~\citep{timer, wu2021autocts, wu2024autocts++, wu2024fully}.
A promising solution is to pre-train a general model on diverse time series datasets, which can then be fine-tuned with minimal data for different downstream scenarios or even used directly without fine-tuning. Following this idea, foundation models for time series forecasting have gained significant attention. Recent efforts have focused on scaling up pre-training datasets and model sizes to enhance generalization performance~\citep{moiral,moment,chronos}. However, excessive scaling introduces high computational costs during training and inference, undermining the practicality of general models, particularly in resource-constrained settings.
Beyond scaling, the design of general time series forecasting models can also be approached through pre-training tasks and downstream transfer adaptation. From these two perspectives, we identify the following challenges.

\begin{figure*}[h]    
    \centerline{\includegraphics[width=1\linewidth]{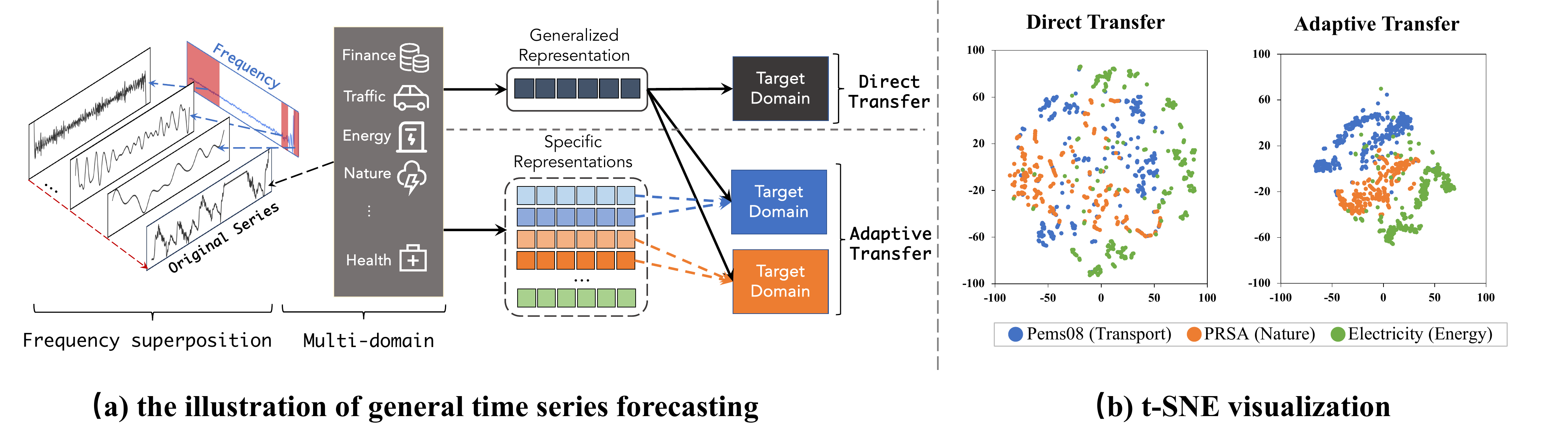}}\vspace{-2mm}
    \caption{\textbf{(a)} Pre-training on multi-domain datasets that exhibit combined frequency. Existing general time series forecasting models only extract generalized representations for direct transfer to various downstream target domains. We propose to learn generalized and specific representations during pre-training, and adaptively transfer them to each target domain. \textbf{(b)} The t-SNE visualization of the hidden representations after direct transfer and adaptive transfer: In direct transfer, representations of different domains are mixed, but in adaptive transfer, they show a clear clustering pattern. The detailed experiment setting is in the Appendix~\ref{setting}}
    \vspace{-5pt}
    \label{fig: intro}
\end{figure*}

\textbf{Obtaining a unified representation from time series data across various domains is challenging}. Time series from each domain involve complex temporal patterns, composed of multiple frequency components combined with each other~\citep{fedformer, wu2024catch}, which is frequency superposition. As shown in Figure~\ref{fig: intro}(a), different frequency components contain distinct semantic information. For example, low and high-frequency components represent long-term trends and rapid variations, respectively~\citep{tfc}. Furthermore, different datasets exhibit diverse frequency distributions, and the significance of low-frequency and high-frequency components for time series modeling varies across domains\citep{not_all}. As a result, large-scale time series data from different domains introduce even more complex temporal patterns and frequency diversity. Existing pre-training frameworks~\citep{simmtm, patchtst, pits}, such as masked modeling and contrastive learning, were proposed to learn a unified representation from time domain. However, these methods overlook the frequency diversity and complexity exhibited in heterogeneous time series that come from various domains,  making it difficult to capture intricate patterns, thus limiting their generalization capabilities.
% Existing pre-training frameworks~\citep{simmtm, patchtst, pits}, such as mask modeling and contrastive learning, were proposed to learn a unified representation from time series data.  However, these methods overlook the frequency diversity and complexity exhibited in heterogeneous time series that come from various domains,  making it difficult to capture intricate patterns, thus limiting their generalization capabilities.

% 且不同的数据集的频率分布是不同的，比如有些趋势强的数据低频数据更有用，而周期强的数据集中高频可能对于理解时序更重要。而大量的多元数据会导致更复杂的时序模式和频率多样性。现有针对时序的预训练框架，如掩码建模与对比学习，仅仅从时域建模，进一步，他们没有考虑多源时序数据的频率多样性，导致其无法识别不同时序数据的关键频域信息，导致其无法捕捉复杂时序模式的关键信息。
% \todo{Existing methods for TSF primarily model time series based on the temporal dependencies among original values in the time domain, which struggle to capture these complex temporal patterns across superimposed frequency components efficiently. }Besides, existing time series pre-training frameworks, such as masked modeling \cite{} and contrastive learning \cite{}, primarily learn representations from the original values which overlooks global information among different frequency components, thus further limiting the model of generalization capabilities.

\textbf{Adaptive transferring information from multi-domain time series to specific downstream scenarios presents a challenge}. Multi-source time series data originate from various domains~\citep{moiral}, whose data exhibit domain-specific information~\citep{timer, DBLP:conf/icde/00010GY0HXJ24, DBLP:journals/pvldb/ZhaoGCHZY23, DBLP:journals/sigmod/GuoJ014}. Information from the same or similar domain as the target domain is useful for improving the model's effectiveness in the target task~\citep{chen2023knowledge}. %However, as shown in Figure~\ref{fig: intro}(a), existing time series pre-training frameworks~\citep{moiral,timer,gpt4ts} focus mainly on learning generalized representation during pre-training and overlook domain-specific representation. Thus, they only transfer the same generalized representation to different target domains, called \textit{direct transfer}, which limits the model's effectiveness in specific downstream tasks. 
However, as shown in Figure~\ref{fig: intro}(a), existing time series pre-training frameworks~\citep{moiral,timer,gpt4ts} primarily focus on learning generalized time series representations during pre-training while overlooking domain-specific representations, called \textit{direct transfer}. While generalized representations are essential, directly transferring them to specific downstream tasks without incorporating domain-specific information leaves room for improvement.
Therefore, it is necessary to learn domain-specific information during pre-training and adaptively transfer the specific representations to target domain, called \textit{adaptive transfer}. Realizing adaptive transfer poses two difficulties:  1) capturing domain-specific information in pre-training. 2) adaptive use of domain-specific information in various downstream tasks.

%左：泛化模型需要从频率叠加的多源数据集进行预训练，提取出泛化特征从而在下游任务进行直接迁移，或者结合特殊特征进行自适应迁移。右：下游三个domain的数据集在经过直接迁移和自适应迁移后特征的tsne可视化
% \begin{figure*}[h]    
%     \centerline{\includegraphics[width=1\linewidth]{Figures/Intro.pdf}}\vspace{-2mm}
%     \caption{\textbf{(a)} Pre-training on multi-domain datasets that exhibit combined frequency. Existing general time series forecasting models only extract generalized representations for direct transfer to various downstream target domains. We propose to learn generalized and specific representations during pre-training, and adaptively transfer them to each target domain. \textbf{(b)} The t-SNE visualization of the hidden representations after direct transfer and adaptive transfer: In direct transfer, representations of different domains are mixed, but in adaptive transfer, they show a clear clustering pattern. \color{blue}{The detailed experiment setting is in the Appendix~\ref{Appendix_intro_setting}}}
%     \vspace{-5pt}
%     \label{fig: intro}
% \end{figure*}

To address these challenges, we propose a register assisted general time series forecasting model with decomposed frequency learning, namely \textbf{ROSE}.
% First, we propose Decomposed Frequency Learning to learn generalized representations. By decomposing frequency components with the Fourier Transform and using decoupled time series values in the time domain for encoding and reconstruction, our model can learn generalized features effectively across different frequency components. It is worth noting that the decoupled time series values contain important semantic features such as trends and periods, which are presented in each frequency component in the frequency domain. Therefore, decomposed frequency learning helps the model learn these valuable semantic features. Meanwhile, we can learn across-period and global information among different frequency components by masking multiple frequency components, further improving the model's generalization capabilities.
%这一段感觉针对我们Decomposed frequency Learning的具体方法介绍相对较少，更多的是Frequency learning的意义
\textbf{First}, we propose Decomposed Frequency Learning that learns generalized representations to solve the issue with coupled semantic information. We decompose individual time series using the Fourier transform with a novel frequency-based masking method, and then convert it back to the time domain to obtain decoupled time series for reconstruction. It makes complex temporal patterns disentangled, thus benefiting the model to learn generalized representations.
%To address these challenges, we propose Prompt-VQ and Frequency-based Learning to construct a general Time series forecasting model (\textbf{ROSE}). Firstly, to address the frequency superposition and diversity in time series datasets, we introduce the innovative task of multi-frequency masking modeling. By applying multiple frequency masks to time series data in the frequency domain and then transforming it back to the time domain for encoding and reconstruction, our model can learn unified generalized features across multiple frequencies. It is worth noting that time series data contains important semantic features such as trends and cycles, which can be represented through the information of various frequencies in the frequency domain. Therefore, masking time series data in the frequency domain helps the model learn these valuable global semantic features. Furthermore, through multiple frequency masking, we can enhance the model's ability to learn across different periods and time scales, thereby improving its generalization capability. 
\textbf{Second}, we introduce Time Series Register (TS-Register) to learn domain-specific information in multi-domain data. By setting up a register, we generate register tokens to learn each domain-specific information during pre-training. In a downstream scenario, the model adaptively selects Top-K vectors from the register that are close to the target domain of interest. 
% During fine-tuning, we incorporate learnable vectors into the selected register tokens to complement target specific information to perform more flexible adaptive transfer. 
During fine-tuning, we adjust the selected register tokens with a novel learnable low-rank matrix, which complements target-specific information to perform more flexible adaptive transfer.
As shown in Figure~\ref{fig: intro}(b), adaptive transfer successfully utilizes domain-specific information in multi-domain time series, which contributes to the model's performance in target tasks. The contributions are summarized as follows:

% \begin{itemize}[leftmargin=3.5mm]
\begin{itemize}
    \item We propose ROSE, a novel light weight general time series forecasting model using multi-domain datasets for pre-training and improving downstream fine-tuning performance and efficiency.
    \item We propose a novel Decomposed Frequency Learning that employs multi-frequency masking to learn complex general temporal patterns from multi-domain data, empowering the model's generalization capability. 
    \item We propose a novel TS-Register to capture domain-specific information in pre-training and enable adaptive transfer of target-oriented specific information for downstream tasks.
    \item Our experiments with 7 real-world benchmarks demonstrate that ROSE achieves state-of-the-art performance in full-shot setting and achieves competitive or superior results in few-shot setting, along with impressive transferability in zero-shot setting.
\end{itemize}

\begin{figure*}[t]
    \centering\vspace{0mm}
    \includegraphics[width=1\linewidth]{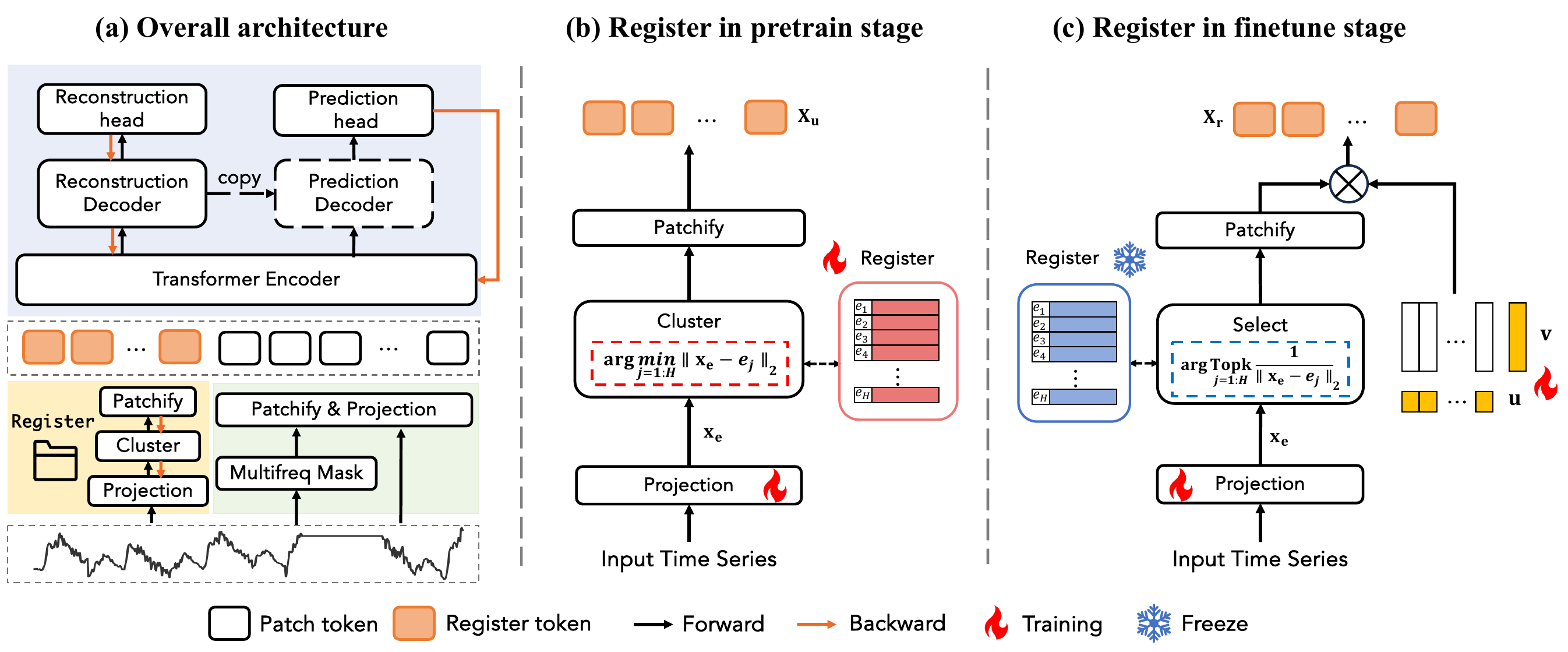}\vspace{-1em}
    \caption{The model architecture of ROSE.}
    \vspace{-10pt}
    \label{fig:Frame_1}
\end{figure*}
\section{Related Work}
\subsection{Traditional Time Series Forecasting}
The statistical time series forecasting models like ARIMA~\citep{ARIMA}, despite their theoretical support, are limited in modeling nonlinearity. With the rise of deep learning, many RNN-based models~\citep{MileTS,wen2017multi,deepar} have been proposed, modeling the sequential data with an autoregressive process. CNN-based models~\citep{moderntcn, scinet} have also received widespread attention due to their ability to capture local features. MICN~\citep{micn} utilizes TCN to capture both local and global features, while TimesNet~\citep{timesnet} focuses on modeling 2D temporal variations. However, both RNNs and CNNs struggle to capture long-term dependencies. Transformer-based models~\citep{fedformer, patchtst, autoformer,itransformer, chen2024pathformer}, with their attention mechanism, can capture long dependencies and extract global information, leading to widespread applications in long-time series prediction. However, this case-by-case paradigm requires meticulous hyperparameter design for different datasets, and its predictive performance can also be affected by data scarcity.
\subsection{Time Series Forecasting Foundation Model}
Pre-training with multiple sources time series has recently received widespread attention~\citep{lag, forecastpfn, timegpt,lptm}. MOMENT~\citep{moment} and MOIRAI~\citep{moiral} adopt a BERT-style pre-training approach, while Timer~\citep{timer}, Chronos~\citep{chronos} and TimsFM~\citep{predct} use a GPT-style pre-training approach, giving rise to improved performance in time series prediction. However, the above methods overlook domain-specific information from multi-source data, thus limiting the performance of the models. Different from previous approaches, ROSE pre-trains on large-scale data from various domains and it considers both generalized representations and domain-specific information, which facilitates flexible adaptive transfer in downstream tasks. 
\section{Methodology}

% \textbf{Problem Definition.} Given a multivariate time series $\mathbf{X}_t=\{\mathbf{x}_{t-L:t}^i\}_{i=1}^C $, where each $\mathbf{x}_{t-L:t}^i \in \mathbb{R}^L$ is a sequence of observations. $L$ denotes the look-back window and $C$ denotes the number of channels. The forecasting task is to predict the future values $\mathbf{\hat{Y}}_t=\{\mathbf{\hat{x}}_{t:t+F}^i\}_{i=1}^C$, where $F$ denotes the forecast horizon. $\mathbf{Y}_t=\{\mathbf{x}_{t:t+F}^i\}_{i=1}^C$ is the ground truth of future.

% The general time series forecasting model is pre-trained with multi-source datasets $\mathbf{D}_\text{pre-train}= \{(\mathbf{X}_t^{j}, \mathbf{Y}_t^{j})\}^{N}_{j=1}$, where $N$ is the number of datasets. For the downstream task, the model is fine-tuned with a training dataset $\mathbf{D}_\text{train}=\{(\mathbf{X}_t^\text{train}, \mathbf{Y}_t^\text{train})\}$, and is tested with $\mathbf{D}_\text{test}=\{(\mathbf{X}_t^\text{test}, \mathbf{Y}_t^\text{test})\}$ to predict $\mathbf{\hat{Y}}_t^\text{test}$, where $\mathbf{D}_\text{pre-train}$, $\mathbf{D}_\text{train}$ and $\mathbf{D}_\text{test}$ are pairwise disjoint. Alternatively, the model could be directly tested using $\mathbf{D}_\text{test}$ without fine-tuning with $\mathbf{D}_\text{train}$ to predict $\mathbf{\hat{Y}}_t^\text{test}$.

\subsection{Architecture}
As illustrated in Figure~\ref{fig:Frame_1}, ROSE adopts an encoder-decoder architecture for time series modeling. Its backbone comprises multiple Transformer layers, which effectively process sequential information and capture temporal dependencies~\citep{transformer}. Both the reconstruction decoder and prediction decoder share the same structure as the Transformer encoder and are designed for reconstruction and prediction tasks, respectively. The reconstruction task enables the model to gain a comprehensive understanding of time series, while the prediction task enhances its few-shot and zero-shot capabilities. ROSE is pre-trained in a channel-independent way, which is widely used in time series forecasting~\citep{patchtst}.

\textbf{Input representations.} To enhance the generalization of ROSE for adaptive transferring from multi-domains to different target domains, we model the inputs $\mathbf{x}$ with \textbf{patch tokens} and \textbf{register tokens}. 
% Patch token is obtained using a linear layer that maps the patches cut from the input sequence to a fixed dimension, which is used to capture local information.  
Patch tokens are obtained by partitioning the time series using patching layers~\citep{patchtst}, to preserve local temporal information. 
% Register tokens are obtained by linear mapping and clustering each of the entire time series input into discrete embedding, to capture domain-specific information, which will be introduced in Section~\ref{ts-register}. 
% \qyy{Register tokens are obtained by linearly mapping the entire time series and then retrieving it from the register, to capture domain-specific information, which will be introduced in Section~\ref{ts-register}. }
Register tokens that 
%retrieved from the register by the embedding of the entire time series, 
capture domain-specific information will be introduced in Section~\ref{ts-register}.

\begin{figure*}[h]    
    \centerline{\includegraphics[width=1\linewidth]{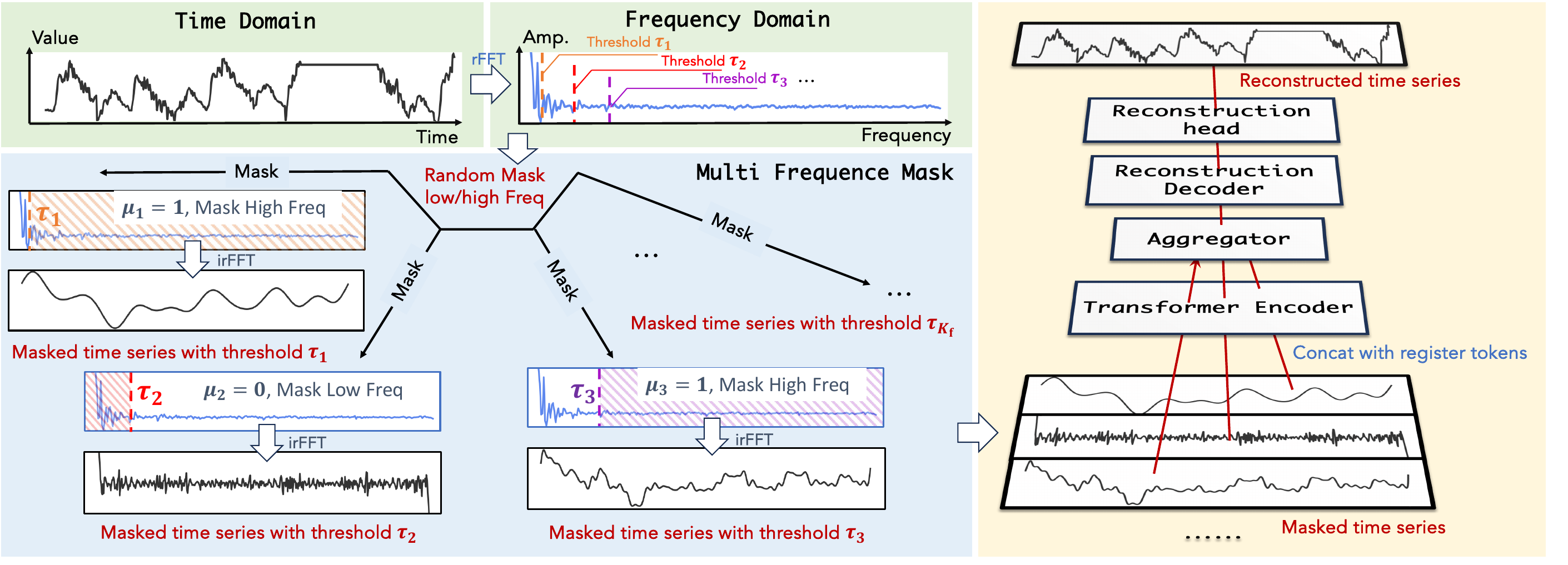}}\vspace{-10pt}
    \caption{An illustration of decomposed frequency learning. Based on the sampled thresholds, we randomly apply low/high-frequency masking to the time series in the frequency domain and then transform it back to the time domain for reconstruction.}
    \vspace{-10pt}
    \label{fig:mask}
\end{figure*}
\subsection{Decomposed Frequency Learning } \label{multi_freq_mask}
As shown in Figure \ref{fig: intro}, time series data are composed of multiple superimposed frequency components, resulting in the overlap of different temporal changes. Furthermore, low-frequency components typically contain information about overall trends and longer-scale variations, and high-frequency components usually contain information about short-term fluctuations and shorter-scale variations, therefore, understanding time series from low and high frequencies separately benefits general time series representation learning.
Based on the observations above, we propose a novel frequency-based masked modeling that randomly mask either high-frequency or low-frequency components of a time series multiple times as the key to enable learning of common time series patterns, such as trends and various long and short term fluctuations.  Finally, reconstruction task assists the model in comprehending the data from various frequency perspectives, enabling it to learn generalized representations. 
In contrast, existing frequency masking methods~\citep{tfc}, which randomly mask frequencies of a single time series once, show limited forecasting effectiveness due to the lack of common pattern learning from heterogeneous time series that come from various domains.

% As shown in Figure~\ref{fig: intro}, time series data are composed of multiple combined frequency components, resulting in overlapping of different temporal variations. Furthermore, low-frequency components typically contain information about overall trends and longer-scale variations, and high-frequency components usually contain information about short-term fluctuations and shorter-scale variations, therefore, understanding time series from low and high frequencies separately benefits general time series representation learning. 
% Based on the observations above, we propose decomposed frequency learning based on multi-freq masking to understand the time series from multiple-frequency perspectives, which enhances the model's ability to learn generalized representations.
% A univariate illustration of decomposed frequency learning. 根据预先定义的阈值，通过随机对时序频域进行高频掩码或低频掩码，并将其转换至时域，将原先耦合的时序信息，分解为包含不同频率分量的时序，从而可以使模型从多个频率角度理解时序
% \begin{figure*}[h]    
%     \centerline{\includegraphics[width=1\linewidth]{Figures/MFM_5.pdf}}\vspace{-10pt}
%     \caption{An illustration of decomposed frequency learning. Based on the sampled thresholds, we randomly apply low/high-frequency masking to the time series in the frequency domain and then transform it back to the time domain for reconstruction.}
%     \vspace{-10pt}
%     \label{fig:mask}
% \end{figure*}

\textbf{Multi-frequency masking.} As shown in the green part of Figure~\ref{fig:mask}, given a time series $\mathbf{x} \in \mathbb{R}^L$, we utilize the Real Fast Fourier Transform (rFFT)~\citep{rfft} to transform it into the frequency domain, giving rise to 
% $\mathbf{x}_{\mathrm{freq}}$ which is a vector containing L/2+1 complex numbers.
$\mathbf{x}_{\mathrm{freq}} \in \mathbb{C}^{L/2+1}$.
\begin{equation}
\mathbf{x}_{\mathrm{freq}} = \mathrm{rFFT}(\mathbf{x}).
\end{equation}
%% 为了将时序的高频信息和低频信息分解出来分别建模，我们set $K$ thresholds $\{ \tau_1,\tau_2,\tau_3...,\tau_K\}$ and $K$ random numbers $\{\mu_1,\mu_2,\mu_3,...,\mu_K\}$ to random masking, generating a mask matrix $M \in R^{t \times l/2}$, where each row corresponds to a mask vector $M(i)$ and each element is 0 or 1.
%
To separately model high-frequency and low-frequency information in time series, we sample $K_\mathrm{f}$ thresholds ${\tau_1,\tau_2,\tau_3,...,\tau_{K_\mathrm{f}}}$ and $K_\mathrm{f}$ random numbers ${\mu_1,\mu_2,\mu_3,...,\mu_{K_\mathrm{f}}}$ for multi-frequency masks, where $\tau \in \mathrm{Uniform}(0,a)$, $a < L/2+1 $ , and $\mu \in \mathrm{Bernoulli}(p)$. Each pair of $\tau_i$  and $\mu_i$  corresponds to the $i_{th}$  frequency mask.  This generates a mask matrix $\mathbf{M} \in \{0,1\}^{K_\mathrm{f} \times (L/2+1)}$, where each row corresponds to the $i_{th}$  frequency mask,  each column corresponds to  the $j_{th}$ frequency, and each element $m_{ij}$ is 0 or 1, meaning that the $j_{th}$ frequency  is masked with the  $i_{th}$  frequency mask
 or not.
\begin{equation}
m_{ij}=\left\{
\begin{aligned}
&\quad \mu_i&,if \,\ j <\tau_i \\
&(1-\mu_i)&,if \,\  j>\tau_i
\end{aligned}\right.
,
\end{equation}
% where $i$,$f$ represents the $f_{th}$ value of the $i_{th}$ mask vector. $\tau_i$ denotes the threshold for the $i_{th}$ frequency domain mask, and $\mu_i \in \mathrm{Bernoulli}(1,p)$ indicates the random number for the $i_{th}$ frequency domain mask.
where $\tau_i$ and $\mu_i$ denote the threshold and random number for the $i_{th}$ frequency domain mask. If $\mu_i = 1$, it means that frequency components above $\tau_i$ will be masked, indicating to mask high frequency , as shown by the threshold $\tau_1$ in Figure~\ref{fig:mask}. Conversely, if $\mu_i = 0$, it signifies that frequency components below $\tau_i$ will be masked, indicating to mask low frequency, exemplified by threshold $\tau_2$ in Figure~\ref{fig:mask}.

% If $\mu_i=1$, it indicates masking of the high frequencies just like threshold 1 in Figure~\ref{fig:mask}, whereas if
%  $\mu_i=0$, it signifies the masking of the low frequencies like threshold 2.
 
After obtaining the mask matrix $\mathbf{M}$, we replicate $\mathbf{x}_\mathrm{freq}$ $K_\mathrm{f}$ times to get the $\mathbf{X}_{\mathrm{freq}} \in \mathbb{C}^{K_\mathrm{f} \times 
 L/2+1}$ and perform element-wise Hadamard product with the mask matrix $\mathbf{M}$ to get masked frequency of time series. Then, we use the inverse Real Fast Fourier Transform (irFFT) to convert the results from the frequency domain back to the time domain and get $K_\mathrm{f}
$ masked sequences $\mathbf{X}_{\mathrm{mask}} = \{\mathbf{x}_{\mathrm{mask}}^i\}_{i=1}^{K_\mathrm{f}} $,  where each $\mathbf{x}_\mathrm{mask}^i\in \mathbb{R}^{L}$ corresponding to masking with a different threshold $\tau_i$.
\begin{equation}
    \mathbf{X}_{\mathrm{mask}}= \mathrm{irFFT}(\mathbf{X}_\mathrm{freq} \odot \mathbf{M}).
\end{equation}
%在得到掩码序列之后，我们将每条序列切分为P个不重合的patch。然后我们使用一层先行层将每个patch映射到固定维度，从而将每条掩码序列转为P个patch tokens。如3.2节所示，我们同时将每条掩码序列输入进TS-Register从而将每条序列转为N个Register tokens。然后我们每条序列得到的patch token与register token 拼接，并输入进transformer encoder得到模型表征。
%We simultaneously input each masked sequence into the TS-Register, whose details are described in Section~\ref{ts-register}, converting each sequence $\mathbf{X}_\mathrm{mask}^i$ into $N_\mathrm{r}$ Register tokens, and we get $\mathcal{X}_\mathrm{mu}=\{\mathbf{X}_{\mathrm{mu}}^i\}_{i=1}^{K_\mathrm{f}} $, each $\mathbf{X}_\mathrm{mu}^i\in \mathbb{R}^{N_\mathrm{r} \times D}$ to capture domain-specific information
%我们复制x_u k times ,其中x_u是由原始序列输入ts-register得到,并会在3.3节详述。
\textbf{Representation learning.} As shown in the yellow part of Figure~\ref{fig:mask}, after obtaining the $K_\mathrm{f}$ masked sequences $\mathbf{X}_{\mathrm{mask}}$, we divide each sequence  $\mathbf{x}_\mathrm{mask}^i$  into $P$ non-overlapping patches, and use a linear layer to transforming them into $P$ patch tokens, and thus we get $\mathcal{X}_\mathrm{mp}=\{\mathbf{X}_{\mathrm{mp}}^i\}_{i=1}^{K_\mathrm{f}}$ to capture general information, where each $\mathbf{X}_\mathrm{mp}^i\in \mathbb{R}^{P \times D}$, and $D$ is the dimension for each patch token. We replicate the register tokens  $\mathbf{X}_{\mathrm{u}}$ $K_\mathrm{f}$ times to get $\mathcal{X}_\mathrm{u} \in \mathbb{R}^{K_{\mathrm{f}} \times N_\mathrm{r} \times D}$, where  $\mathbf{X}_{\mathrm{u}}$ is obtained by inputting the original sequence into the TS-Register, as detailed in Section~\ref{ts-register}. Then, we concatenate the patch tokens $\mathcal{X}_\mathrm{mp}$ with the register tokens $\mathcal{X}_\mathrm{u}$, and feed them into the Transformer encoder to obtain the representation of each masked series. These representations are then averaging aggregated to yield a unified representation $\mathbf{S}_\mathrm{m} \in \mathbb{R}^{(N_\mathrm{r}+P) \times D}$. 
% The aggregator is the averaging operation.
 \begin{equation}
\mathbf{S}_\mathrm{m}=\mathrm{Aggregator(Encoder(Concat(\mathcal{X}_{mp}, \mathcal{X}_{u})))}.
 \end{equation}
 
 \textbf{Reconstruction task.} After obtaining the representation $\mathbf{S}_\mathrm{m}$, we feed it into the reconstruction decoder, which shares same stucture as the Tranformer encoder, and ultimately reconstruct the original sequence $\mathbf{\hat{x}} \in \mathbb{R}^{L}$ through the reconstruction head, which is a linear layer. As frequency domain masking affects the overall time series, we compute the Mean Squared Error (MSE) reconstruction loss for the entire time series. 
  \begin{equation}
    \mathcal{L}_\mathrm{reconstruction}=\mathrm{||\mathbf{x}-\hat{\mathbf{x}}||_2^2}.
 \end{equation}

\subsection{Time Series Register} \label{ts-register}

By decomposed frequency learning, we can obtain the general representations. Additionally, we propose the TS-Register that learns domain-specific information from the multi-domain datasets for adaptive transfer. It clusters domain-specific information from the multi-domain datasets into register tokens and stores such domain-specific information in the register during pre-training. Then, it adaptively selects domain-specific information from the register via a Top-K selection strategy to enhance the performance in the target domain. A novel learnable low-rank matrix is proposed to set to complement the downstream dataset-specific information through fine-tuning.

%, shared by input data 
We set up a randomly initialized register $\mathbf{E} \in \mathbb{R}^{H\times D_\mathrm{r}}$ with $H$ cluster center vectors $\mathbf{e}_i \in \mathbb{R}^{D_\mathrm{r}}, i \in \{1, 2, …, H\}$. Each of input time series 
$\mathbf{x} \in \mathbb{R}^{L}$
is projected into a data-dependent embedding 
$\mathbf{x}_\mathrm{e} \in \mathbb{R}^{D_\mathrm{r}}$ 
through a linear layer.

\textbf{Pre-training stage.} As shown in Figure~\ref{fig:Frame_1}(b), we use the register to cluster these data-dependent embeddings, which generate domain-specific information, and store them in pre-training.
% Specifically, during the pre-training process, each of the entire input sequence 
% $\mathbf{x} \in \mathbb{R}^{l}$
% is projected into a data-dependent embedding 
% $\mathbf{z}_\mathrm{e} \in \mathbb{R}^{D_\mathrm{r}}$ 
% through a linear layer. We set up a randomly initialized register codebook $\mathbf{E} \in \mathbb{R}^{H\times D_\mathrm{r}}$ with $H$ cluster center vectors $\mathbf{e}_i \in \mathbb{R}^{D_\mathrm{r}}, i \in \{1, 2, …, H\}$, shared by all input data. 
Specifically, We find a cluster center vector $\mathbf{e}_\delta$ from the register $\mathbf{E}$ where we use $\delta$ to denote the cluster that the data-dependent embedding $\mathbf{x}_\mathrm{e}$ belongs to.
%$\mathbf{e}_\delta$ , whose process is shown below
%\textbf{Pre-training stage}. As shown in Figure \ref{fig:Frame_1} (b), to obtain domain-specific information during pre-training, we utilize a linear layer to map the entire input series into data-dependent vectors with fixed lengths. %Since real-world temporal domains are quantized and 
%Since quantizing the vector allows us to cluster and retain only important and global features \cite{vqvit}, we quantize these vectors using a codebook as a register, which enables us to store representative domain-specific information. The quantized vectors are invariant under small perturbations, which promotes better representation of domain-specific information and robustness of the vectors in the register and avoids their over-reliance on detailed information about specific datasets.
% \begin{equation} %\mathbf{e}_\delta\ ,
%  \delta=\underset{j=1:H}{\arg \operatorname{min} }\left\|\mathbf{x}_\mathrm{e}-\mathbf{e}_j\right\|_2.
% \label{argmin}
% \end{equation}
% \vspace{-8pt}
\begin{equation}
\mathcal{L}_{\text{register}}=\|\mathbf{x}_\mathrm{e}-\mathbf{e}_\delta\|^2_2,\ \  \delta=\underset{j=1:H}{\arg \operatorname{min} }\left\|\mathbf{x}_\mathrm{e}-\mathbf{e}_j\right\|_2.
\label{loss_register}
\end{equation}
To update the cluster center vectors in the register $\mathbf{E}$ that represents the domain information of the pre-trained datasets, we set the loss function shown in Equation~\ref{loss_register} that minimizes the distance between the embedding $\mathbf{x}_\mathrm{e}$ and the cluster center $\mathbf{e}_\delta$. 
To solve the problem that the gradient of the $\arg\operatorname{min}$ function cannot be backpropagated, we use the stop gradient operation to pass the gradient of $\mathbf{e}_\delta$ directly to $\mathbf{x}_\mathrm{e}$.

% $\text{sg}()$ refers to the stop gradient operation, and is used to solve the problem that the gradient of the $\arg\operatorname{min}$ function cannot be backpropagated~\citep{gumbel}. The first term is used to update the register $\mathbf{E}$, and the second term is used to update the parameters of the linear layer that learns $\mathbf{x}_\mathrm{e}$.

In this way, the vectors in the register $\mathbf{E}$ cluster the embeddings of different data and learn the domain-specific centers for pre-trained datasets, which can represent domain-specific information. As a vector in the register $\mathbf{E}$, $\mathbf{e}_\delta$ represents the domain-specific information for input $\mathbf{x}$. $\mathbf{e}_\delta$ is invariant under small perturbations in $\mathbf{x}_\mathrm{e}$ that represents $\mathbf{x}$, which promotes better representation of domain-specific information and robustness of the vectors in the register. This also avoids their over-reliance on detailed information about specific datasets.

The cluster center vector $\mathbf{e}_\delta$ is then patched into $\mathbf{X}_\mathrm{u} \in \mathbb{R}^{N_\text{r}\times D}$, where $N_\text{r}$ is the number of the register tokens and $D$ is the dimensionality of Transformer latent space. $\mathbf{X}_\mathrm{u}$ is called register tokens, which are used as the prefix of the patch tokens $\mathbf{X}_\mathrm{p} \in \mathbb{R}^{P \times D}$ and input for the Transformer encoder to provide domain-specific information.

\textbf{Fine-tuning stage}. As shown in Figure~\ref{fig:Frame_1}(c), after obtaining a register $\mathbf{E}$ that contains domain-specific information through pre-training, we freeze the register parameters to adaptively use this domain-specific information in the downstream tasks. 

Since the target domain may not strictly fit one of the upstream domains, we propose a novel embedding learning of the downstream data by employing a Top-K strategy that selects $k$ similar vectors in the register. As shown in Equation~\ref{topk}, the embedding of input time series $\mathbf{x}_\mathrm{e}$ picks the $k$ nearest vectors in the register $\mathbf{E}$, and uses their average as $\bar{\mathbf{e}}_k$ to represent the domain-specific information from the pre-train stage. $\bar{\mathbf{e}}_k$ is also patched into $\mathbf{X}_\mathrm{d} \in \mathbb{R}^{N_\text{r}\times D}$ and is used as the \textbf{domain specific register tokens}.
\begin{equation}
\bar{\mathbf{e}}_k=\frac{1}{k}\sum_{i=1}^{k} \mathbf{e}_{\delta_i}, \ \ \left\{\delta_1, \cdots, \delta_k\right\}=\underset{j=1:H}{\arg \operatorname{Topk}}(\frac{1}{\left\|\mathbf{x}_\mathrm{e}-\mathbf{e}_j\right\|_2}).
\label{topk}
\end{equation}
Since the downstream data has its own specific information at the dataset level in addition to the domain level, this may not be fully represented by the domain information obtained from the pre-trained dataset alone. Therefore, we innovatively set a learnable matrix $\mathbf{A} \in \mathbb{R}^{N_\mathrm{r} \times D}$ to adjust $\mathbf{X}_\mathrm{d}$ to complement the \textbf{specific information of downstream data}. Since the pre-trained model has a very low intrinsic dimension~\citep{intrinsic}, in order to get better fine-tuning results, $\mathbf{A}$ is set as a low-rank matrix:
\begin{equation}
\mathbf{A}=\mathbf{u} \times \mathbf{v^\mathrm{T}},
\label{delta_z}
\end{equation}
where $\mathbf{u} \in \mathbb{R}^{N_\mathrm{r}}$ and $\mathbf{v} \in \mathbb{R}^{D}$, and only the vectors $\mathbf{u}$ and $\mathbf{v}$ need to be retrained in the fine-tuning step.
As illustrated in Equation~\ref{Hadamard product}, the register token $\mathbf{X}_\mathrm{r}$ of the downstream scenario is obtained by doing the Hadamard product of $\mathbf{X}_\mathrm{d}$, which represents the domain-specific information obtained at the pre-train stage, and $\mathbf{A}$, which represents the downstream dataset-specific information.
% , where $\circ$ denotes the Hadamard product.
\begin{equation} \label{Hadamard product}
\mathbf{X}_\mathrm{r}=\mathbf{X}_\mathrm{d}\odot \mathbf{A}.
\end{equation}

\subsection{Training}

To improve the prediction performance in zero-shot and few-shot settings, we co-train supervised prediction with self-supervised reconstruction that uses multi-frequency masking to learn unified features that are more applicable to the downstream prediction task. 
% We normalize time series by employing the REVIN~\citep{revin} that is commonly used by the state-of-the-art time series models, which first normalizes each input sample and subsequently applies inverse normalization to recover the model's output.

\textbf{Prediction task.}  The input time series $\mathbf{x} \in \mathbb{R}^{L}$ is sliced into $P$ non-overlapping patches and then mapped to $\mathbf{X}_\mathrm{p} \in \mathbb{R}^{P \times D}$.
% It should be noted that the input sequence in the prediction task is not masked. 
Based on common forecasting needs~\citep{tfb}, we set up four prediction heads mapping to prediction lengths of $\{96, 192, 336, 720\}$ to accomplish the prediction task. Patch tokens $\mathbf{X}_\mathrm{p}$ are concatenated with the register tokens $\mathbf{X}_\mathrm{u}$  and then successively fed into the Transformer encoder to yield the representation $\mathbf{S} \in \mathbb{R}^{(N_\mathrm{r}+P) \times D}$:
 \begin{equation}
     \mathbf{S}=\mathrm{Encoder(Concatenate(\mathcal{X}_{p}, \mathcal{X}_{u}))}.
 \end{equation}
We feed the representation $\mathbf{S}$ into the prediction decoder and prediction heads to obtain four prediction results $\mathbf{\hat{Y}}_F$, where $F\in \{96,192,336,720\}$. With the ground truth $\mathbf{Y}_F$, the prediction loss $\mathcal{L}_{\text{prediction}}$ is shown in Equation~\ref{pred_loss}.
 \begin{equation}
     \mathcal{L}_{\text{prediction}}=\sum_{F \in \{96,192,336,720\}}||{\mathbf{Y}_F-{\mathbf{\hat{Y}}_F}}||_2^2.
\label{pred_loss}
 \end{equation}
\textbf{Pre-training.} The reconstruction task learns generalized features through the Transformer encoder and reconstruction decoder. To utilize these features for the prediction task, the parameters of the reconstruction decoder are copied to the prediction decoder during forward propagation. To avoid prediction training affecting the generalization performance of the model, the gradients of the prediction heads are skipped at back-propagation. The overall loss of ROSE in pre-training stage is shown in Equation~\ref{total_loss}.
 \begin{equation} \label{total_loss}
\mathcal{L_\text{pre-train}}=\mathcal{L}_{\text{reconstruction}}+\mathcal{L}_{\text{prediction}}+\mathcal{L}_{\text{register}}.
 \end{equation}
 %Since both prediction and reconstruction loss are MSEs between sequences and have the same magnitude, their coefficients are set to 1.
\textbf{Fine-tuning.} We only perform a prediction task in fine-tuning. Patch tokens $\mathbf{X}_\text{p}$ are concatenated with the adjusted register tokens $\mathbf{X}_\mathrm{r}$. For a downstream task with a fixed prediction length, we use the corresponding pre-trained prediction head to fine-tune the model.

%  \begin{equation} \label{total_loss_finetune}
% \mathcal{L_\mathrm{fine-tune}}=||{\mathbf{Y}_F-{\mathbf{\hat{Y}}_F}}||_2^2.
%  \end{equation}

\section{Experiments}
\label{experiments}
\textbf{Pre-training datasets.} The datasets are crucial for pre-training a general time series forecasting model. In light of this, we gather many publicly available datasets from various domains, including energy, nature, health, transport, web, economics, etc.  The details of these datasets are shown in the Appendix~\ref{pretraining_datasets}. To enhance data utilization, we downsample fine-grained datasets to coarser granularity, resulting in approximately 887 million time points.

\textbf{Evaluation datasets.} To conduct comprehensive and fair comparisons for different models, we conduct experiments on seven well-known forecasting benchmarks as the target datasets, including Weather, Traffic, Electricity, and ETT (4 subsets), which cover multiple domains.

% \textbf{Baselines.} We select the state-of-the-art models as our baselines, including Transformer-based models: Timer~\citep{timer}, iTransformer~\citep{itransformer}, and PatchTST~\citep{patchtst}; CNN-based model: TimesNet~\citep{timesnet}; MLP-based models: FITS~\citep{fits}, TIDE~\citep{tide}, and DLinear~\citep{dlinear}. It is worth noting that Timer is a recent general time series model with a GPT-style architecture. Since the size of the look-back window can affect the performance of different models, we choose the look-back window size in $\{96, 336, 512, 720\}$ for each baseline that achieves its best results for fair comparisons.

\textbf{Baselines.} We select the state-of-the-art models as baselines in full-shot and few-shot setting, including four specific models: iTransformer~\citep{itransformer}, PatchTST~\citep{patchtst}, TimesNet~\citep{timesnet}, and DLinear~\citep{dlinear}, and two LLM-based models: GPT4TS~\citep{gpt4ts} and $\textbf{S}^2$IP-LLM~\citep{s2ipllm}. In addition, we select five foundation models for comparison in zero-shot setting, including Timer~\citep{timer}, MOIRAI~\citep{moiral}, Chronos~\citep{chronos}, TimesFM~\citep{timefm}, and Moment~\citep{moment}.

\textbf{Setup.} Consistent with previous works, we adopted Mean Squared Error (MSE) and Mean Absolute Error (MAE) as evaluation metrics. Due to ROSE mostly aims at long-term predictions, for fair comparison, all methods fix the look-back window $L$ = 512 and predict the future values with lengths $F=\{96, 192, 336, 720\}$. More implementation details are presented in the Appendix~\ref{setting}.

\subsection{In Distribution Forecasting}

\textbf{Setting.} In full-shot setting, we utilize full downstream data to fine-tune pre-trained ROSE and baselines.
% with full downstream data. 
In few-shot setting, we fine-tune all models with only 10\% train data. 
The \textit {"Drop Last"} issue is reported by several researchers~\cite{tfb, qiu2025tab, li2025TSFM-Bench}. That is, in some previous works evaluating the model on test set with drop-last=True setting may cause additional errors related to test batch size. In our experiment, to ensure fair comparison in the future, we set the drop last to False for all baselines to avoid this issue.

\textbf{Full-shot results.}  As shown in Table \ref{tab:fullshot}, we also present the results of the ROSE in 10\% few-shot setting. Key observations are summarized as follows. First, as a general forecasting model, ROSE \textbf{\textit{achieves superior performance compared to the six state-of-the-art baselines with full-data training, achieving an average MSE reduction of 15\%,}} which shows that our decomposed frequency learning and register help to learn generalized representations from large-scale datasets and adaptively transfer the multi-domain information to specific downstream scenarios. Second, we observe that ROSE in 10\% few-shot setting shockingly \textbf{\textit{improves a large margin as MSE reduction in average exceeding 12\% over the baselines trained with full data.}} This observation validates the transferability of ROSE pre-trained with large multi-source data.
\begin{table*}[h]
  \centering
  \vspace{-10pt}
  \caption{The results for ROSE in full-shot setting and 10\% few-shot setting, compared with other methods in full-shot setting. The average results of all predicted lengths are listed here.} 
  \resizebox{\linewidth}{!}{
    \begin{tabular}{c|cc|cc|cc|cc|cc|cc|cc|cc}
    \toprule
    Models & \multicolumn{2}{c|}{ROSE} & \multicolumn{2}{c|}{ROSE (10\%)} & \multicolumn{2}{c|}{ITransformer} & \multicolumn{2}{c|}{PatchTST} & \multicolumn{2}{c|}{Timesnet} & \multicolumn{2}{c|}{Dlinear} & \multicolumn{2}{c|}{GPT4TS} & \multicolumn{2}{c}{$\textbf{S}^2$IP-LLM} \\
    \midrule
    Metric & MSE   & MAE   & MSE   & MAE   & MSE   & MAE   & MSE   & MAE   & MSE   & MAE   & MSE   & MAE   & MSE   & MAE & MSE   & MAE \\
    \midrule
    ETTh1 & \textcolor[rgb]{ 1,  0,  0}{\textbf{0.391}} & \textcolor[rgb]{ 1,  0,  0}{\textbf{0.414}} & \textcolor[rgb]{ .161,  .447,  .957}{\underline{0.397}} & \textcolor[rgb]{ .161,  .447,  .957}{\underline{0.419}} & 0.439  & 0.448  & 0.413 & 0.434 & 0.582  & 0.533  & 0.416  & 0.436  & 0.427  & 0.426  & 0.406  & 0.427  \\
    \midrule
    ETTh2 & \textcolor[rgb]{ 1,  0,  0}{\textbf{0.331}} & \textcolor[rgb]{ 1,  0,  0}{\textbf{0.374}} & \textcolor[rgb]{ .161,  .447,  .957}{\underline{0.335}} & \textcolor[rgb]{ .161,  .447,  .957}{\underline{0.380}} & 0.374  & 0.406  & \textcolor[rgb]{ 1,  0,  0}{\textbf{0.331}} & 0.381 & 0.409  & 0.438  & 0.508  & 0.485  & 0.354  & 0.394  & 0.347  & 0.391  \\
    \midrule
    ETTm1 & \textcolor[rgb]{ 1,  0,  0}{\textbf{0.341}} & \textcolor[rgb]{ 1,  0,  0}{\textbf{0.367}} & 0.349 & \textcolor[rgb]{ .161,  .447,  .957}{\underline{0.372}} & 0.362  & 0.391  & 0.353 & 0.382 & 0.490  & 0.464  & 0.356  & 0.378  & 0.352  & 0.383  & \textcolor[rgb]{ .161,  .447,  .957}{\underline{0.343 }} & 0.379  \\
    \midrule
    ETTm2 & \textcolor[rgb]{ 1,  0,  0}{\textbf{0.246}} & \textcolor[rgb]{ 1,  0,  0}{\textbf{0.305}} & \textcolor[rgb]{ .161,  .447,  .957}{\underline{0.250}} & \textcolor[rgb]{ .161,  .447,  .957}{\underline{0.308}} & 0.269  & 0.329  & 0.256 & 0.317 & 0.317  & 0.358  & 0.259  & 0.325  & 0.266  & 0.326  & 0.257  & 0.319  \\
    \midrule
    Weather & \textcolor[rgb]{ 1,  0,  0}{\textbf{0.217}} & \textcolor[rgb]{ 1,  0,  0}{\textbf{0.251}} & 0.224 & \textcolor[rgb]{ .161,  .447,  .957}{\underline{0.252}} & 0.233  & 0.271  & 0.226 & 0.264 & 0.329  & 0.336  & 0.239  & 0.289  & 0.237  & 0.270  & \textcolor[rgb]{ .161,  .447,  .957}{\underline{0.222 }} & 0.259  \\
    \midrule
    Electricity & \textcolor[rgb]{ 1,  0,  0}{\textbf{0.155 }} & \textcolor[rgb]{ 1,  0,  0}{\textbf{0.248 }} & 0.164  & \textcolor[rgb]{ .161,  .447,  .957}{\underline{0.253 }} & 0.164  & 0.261  & \textcolor[rgb]{ .161,  .447,  .957}{\underline{0.159}} & \textcolor[rgb]{ .161,  .447,  .957}{\underline{0.253}} & 0.195  & 0.296  & 0.166  & 0.267  & 0.167  & 0.263  & 0.161  & 0.257  \\
    \midrule
    Traffic & \textcolor[rgb]{ 1,  0,  0}{\textbf{0.390}} & \textcolor[rgb]{ 1,  0,  0}{\textbf{0.264}} & 0.418 & \textcolor[rgb]{ .161,  .447,  .957}{\underline{0.278}} & 0.397  & 0.282  & \textcolor[rgb]{ .161,  .447,  .957}{\underline{0.391}} & \textcolor[rgb]{ 1,  0,  0}{\textbf{0.264}} & 0.623  & 0.333  & 0.433  & 0.305  & 0.414  & 0.294  & 0.405  & 0.286  \\
    \bottomrule

    \end{tabular}}%
    \vspace{-10pt}
  \label{tab:fullshot}%
\end{table*}%

\begin{figure}[h]
    \centering
    \vspace{-5pt}
    \includegraphics[width=1\linewidth]{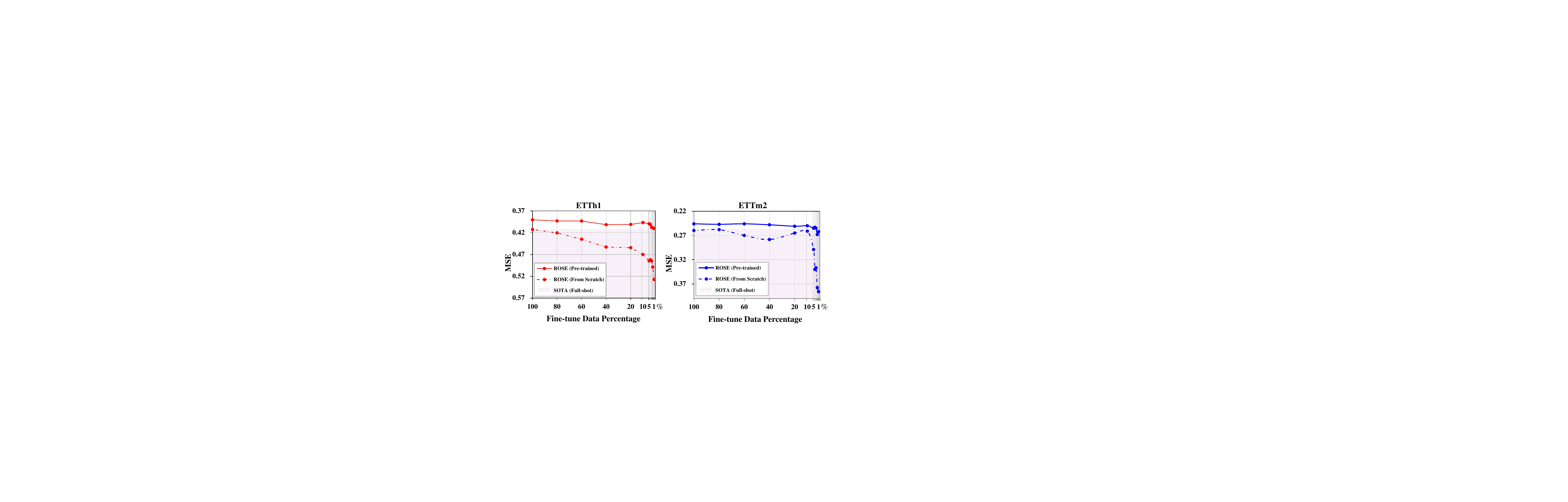}
    \vspace{-15pt}
    \caption{The forecasting results of ROSE obtained by training from scratch and fine-tuning from the pre-trained model. The right, upper corner is the best case.}
    \label{fig: Finetune-percentage}
    \vspace{-1em}
\end{figure}

\textbf{Few-shot results.} The results under the 10\% few-shot setting are presented in Table \ref{tab:fewshot_few_results} in Appendix~\ref{fewshot}. ROSE outperforms advanced models when training data is scarce in the target domain.
% \begin{figure}[h]
%     \centering
%     \vspace{-5pt}
%     \includegraphics[width=1\linewidth]{Figures/Finetune_rate.pdf}
%     \vspace{-15pt}
%     \caption{The forecasting results of ROSE obtained by training from scratch and fine-tuning from the pre-trained model. The right, upper corner is the best case.}
%     \label{fig: Finetune-percentage}
%     \vspace{-1em}
% \end{figure}
Figure \ref{fig: Finetune-percentage} shows the performance of pre-trained ROSE and ROSE trained from scratch on ETTh1 and ETTm2 with different fine-tuning data percentages, noting the best baselines in full-shot setting. 
The pre-trained ROSE shows stable, superior performance even with limited fine-tuning samples. Specifically, the pre-trained ROSE \textbf{\textit{exceeds SOTA performance with only 1\% train data for ETTh1 and 2\% for ETTm2.}} Moreover, compared to the ROSE trained from scratch, the pre-trained ROSE exhibits a slower decline in prediction performance with the reduction of fine-tuning data, demonstrating the impressive generalization ability of ROSE through pre-training.

\vspace{-1em}
\subsection{Zero-shot Forecasting}
\textbf{Setting.} In this section, to ensure a fair comparison, we conduct zero-shot predictions for each foundational model on downstream datasets not included in their pre-training data. It is worth noting that, unlike a few foundation models~\citep{moiral} that require much longer inputs to achieve better predictive performance, we fix the input length of all baselines to 512 without considering longer input lengths, as many real-world scenarios could offer very limited samples.

% 在这节中，为了公平对比，我们让每个基础模型在没有被其预训练数据包含的下游数据集进行零样本预测。值得注意的是，与少数一些需要much longer输入才能达到较好预测表现的foundation models不同的是，as many real-world scenarios could offer very limited samples，我们固定所有基础模型的输入长度为512而不考虑更长的输入长度。

% 我们能够看到我们的模型在大多数数据集上体现出良好的性能。值得注意的是，ROSE相比于其他基础模型，非常的轻量以及高效，具体将会于section 4.3中进行详细分析。

\textbf{Results.} As shown in Table \ref{tab:zeroshot}, ROSE significantly outperforms across the majority of datasets, \textbf{\textit{achieving an average reduction of 15\% in MSE.}} In comparison to Timer and Moirai, ROSE achieves average MSE reductions of 9\% and 6\%, respectively, and demonstrates a remarkable 43\% relative improvement over Moment. Notably, ROSE stands out not only for its superior performance but also for its exceptionally lightweight and efficient design, which sets it apart from other foundational models. Detailed analysis of these aspects will be presented in Section~\ref{analysis}.

\begin{table}[h]
  \centering
  \vspace{-5pt}
  \caption{The results for ROSE and other foundation models in the zero-shot setting. The average results of all predicted lengths are listed here. We use '-' to indicate that the dataset has been involved in the model's pre-training, and thus not used for testing.}
  \resizebox{1\linewidth}{!}{
    \begin{tabular}{c|c|c|c|c|c|c}
    \toprule
    Models & \multicolumn{1}{c|}{ROSE} & \multicolumn{1}{c|}{Timer} & \multicolumn{1}{c|}{MOIRAI} & \multicolumn{1}{c|}{Chronos} & \multicolumn{1}{c|}{TimesFM} & \multicolumn{1}{c}{Moment} \\
    \midrule
    Metric & MSE      & MSE    & MSE     & MSE    & MSE     & MSE    \\
    \midrule
    ETTh1 & \textcolor[rgb]{ 1,  0,  0}{\textbf{0.401 }} &  \textcolor[rgb]{ .161,  .447,  .957}{\underline{0.451 }} &  0.475   & 0.560   & 0.489   & 0.708   \\
    \midrule
    ETTh2 & \textcolor[rgb]{ 1,  0,  0}{\textbf{0.346 }} &  \textcolor[rgb]{ .161,  .447,  .957}{\underline{0.366 }} &  0.379  &  0.392   & 0.396    & 0.392  \\
    \midrule
    ETTm1 & \textcolor[rgb]{ .161,  .447,  .957}{\underline{0.525 }} &  0.544   & 0.714  & 0.636  & \textcolor[rgb]{ 1,  0,  0}{\textbf{0.434 }}  & 0.697  \\
    \midrule
    ETTm2 & \textcolor[rgb]{ 1,  0,  0}{\textbf{0.299 }} & 0.360    & 0.343   & \textcolor[rgb]{ .161,  .447,  .957}{\underline{0.313 }}   & 0.320  & 0.319    \\
    \midrule
    Weather & \textcolor[rgb]{ 1,  0,  0}{\textbf{0.265 }}  & 0.292  & \textcolor[rgb]{ .161,  .447,  .957}{\underline{0.267 }} & 0.288    & -       & 0.291  \\
    \midrule
    Electricity & \textcolor[rgb]{ 1,  0,  0}{\textbf{0.234 }} & 0.297   & \textcolor[rgb]{ .161,  .447,  .957}{\underline{0.241 }}   & 0.245   & -      & 0.861   \\
    \midrule
    Traffic & \textcolor[rgb]{ 1,  0,  0}{\textbf{0.588 }}  & \textcolor[rgb]{ .161,  .447,  .957}{\underline{0.613 }}   & -     & 0.615  & -     & 1.411  \\
    \bottomrule

    \end{tabular}}%

  \label{tab:zeroshot}%
\end{table}%

\subsection{Model Analysis}
\label{analysis}

\textbf{Efficiency analysis.}
%由于采用了重构与预测双预训练任务，使得我们的模型在具备极强泛化性的同时，不需要任何下游数据进行微调即可进行预测。为了验证ROSE zero-shot的能力以及对模型的推理性能进行分析，我们对近期展示出Zeroshot能力的的模型以及进行对比，具体为Timer，MOIRAI，PatchTST和GPT4TS。其中Timer与MOIRAI我们直接采用开源的参数，而PatchTST与GPT4TS则采用与我们相同的预训练数据集，只在预测任务上进行训练，具体实施细节见Appendix。结果如图6所示，我们发现ROSE在模型参数和推理时间远小于其余模型除了PatchTST的情况下，在Zeroshot情景下取得了接近最优的结果。这证明了我们模型具备极强的泛化性，以及可扩展性。
%我们对比了所有模型的参数量、推理速度（perbatch）以及在ETT（4 subsets）上的zeroshot性能均值。具体实验细节和结果在附录。
%如图所示，ROSE在包含最少参数量以及显著的效率优势的情况下展现了整体最优越的zeroshot性能。ROSE拥有毫秒级别的推理速度，为次快模型（moirai）的十分之一。这肯定了ROSE强大的大的通用性和可扩展性。
% To exhibit the performance and efficiency advantages of ROSE in time series forecasting, we evaluate the number of parameters, inference time per batch, and average performance on ETT (4 subsets) in the zero-shot setting for different foundation models. Specific implementation details and results can be found in Appendix~\ref{appendix_zeroshot}.
% To exhibit the performance and efficiency advantages of ROSE in time series forecasting, we evaluate the number of parameters for ROSE and all baselines.
% Moreover, for ROSE and each foundation models, we evaluate its average performance on ETTh1 and ETTh2 datasets in zero-shot setting, as well as its average testing time.
To exhibit the performance and efficiency advantages of ROSE, we compare its parameter count to other foundation models and evaluate their performance and testing time averaged on ETTh1 and ETTh2 datasets in zero-setting. 
Similarly, for each specific model, we evaluate its parameter count as well as its performance in full-shot setting and training-to-testing time averaged on the same datasets.
Specific implementation details and results can be found in Appendix~\ref{appendix_zeroshot}.

\begin{figure}[h]
    \centering
    \vspace{-10pt}
    \includegraphics[width=0.9\linewidth]{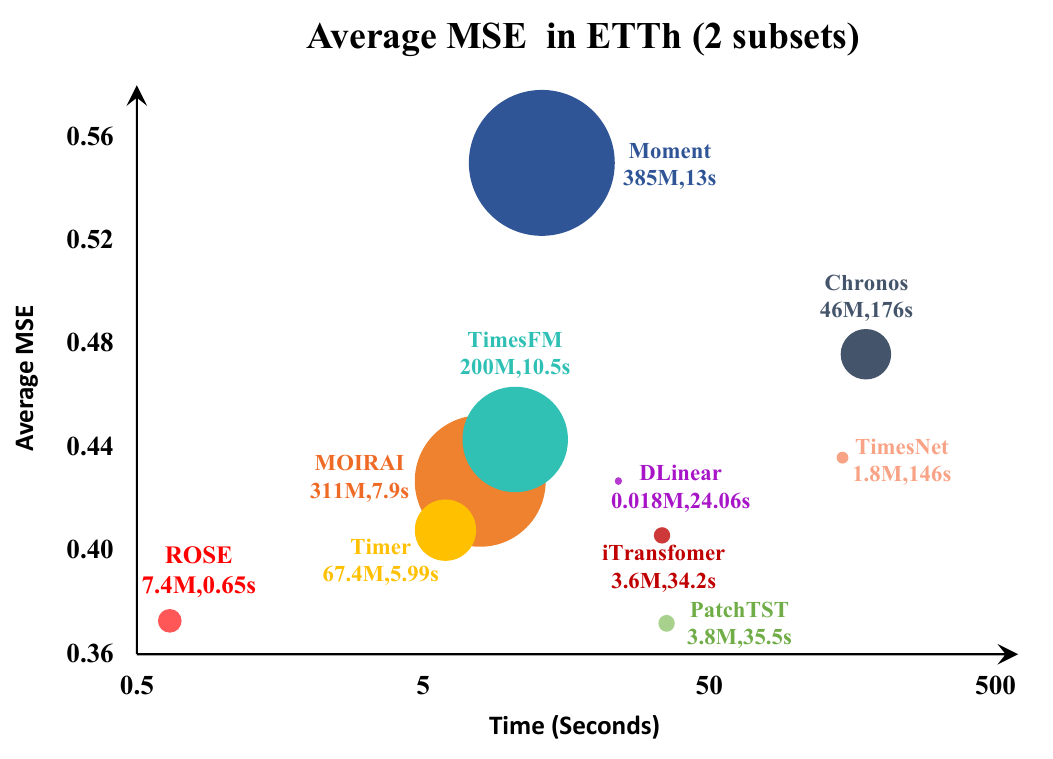}
    \vspace{-15pt}
    \caption{Model performance, number of parameter and efficiency comparison. }
    \vspace{-10pt}
    \label{fig:Efficiency}
\end{figure}
% As shown in Figure \ref{fig:Efficiency}, ROSE is a light weight general model with 7.4M parameters and millisecond-level inference speeds. ROSE's inference speed and number of parameters are only about one-tenth of the second fastest/smallest model (Timer). Importantly, ROSE exhibits superior zero-shot performance while using the least number of parameters, which shows significant efficiency advantages compared to other foundation models.
% This well demonstrates the robust generalization and scalability of ROSE. Compared to the too-large-scale foundation models, ROSE may better meet the need for general models in real scenarios, especially when the resources are limited.
%

As shown in Figure~\ref{fig:Efficiency}, ROSE is a lightweight general model with 7.4M parameters and short inference time, which are only about one-tenth of the second fastest/smallest foundation model (Timer). Importantly, ROSE uses the least number of parameters among foundation models, with its parameter count approaching that of specific models, while exhibiting superior zero-shot performance.
% which shows significant efficiency advantages compared to other foundation models. This well demonstrates the robust generalization and scalability of ROSE.
% which shows the robust generalization and scalability of ROSE.
This is attributed to our proposed decomposed frequency learning that enhances the comprehension of time series. Concurrently, the TS-Register achieves the adaptive transfer thus efficiently adapting to downstream tasks without the need of scaling up to achieve strong generalizability.
Compared to foundation models with large scale, ROSE
% , with competitive performance against specific models trained separately for downstream data, 
may better meet the need for general models in real scenarios that require high computational and parameter efficiency as well as high prediction accuracy with scarce downstream data.
% with competitive performance against specific models trained separately for downstream data,  low memory requirements, as well as substantial high efficiency.

\begin{figure}[h]    
    \vspace{-1em}
    \centerline{\includegraphics[width=1\linewidth]{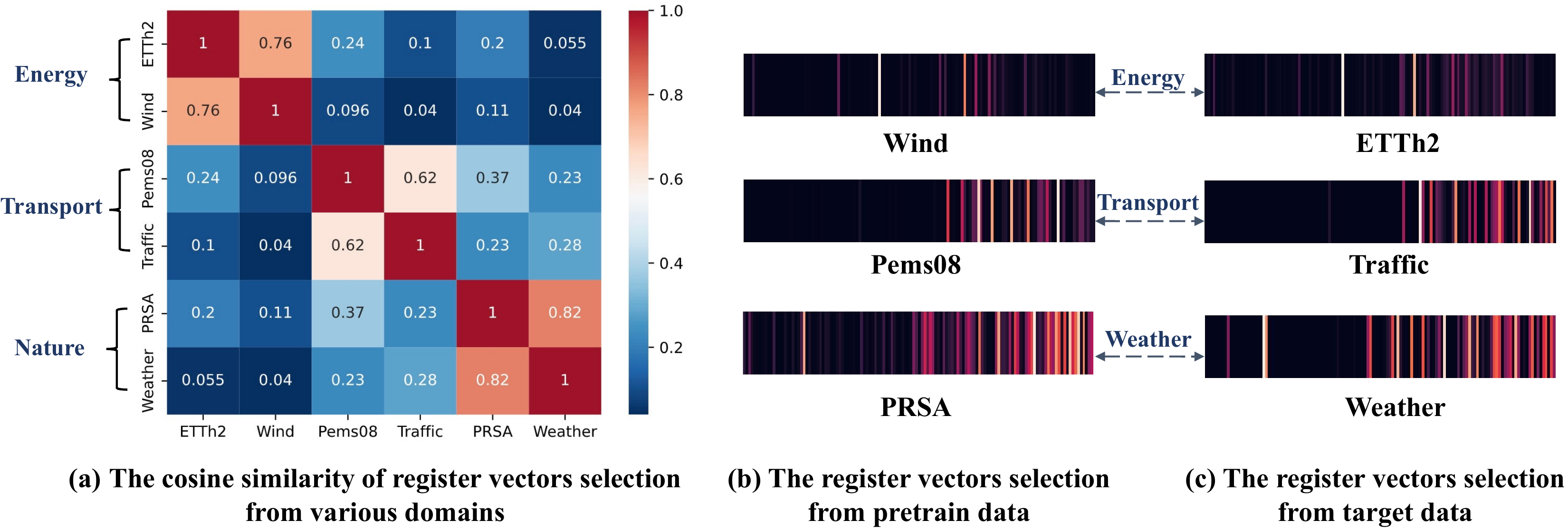}}\vspace{-1mm}
    \caption{Visualization of TS-Register. The calculation of cosine similarity is in the Appendix~\ref{cosine}.}
    \vspace{-1em}
    \label{fig:visual}
\end{figure}

\textbf{Visualization of TS-Register.} To validate the TS-Register's capability to transfer domain-specific information adaptively from pre-training datasets to target datasets, we visualize the cosine similarity of register vector selections from datasets across different domains. As shown in Figure~\ref{fig:visual}(a), the cosine similarity is higher for datasets within the same domain and lower between different domains. We also visualize the register vector selections from different datasets in Figures~\ref{fig:visual}(b) and (c), where datasets from the same domain show similar visualizations. This confirms the TS-Register's capability of adaptive transfer from multi-source to target datasets across various domains.

% The calculation of cosine similarity is in the Appendix

\textbf{Scalability and sensitivity.} The scalability analysis of ROSE's model size and pre-training data size are presented in Appendix~\ref{appendix_scalability}. The sensitivity analyses for the upper bound $a$ of the thresholds, the number of masked series $K_\mathrm{f}$, the number of register tokens $N_\text{r}$, the size of register $H$ and number of selections $k$ in Top-K strategy are presented in Appendix~\ref{appendix_sensitivity}. 
\subsection{Ablation Studies}
%为了验证模型架构的有效性，我们针对重构任务，预测任务以及AQP在10%数据的few shot场景下进行了消融实验。表三展示了不同模块的作用。AQP模块能够在预训练的过程中将时序数据不同domain的信息进行整合聚类，从而帮助下游预测任务对不同数据集进行自适应迁移，我们在下文对此进行了可视化分析。在W/O预测任务的设置之下，我们发现ETTh1和ETTh2的效果显著下降，可能的原因是下游需要需要重新训练预测头，而这两个数据集在fewshot场景下的样本数量较少，从而导致预测头训练不充分，因此预测任务可以有效提升数据稀缺情况下模型的性能。此外，我们发现在W/O重构任务的设置之下，ETTm1和ETTm2的发生了明显的负迁移现象。这是因为预测任务更容易收到预训练数据中数据偏差影响。
\textbf{Model architecture.} To validate effectiveness of our model design, we perform ablation studies on TS-Register, prediction tasks, and reconstruction task in 10\% few-shot setting. Table~\ref{tab:ablation_1}  shows the impact of each module. The TS-Register leverages multi-domain information during pre-training, aiding adaptive transfer to downstream datasets, as further discussed in Section~\ref{analysis}. The prediction tasks enhance performance in data-scarce situations. Without it, performance significantly drops on ETTh1 and ETTh2 with limited samples. Without the reconstruction task, our model shows negative transfer effects on ETTm1 and ETTm2, likely due to the prediction task making the model more susceptible to pre-training data biases.

\textbf{Masking method.} 
%进一步，为了验证MFM的有效性，我们选取了两种主流的时域掩码建模方法，分别为Patch Mask【引用】与Multi Patch mask【引用】，以及只采用单次频率掩码重构的Freq Mask方法来替换MFM。首先，我们发现从频域建模比从时域建模更能提升模型效果。同时我们发现，多次频域掩码使模型达到了更好的效果，这证明了多频率信息视角有助于模型更好的理解时序。
% To further validate the effectiveness of decomposed frequency learning, we replaced multi-frequency mask with different mask modeling methods, including two mainstream time-domain mask modeling methods: patch mask \cite{patchtst} and multi patch mask \cite{simmtm}, as well as the single frequency mask method, which mask time series in frequency domain only once. The results indicate that modeling in the frequency domain yields superior improvements in model performance compared to time-domain modeling. Furthermore, we observe that applying multi-frequency mask results in enhanced outcomes,  thus affirming that incorporating a multi-frequency perspective significantly aids the model in more accurately comprehending temporal patterns.
To further validate the effectiveness of decomposed frequency learning, we replace the multi-frequency masking with different masking methods, including two mainstream time-domain methods: patch masking~\citep{patchtst} and multi-patch masking~\citep{simmtm}, as well as random frequency masking~\citep{fraug}. The results in Table~\ref{tab:ablation_2} show that random frequency masking and patch masking led to negative transfer on ETTm1 and ETTm2, likely due to significant disruption of the original time series, causing overfitting. In contrast, multi-patch masking and multi-frequency masking resulted in positive transfer across all datasets by preventing excessive disruption. Multi-frequency masking achieved better results, demonstrating its ability to help the model understand temporal patterns from a multi-frequency perspective. We also compare with some other pre-training tasks in Table~\ref{Appendix_ablation_tables} in Appendix~\ref{Appendix_ablation}.

\begin{table}[h]
  \centering
  \vspace{-5pt}
  \caption{Ablations on key components of model architecture, including TS-register, prediction task and reconstruction task. The average results of all predicted lengths are listed here.}
  \resizebox{1\linewidth}{!}{
    \begin{tabular}{c|c|c|c|c|c|c|c|c|c}
    \toprule
    \multicolumn{2}{c|}{\multirow{2}[4]{*}{Design}} & \multicolumn{2}{c|}{ETTm1} & \multicolumn{2}{c|}{ETTm2} & \multicolumn{2}{c|}{ETTh1} & \multicolumn{2}{c}{ETTh2} \\
\cmidrule{3-10}    \multicolumn{2}{c|}{} & MSE   & MAE   & MSE   & MAE   & MSE   & MAE   & MSE   & MAE \\
    \midrule
    \multicolumn{2}{c|}{ROSE} & \textbf{0.349 } & \textbf{0.372 } & \textbf{0.250 } & \textbf{0.308 } & \textbf{0.397 } & \textbf{0.419 } & \textbf{0.335 } & \textbf{0.380 } \\
    \midrule
    \multicolumn{1}{c|}{\multirow{3}[6]{*}{w/o}} & TS-Register & 0.354 & 0.378 & 0.256 & 0.312 & 0.418 & 0.427 & 0.355 & 0.390 \\
\cmidrule{2-10}          & Prediction Task & 0.360  & 0.384 & 0.257 & 0.314 & 0.422 & 0.438 & 0.372 & 0.410 \\
\cmidrule{2-10}          & Reconstruction Task & 0.387 & 0.403 & 0.269 & 0.327 & 0.412 & 0.428 & 0.361 & 0.399 \\
    \midrule
    \multicolumn{2}{c|}{From scratch} & 0.371  & 0.391  & 0.261  & 0.318  & 0.470  & 0.480  & 0.400  & 0.425  \\
    \bottomrule
  \end{tabular}}%
  \vspace{-12pt}
  \label{tab:ablation_1}%
\end{table}%
\begin{table}[h]
  \centering
  \vspace{-5pt}
  \caption{Ablations on decomposed frequency learning, where we replace Multi-freq masking with other masking methods. The average results of all predicted lengths are listed here.}
  \resizebox{1\linewidth}{!}{
    \begin{tabular}{c|c|c|c|c|c|c|c|c|c}
    \toprule
    \multicolumn{2}{c|}{\multirow{2}[4]{*}{Design}} & \multicolumn{2}{c|}{ETTm1} & \multicolumn{2}{c|}{ETTm2} & \multicolumn{2}{c|}{ETTh1} & \multicolumn{2}{c}{ETTh2} \\
\cmidrule{3-10}    \multicolumn{2}{c|}{} & MSE   & MAE   & MSE   & MAE   & MSE   & MAE   & MSE   & MAE \\
    \midrule
    \multicolumn{2}{c|}{ROSE} & \textbf{0.349 } & \textbf{0.372 } & \textbf{0.250 } & \textbf{0.308 } & \textbf{0.397 } & \textbf{0.419 } & \textbf{0.335 } & \textbf{0.380 } \\
    \midrule
    \multicolumn{2}{c|}{Random Freq Masking}& 0.381 & 0.397 & 0.261 & 0.324 & 0.410 & 0.427 & 0.374 & 0.405 \\
    \midrule
    \multicolumn{2}{c|}{Multi-Patch Masking}& 0.356 & 0.379 & 0.259 & 0.316 & 0.404 & 0.426 & 0.349 & 0.389 \\
    \midrule
    \multicolumn{2}{c|}{Patch Masking}& 0.378 & 0.400   & 0.261 & 0.319 & 0.408 & 0.432 & 0.375 & 0.407 \\
%     \multicolumn{1}{c|}{\multirow{3}[6]{*}{\makecell[c]{Replace \\ Multi-Frequency Masking}}} & \makecell[c]{Random Freq Masking}& 0.381 & 0.397 & 0.261 & 0.324 & 0.410 & 0.427 & 0.374 & 0.405 \\
% \cmidrule{2-10}          & Multi-Patch Masking & 0.356 & 0.379 & 0.259 & 0.316 & 0.404 & 0.426 & 0.349 & 0.389 \\
% \cmidrule{2-10}          & Patch Masking & 0.378 & 0.400   & 0.261 & 0.319 & 0.408 & 0.432 & 0.375 & 0.407 \\
    \midrule
    \multicolumn{2}{c|}{From scratch} & 0.371  & 0.391  & 0.261  & 0.318  & 0.470  & 0.480  & 0.400  & 0.425  \\
    \bottomrule
  \end{tabular}}%
  \vspace{-12pt}
  \label{tab:ablation_2}%
\end{table}%

\section{Conclusion and Future Work}
In this work, we propose ROSE, a novel general model, addressing the challenges of leveraging multi-domain datasets for enhancing downstream prediction task performance. ROSE utilizes decomposed frequency learning and TS-Register to capture generalized and domain-specific representations, enabling improved fine-tuning results, especially in data-scarce scenarios. Our experiments demonstrate ROSE's superior performance over baselines with both full-data and few-data fine-tuning, as well as its impressive zero-shot capabilities. Future efforts will concentrate on expanding pre-training datasets and extending ROSE's applicability across diverse time series analysis tasks, e.g., classification. 
We provide our code at \url{https://github.com/decisionintelligence/ROSE}.

\section*{Impact Statement}

This paper presents work whose goal is to advance the field of Machine Learning. There are many potential societal consequences of our work, none which we feel must be specifically highlighted here.

\section*{Acknowledgements}
This work was partially supported by National Natural Science Foundation of China (62372179). Chenjuan Guo is the corresponding author of the work.

\nocite{langley00}

\bibliography{example_paper}
\bibliographystyle{icml2025}

%%%%%%%%%%%%%%%%%%%%%%%%%%%%%%%%%%%%%%%%%%%%%%%%%%%%%%%%%%%%%%%%%%%%%%%%%%%%%%%
%%%%%%%%%%%%%%%%%%%%%%%%%%%%%%%%%%%%%%%%%%%%%%%%%%%%%%%%%%%%%%%%%%%%%%%%%%%%%%%
% APPENDIX
%%%%%%%%%%%%%%%%%%%%%%%%%%%%%%%%%%%%%%%%%%%%%%%%%%%%%%%%%%%%%%%%%%%%%%%%%%%%%%%
%%%%%%%%%%%%%%%%%%%%%%%%%%%%%%%%%%%%%%%%%%%%%%%%%%%%%%%%%%%%%%%%%%%%%%%%%%%%%%%
\newpage
\appendix
\onecolumn
\section{Appendix}
\subsection{Implementation Details}

\subsubsection{Pre-training Datasets}
\label{pretraining_datasets}

We use multi-source datasets in pre-training which contain subsets of Monash~\citep{monash}, UEA~\citep{uea} and UCR~\citep{ucr} time series datasets, as well as some other time series classical datasets~\citep{prsa,tdbrain,pems,fred, nn5}. The final list of all pre-training datasets is shown in Table~\ref{tab: pretraining datasets}. There is no overlap between the pre-training datasets and the target datasets. It is worth noting that the dataset weather in the pre-training dataset is a univariate dataset, which is different to the multivariate dataset weather in the target task.
The pre-trained datasets can be categorized into 6 different domains according to their sources: Energy, Nature, Health, Transport, and Web. The sampling frequencies of the datasets show a remarkable diversity, ranging from millisecond samples to monthly samples, which reflects the diverse application scenarios and complexity of the real world. For all pre-training datasets, we split them into univariate sequences and train them in a channel-independent manner.

\begin{table*}[h]
  \centering
  \vspace{-1em}
  \caption{List of pretraining datasets.}
  \resizebox{0.7\linewidth}{!}{
    \begin{tabular}{c|c|c|c|c}
    \toprule
    Domain & Dataset & Frequency & Time Pionts & Source \\
    \midrule
    \multirow{4}[8]{*}{Energy} & Aus. Electricity Demand & Half Hourly & 1155264 & Monash\citep{monash} \\
\cmidrule{2-5}          & Wind  & 4 Seconds & 7397147 & Monash\citep{monash} \\
\cmidrule{2-5}          & Wind Farms & Minutely & 172178060 & Monash\citep{monash} \\
\cmidrule{2-5}          & Solar & 10 Minutes & 7200720 & Monash\citep{monash} \\
\cmidrule{2-5}          & Solar Power & 4 Seconds & 7397222 & Monash\citep{monash} \\
\cmidrule{2-5}          & London Smart Meters & Half Hourly & 166527216 & Monash\citep{monash} \\
    \midrule
    \multirow{11}[22]{*}{Nature} & Phoneme & -     & 2160640 & UCR\cite{ucr} \\
\cmidrule{2-5}          & EigenWorms & -     & 27947136 & UEA\citep{uea} \\
\cmidrule{2-5}          & PRSA  & Hourly & 4628448 & \citep{prsa} \\
\cmidrule{2-5}          & Temperature Rain & Daily & 23252200 & Monash\citep{monash} \\
\cmidrule{2-5}          & StarLightCurves & -     & 9457664 & UCR\citep{ucr} \\
\cmidrule{2-5}          & Worms & 0.033 Seconds & 232200 & UCR\citep{ucr} \\
\cmidrule{2-5}          & Saugeen River Flow & Daily & 23741 & Monash\citep{monash} \\
\cmidrule{2-5}          & Sunspot & Daily & 73924 & Monash\citep{monash} \\
\cmidrule{2-5}          & Weather & Daily & 43032000 & Monash\citep{monash} \\
\cmidrule{2-5}          & KDD Cup 2018 & Daily & 2942364 & Monash\cite{monash} \\
\cmidrule{2-5}          & US Births & Daily & 7305  & Monash\citep{monash} \\
    \midrule
    \multirow{7}[14]{*}{Health} & MotorImagery & 0.001 Seconds & 72576000 & UEA\citep{uea} \\
\cmidrule{2-5}          & SelfRegulationSCP1 & 0.004 Seconds & 3015936 & UEA\citep{uea} \\
\cmidrule{2-5}          & SelfRegulationSCP2 & 0.004 Seconds & 3064320 & UEA\citep{uea} \\
\cmidrule{2-5}          & AtrialFibrillation & 0.008 Seconds & 38400 & UEA\citep{uea} \\
\cmidrule{2-5}          & PigArtPressure & -     & 624000 & UCR\citep{ucr} \\
\cmidrule{2-5}          & PIGCVP & -     & 624000 & UCR\citep{ucr} \\
\cmidrule{2-5}          & TDbrain & 0.002 Seconds & 79232703 & \citep{tdbrain} \\
    \midrule
    \multirow{6}[12]{*}{Transport} & Pems03 & 5 Minute & 9382464 & \citep{pems} \\
\cmidrule{2-5}          & Pems04 & 5 Minute & 5216544 & \citep{pems} \\
\cmidrule{2-5}          & Pems07 & 5 Minute & 24921792 & \citep{pems} \\
\cmidrule{2-5}          & Pems08 & 5 Minute & 3035520 & \citep{pems} \\
\cmidrule{2-5}          & Pems-bay & 5 Minute & 16937700 & \citep{pems} \\
\cmidrule{2-5}          & Pedestrian\_Counts & Hourly & 3132346 & Monash\citep{monash} \\
    \midrule
    Web   & Web Traffic & Daily & 116485589 & Monash\citep{monash} \\
    \midrule
    \multirow{3}[6]{*}{Economic} & FRED\_MD & Monthly & 77896 & \citep{fred} \\
\cmidrule{2-5}          & Bitcoin & Daily & 75364 & Monash\citep{monash} \\
\cmidrule{2-5}          & NN5   & Daily & 87801 & \citep{nn5} \\
    
    \bottomrule

    \end{tabular}}%
  \label{tab: pretraining datasets}%
\end{table*}%

\subsubsection{Evaluation Datasets}
\label{evaluation_datasets}

We use the following 7 multivariate time-series datasets for downstream fine-tuning and forecasting: ETT datasets\footnote{https://github.com/zhouhaoyi/ETDataset} contain 7 variates collected from two different electric transformers from July 2016 to July 2018. 
It consists of four subsets, of which ETTh1/ETTh2 are recorded hourly and ETTm1/ETTm2 are recorded every 15 minutes. 
Traffic\footnote{https://pems.dot.ca.gov/} contains road occupancy rates measured by 862 sensors on freeways in the San Francisco Bay Area from 2015 to 2016, recorded hourly. 
Weather\footnote{https://www.bgc-jena.mpg.de/wetter/} collects 21 meteorological indicators, such as temperature and barometric pressure, for Germany in 2020, recorded every 10 minutes. 
Electricity\footnote{https://archive.ics.uci.edu/ml/datasets/ElectricityLoadDiagrams20112014} contains the electricity consumption of 321 customers from July 2016 to July 2019, recorded hourly. We split each evaluation dataset into train-validation-test sets and detailed statistics of evaluation datasets are shown in Table~\ref{tab: Evaluation Datasets}.

\begin{table}[htbp]
  \centering
  \caption{The statistics of evaluation datasets.}
  \resizebox{0.65\linewidth}{!}{
    \begin{tabular}{c|c|c|c|c|c|c|c}
    \toprule
    Dataset & ETTm1 & ETTm2 & ETTh1 & ETTh2 & Traffic & Weather  & Electricity \\
    \midrule
    Variables & 7     & 7     & 7     & 7     & 862   & 21  & 321 \\
    \midrule
    Timestamps & 69680 & 69680 & 17420 & 17420 & 17544 & 52696 & 26304 \\
    \midrule
    Split Ratio & 6:2:2 & 6:2:2 & 6:2:2 & 6:2:2 & 7:1:2 & 7:1:2 & 7:1:2 \\
    \bottomrule
    \end{tabular}}%
  \label{tab: Evaluation Datasets}%
\end{table}%

\subsubsection{Setting}
\label{setting}
We implemented ROSE in PyTorch~\citep{pytorch} and all the experiments were conducted on 8 NVIDIA A800 80GB GPU. We used ADAM~\citep{adam}  with an initial learning rate of $5 \times 10^{-4}$ and implemented learning rate decay using the StepLR method to implement learning rate decaying pre-training. 
By default, ROSE contains 3 encoder layers and 3 decoder layers with head number of 16 and the dimension of latent space $D = 256$. The patch size for patching is set to 64.

\textbf{Pre-training.} We use $N_\mathrm{r} = 3$ as the number of register tokens and $P = 8$ as the path tokens. We set the input length to 512 for the supervised prediction task with target lengths of 96, 192, 336, and 720. We also set the input length to 512 and mask number $K_\mathbf{f}= 4$. The batch size is set to 8192 in pre-training.

\textbf{Fine-tuning.} We fix the lookback window to 512, and perform predictions with target lengths of 96, 192, 336, and 720, respectively. The number of register tokens $N_\mathrm{r}$ and patch tokens $P$ is the same as in pre-training, and the parameter $k = 3$ in TopK is set when selection vectors are performed in the register.

\textbf{The t-SNE visualization.} We select three datasets (Pems08, PSRA, Electricity) from transport, nature and energy domains respectively and compare the differences in hidden representations between direct transfer and adaptive transfer. Specifically, direct transfer refers to the case where domain specific information is not considered, while adaptive transfer considers domain specific information that is learned by register tokens. We visualized the output of the encoder's hidden representations using t-SNE.

\subsubsection{Baselines}
We select the state-of-the-art models as our baselines in full-shot and few-shot setting, including four specific models: iTransformer~\citep{itransformer}, PatchTST~\citep{patchtst}, TimesNet~\citep{timesnet}, and DLinear~\citep{dlinear}, and two LLM-based models: GPT4TS~\citep{gpt4ts} and $\textbf{S}^2$IP-LLM~\citep{s2ipllm}. In addition, we selected five foundation models for comparison in zero-shot setting, including Timer~\citep{timer}, MOIRAI~\citep{moiral}, Chronos~\citep{chronos}, TimesFM~\citep{timefm} and Moment~\citep{moment}. The zero-shot experiment of Moment is designed based on the reconstruction task in the pre-training phase. Specifically, to ensure consistency between pre-training (reconstruction) and downstream prediction tasks, we actively mask the time periods to be predicted in the input sequence, and directly use the model's reconstruction output for this part as the prediction value.
Moment itself is not designed for zero-shot prediction and does not officially support zero-shot forecasting in this manner. The specific code base for these models is listed in Table~\ref{tab: baseline}:

% Table generated by Excel2LaTeX from sheet 'Baseline'
\begin{table}[h]
  \centering
  \caption{Code repositories for baselines.}
  \resizebox{0.75\linewidth}{!}{
    \begin{tabular}{c|c|c}
    \toprule
    Model Types & Models & \multicolumn{1}{c}{Code Repositories} \\
    \midrule
    \multirow{4}[8]{*}{Small Model} & iTransformer & \multicolumn{1}{c}{\textcolor[rgb]{ 0,  0,  1}{https://github.com/thuml/iTransformer}} \\
\cmidrule{2-3}          & PatchTST & \multicolumn{1}{c}{\textcolor[rgb]{ 0,  0,  1}{https://github.com/yuqinie98/PatchTST}} \\
\cmidrule{2-3}          & TimesNet & \multicolumn{1}{c}{\textcolor[rgb]{ 0,  0,  1}{https://github.com/thuml/TimesNet}} \\
\cmidrule{2-3}          & Dlinear & \multicolumn{1}{c}{\textcolor[rgb]{ 0,  0,  1}{https://github.com/cure-lab/LTSF-Linear}} \\
    \midrule
    \multirow{6}[12]{*}{Foundation Model} & Timer & \multicolumn{1}{c}{\textcolor[rgb]{ 0,  0,  1}{https://github.com/thuml/Large-Time-Series-Model}} \\
\cmidrule{2-3}          & MOIRAI & \multicolumn{1}{c}{\textcolor[rgb]{ 0,  0,  1}{https://github.com/redoules/moirai}} \\
\cmidrule{2-3}          & Chronos & \multicolumn{1}{c}{\textcolor[rgb]{ 0,  0,  1}{https://github.com/amazon-science/chronos-forecasting}} \\
\cmidrule{2-3}          & TimesFM & \multicolumn{1}{c}{\textcolor[rgb]{ 0,  0,  1}{https://github.com/google-research/timesfm/}} \\
\cmidrule{2-3}          & Moment & \multicolumn{1}{c}{\textcolor[rgb]{ 0,  0,  1}{https://anonymous.4open.science/r/BETT-773F/README.md}} \\
    \midrule
    \multirow{2}[4]{*}{LLM-based Model} & GPT4TS & \multicolumn{1}{c}{\textcolor[rgb]{ 0,  0,  1}{https://github.com/DAMO-DI-ML/NeurIPS2023-One-Fits-All}} \\
\cmidrule{2-3}          & S2IP-LLM & \textcolor[rgb]{ 0,  0,  1}{https://github.com/panzijie825/S2IP-LLM} \\
    \bottomrule

    \end{tabular}}%
  \label{tab: baseline}%
\end{table}%

\subsection{Efficiency Analysis}
\label{appendix_zeroshot}
%为了探究模型效率以及验证ROSE 的zero-shot的能力，我们选取了近期具备zero-shot能力的模型，包括 Timer， MOIRAI，以及具备zero-shot潜力的模型包括Patchtst以及GPT4TS，在ETT数据集上进行比较。
%我们对Timer及MOIRAI直接选取了他们最大参数量的模型设置作为baseline。而GPT4TS与PatchTST则使用了预测任务在我们的预训练数据集上进行进行预训练，从而具备zero-shot的能力。具体setting如表所示。
%输入长度：值得注意是的，MOIRAI的结果在5000左右会达到最好，然而这对于其他方法是不公平的，因为如果具备如此长的输入，其余模型在此输入下进行微调可以达到更好的预测结果。因此，为了公平比较，我们的输入长度从96，192，336，512，672，720六个长度中选择最好的结果。
%实验结果：
%作为foundation model 推理效率非常重要，因此我们对ROSE以及五个基础模型在ETT数据集上测试了推理时间。And we evaluate their inference efficiency by recording inference time per batch on the ETTh2 dataset with a batch size of 32.具体结果如表示，我们发现ROSE推理速度比baselines都要快，即使比第二块的Timer也要快十倍左右。在这里引起我们的思考，时序基础模型是否需要非常巨大的参数量，以及现有的时序基础模型是否验证了其架构在时序数据上的scaling law。
As an important aspect of foundation models, inference efficiency is crucial. Therefore, we evaluate the testing time of ROSE and five foundation models in the ETTh1 and ETTh2 dataset in zero-shot setting.
Similarly, we evaluate the time of the entire process of training, validation, and testing for four specific models in the same datasets in full-shot setting. The above experiments all set the batch size to 32.
% We evaluate their inference efficiency by recording the inference time per batch on the ETTh2 dataset with a batch size of 32. 
The specific results are shown in Table~\ref{tab:zero_shot_baseline} and Figure~\ref{fig:Efficiency}. We observe that ROSE maintains its advantage in zero-shot performance while also being significantly faster compared to the baselines, even being approximately ten times faster than the second-fastest foundation model, Timer. This raises our reflection on whether time-series foundation models require extremely large parameter sizes and whether existing time-series foundation models have validated their architectures' scaling laws on time-series data.
% Table generated by Excel2LaTeX from sheet 'Sheet3'
\begin{table}[h]
  \centering
  \vspace{-1em}
  \caption{Efficiency analysis.}
  \resizebox{0.5\linewidth}{!}{
    \begin{tabular}{cccc}
    \toprule
    Model & Parameters & Pre-train datasize & Averaged time \\
    \midrule
    ROSE  & 7.4M  & 0.89B & 0.652s \\
    MOIRAI & 311M  & 27B   & 7.920s \\
    Timer & 67.4M & 1B    & 5.989s \\
    Chronos & 46M   & 84B   & 176s \\
    TimesFM & 200M  & 100B  & 10.5s \\
    Moment & 385M  & 1.13B & 13s \\
    Itransformer & 3.8M  & -     & 34.18s \\
    PatchTST & 3.2M  & -     & 35.47s \\
    TimesNet & 1.8M  & -     & 146s \\
    Dlinear & 0.018M & -     & 24.06s \\
    \bottomrule
    \end{tabular}}%
  \label{tab:zero_shot_baseline}%
\end{table}%

\subsection{Calculation of Cosine Similarity}
\label{cosine}
In Figure~\ref{fig:visual}, we visualize the cosine similarity
of register vector selections.
For each sample in a dataset, during the inference process, $k$ vectors are selected from the register based on the Top-K strategy. We iterate through all samples in the dataset and count the number of times each vector is selected, which allows us to obtain a record vector of length equivalent to the size of the register for each dataset. The $i_{th}$ position in the record vector represents the number of times the $i_{th}$ vector in the register has been selected by samples in the dataset. For a pair of datasets, we can obtain a unique record vector for each dataset, and then we are able to calculate the cosine similarity of two vectors.

\subsection{Scalability}
\label{appendix_scalability}
Scalability is crucial for a general model, enabling significant performance improvements by expanding pre-training data and model sizes. To investigate the scalability of ROSE, we increased both the model size and dataset size and evaluated its predictive performance on four ETT datasets.
\begin{figure}[h]    
    \centerline{\includegraphics[width=0.9\linewidth]{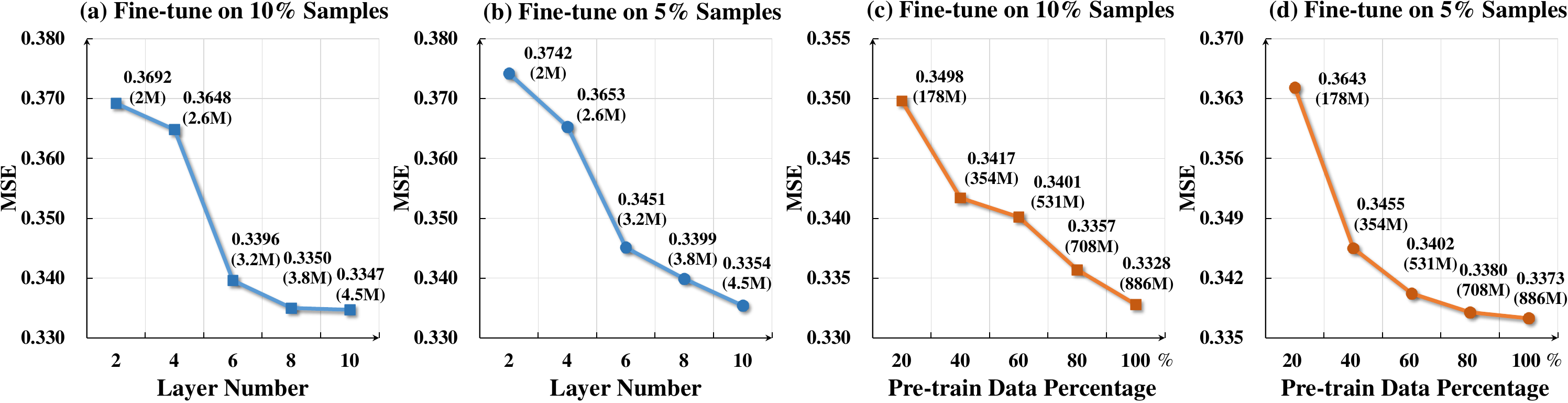}}\vspace{-2mm}
    \caption{(a)/(b): Larger ROSE demonstrates better performance on downstream forecasting. (c)/(d): ROSE pre-trained on larger datasets demonstrates better performance on downstream forecasting.}
    \label{fig: scalability}
    \vspace{-10pt}
\end{figure}

\textbf{Model size.} Constrained by computational resources, we use 40\% pre-training datasets. The results are shown in Figure \ref{fig: scalability}(a) and (b). When maintaining the model dimension, we increased the model layers, increasing model parameters from 2M to 4.5M. This led to 10.37\% and 9.34\% improvements in the few-shot scenario with 5\% and 10\%  downstream data, respectively.

\textbf{Data size.} When keeping the model size, we increase the size of the pre-training datasets from 178M to 887M. The results are shown in Figure \ref{fig: scalability}(c) and (d). The performance of our model steadily improves with the increase in dataset size and achieves improvements of 7.4\% and 4.8\% respectively.
\subsection{Sensitivity}
\label{appendix_sensitivity}
We perform the sensitivity analyses for the upper bound $a$ of the thresholds, the number of masked series $K_{\mathrm{f}}$, the number of register tokens $N_\text{r}$, the size of register $H$ and the number of selections $k$ in Top-K strategy. All the sensitivity experiments present the average results on the four ETT datasets: ETTh1, ETTh2, ETTm1 and ETTm2 under 10\% few-shot setting.

\begin{figure}[h]    
    \centerline{\includegraphics[width=0.8\linewidth]{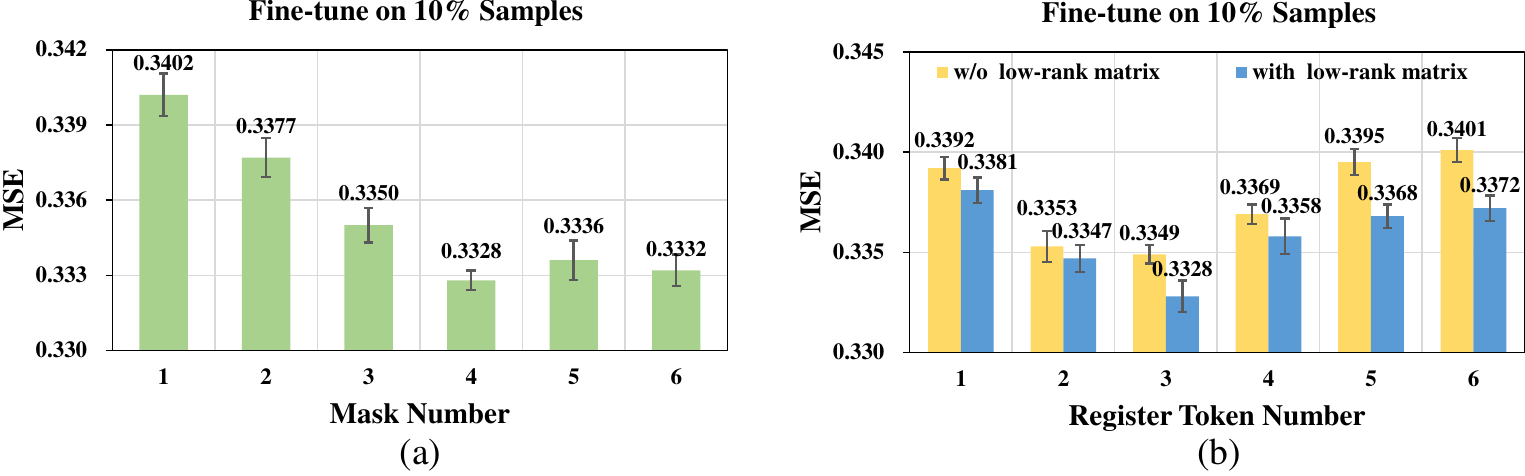}}\vspace{-2mm}
    \vspace{-5pt}
    \caption{(a): Analysis of the number of masked series. (b): Analysis of the number of register tokens.}
    \label{fig: sensitivity}
    \vspace{-10pt}
\end{figure}

%如3.3节所描述，我们提出了decomposed frequency learning ，通过多个阈值在频域对时序进行随机高低频掩码从而将原始时序分解为多个频率分项，使得模型能够从多个频率视角理解时序。In this experiment， we study the influence of the number of masked series one downstream performance. We train ROSE with 1，2，3，4，5， or 6 mask series. On the left side of Fig. 8, we report the results of this analysis. 我们发现随着掩码序列数量的增加，下游效果逐渐变好，这是由于模型能够从分解的频率分项中更好的理解模型，从而导致模型的泛化能力增加。然而随着掩码序列继续增加，下游效果并没有产生显著变化，甚至产生负面影响，这可能是因为掩码序列过多，导致信息冗余。In all our experiments， we kept 4 mask series.
\textbf{Number of masked series.} 
As described in Section \ref{multi_freq_mask}, we propose decomposed frequency learning, which employs multiple thresholds to randomly mask high and low frequencies in the frequency domain, thereby decomposing the original time series into multiple frequency components. This allows the model to understand the time series from multiple frequency perspectives. In this experiment, we study the influence of the number of masked series $K_\mathrm{f}$ on downstream performance. We train ROSE with 1, 2, 3, 4, 5, or 6 mask series. We report the results of this analysis in Figure~\ref{fig: sensitivity}(a). We find that as the number of masked sequences increases, the downstream performance gradually improves. This is because the model can better understand the time series from the decomposed frequency components, which enhances the model's generalization ability. However, more masked series do not bring better downstream performance. This could be due to an excessive number of masked sequences leading to information redundancy. In all our experiments, we keep 4 mask series.

\textbf{Number of register tokens.}
The TS-register module presented in Section~\ref{ts-register} supports the configuration of an arbitrary number of register tokens. In Figure~\ref{fig: sensitivity}(b), we visualize the relationship between the performance on the ETT datasets under a 10\% few-shot setting and the number of register tokens. It is observed that when the number of register tokens ranges from 1 to 6, the model's performance remains relatively stable, with an optimal outcome achieved when the number is set to 3. This phenomenon may be because when the number of register tokens is too small, they contain insufficient domain-specific information, which limits their effectiveness in enhancing the model's performance. Conversely, an excess of register tokens may introduce redundant information, hindering the accurate representation of domain-specific information.
Additionally, we compared the results without the adjustment of a low-rank matrix on the register tokens and found that the incorporation of a low-rank matrix adjustment led to improvements across all quantities of register tokens. This finding underscores the significance of utilizing a low-rank matrix to supplement the register tokens with downstream data-specific information.

\begin{figure}[h]    
    \centerline{\includegraphics[width=1\linewidth]{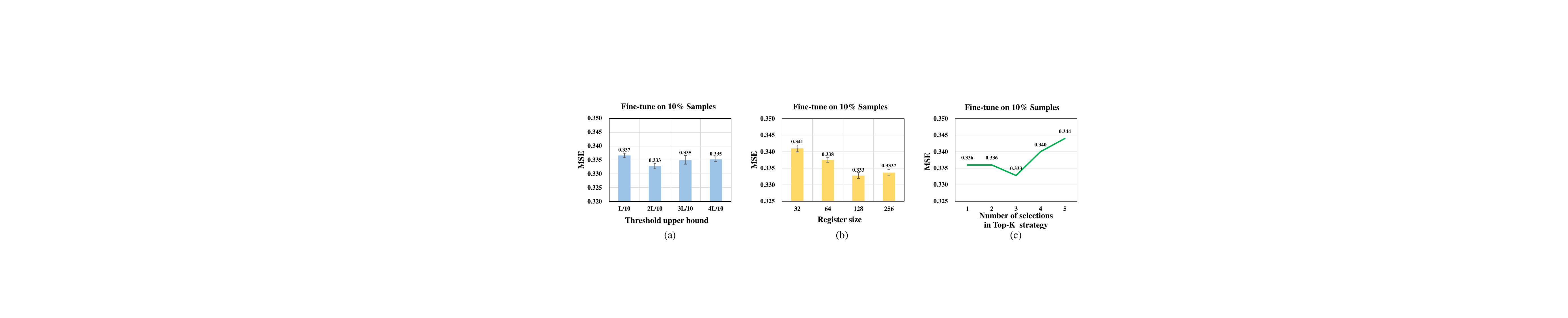}}\vspace{-2mm}
    \vspace{-5pt}
    \caption{(a): Analysis of the threshold upper bound. (b): Analysis of the size of register. (c): Analysis of the number of selections in Top-K strategy.}
    \label{fig: sensitivity_2}
\end{figure}

%我们发现阈值上界对整体效果影响并不大。但是由于一般而言，时序低频信息密度比高频信息密度高。因此上界应偏向低频，使得低频部分和高频部分的信息含量保持平衡。但是同时不应过于偏向左侧，我们发现阈值上界为$L/10$的效果比$L/5$要差，这是因为过于偏左侧会使得低频部分信息含量过少，从而使得重构任务过于困难或者过于简单。通过实验，我们推荐使用$L/5$作为阈值上界。
\textbf{Thresholds upper bound.}  Figure~\ref{fig: sensitivity_2}(a) illustrates the relationship between threshold upper bound and model performance. We have observed that the upper bound of the threshold has a minimal impact on the model's performance. Generally, the information density is higher in low-frequency components compared to high-frequency ones. Therefore, the upper bound of the threshold should be biased towards the low-frequency range to balance the information content between low-frequency and high-frequency components. However, this bias should not be excessive. Our experiments indicate that an upper bound of $L/10$ performs worse than $L/5$ as an overly left-skewed threshold results in insufficient information in the low-frequency range, making the reconstruction task either too difficult or too simple. Based on our findings, we recommend using $L/5$ as the upper bound for the threshold.

\textbf{Register size.} Figure~\ref{fig: sensitivity_2}(b) illustrates the relationship between register size and model performance. The register size determines the upper limit of domain-specific information that the register can store. We can observe that there is a significant improvement in the model effect when the register size is increased from 32 to 128. When the register size exceeds 128, the improvement of the model effect with the increase of register size is no longer obvious. Therefore, we believe that 128 is an appropriate register size for the current pre-training datasets.

\textbf{Number of selections in Top-K strategy.} 
Figure~\ref{fig: sensitivity_2}(c) illustrates the relationship between the number of selections $k$ in Top-K strategy and model performance when we use the register to realize adaptive transfer of domain-specific information in downstream tasks. It can be seen that the model effect performance peaks at 3 tokens at $k$ = 3, which has some advantages over selecting once ($k$ = 1), indicating that the TopK strategy can compensate for the problem of incomplete matching of upstream and downstream domains to some extent. However, too large $k$ will also introduce redundant information and limit the accuracy of domain-specific information transfer.

\subsection{Short-term Forecasting}
\label{appendix_shot-term forcasting}

We also try to apply ROSE to short-term forecasting on the M4~\citep{M4} dataset, which contains the yearly, quarterly and monthly collected univariate marketing data. We follow TimesNet's~\citep{timesnet} setting and metrics (SMAPE, MASE and OWA) for testing. As shown in Table~\ref{tab: Short-term forecasting}, ROSE also exhibits competitive performance on the M4 dataset compared to the baselines.

\begin{table*}[htbp]
    
  \centering
  \caption{Full results on Short-term forecasting.}
  \resizebox{1\linewidth}{!}{
    \begin{tabular}{c|ccc|ccc|ccc|ccc|ccc|ccc|ccc}
    \toprule
    \multirow{2}[4]{*}{Metric} & \multicolumn{3}{c|}{ROSE} & \multicolumn{3}{c|}{iTransformer} & \multicolumn{3}{c|}{PatchTST} & \multicolumn{3}{c|}{Timesnet} & \multicolumn{3}{c|}{Dlinear} & \multicolumn{3}{c|}{GPT4TS} & \multicolumn{3}{c}{S2IP-LLM} \\
\cmidrule{2-22}    \multicolumn{1}{c|}{} & SMAPE & MASE  & OWA   & SMAPE & MASE  & OWA   & SMAPE & MASE  & OWA   & SMAPE & MASE  & OWA   & SMAPE & MASE  & OWA   & SMAPE & MASE  & OWA   & SMAPE & MASE  & OWA \\
    \midrule
    Yearly & \textcolor[rgb]{ .31,  .506,  .741}{\underline{13.302 }} & 3.014 & 0.833 & \textcolor[rgb]{ 1,  0,  0}{\textbf{13.238}} & \textcolor[rgb]{ 1,  0,  0}{\textbf{2.952}} & 0.823 & 16.766 & 4.331 & 1.018 & 13.387 & \textcolor[rgb]{ .31,  .506,  .741}{\underline{2.996}} & \textcolor[rgb]{ 1,  0,  0}{\textbf{0.786}} & 16.965 & 4.283 & 1.058 & 13.531 & 3.015 & 0.793 & 13.413 & 3.024 & \textcolor[rgb]{ .31,  .506,  .741}{\underline{0.792}} \\
    Quarterly & \textcolor[rgb]{ 1,  0,  0}{\textbf{9.998}} & \textcolor[rgb]{ 1,  0,  0}{\textbf{1.165}} & \textcolor[rgb]{ 1,  0,  0}{\textbf{0.885}} & \textcolor[rgb]{ .31,  .506,  .741}{\underline{10.001}} & 1.278 & 0.949 & 12.132 & 1.513 & 0.966 & 10.100  & \textcolor[rgb]{ .31,  .506,  .741}{\underline{1.182}} & \textcolor[rgb]{ .31,  .506,  .741}{\underline{0.890}} & 12.145 & 1.520  & 1.106 & 10.100  & 1.194 & 0.898 & 10.352 & 1.228 & 0.922 \\
    Monthly & \textcolor[rgb]{ 1,  0,  0}{\textbf{12.650}} & \textcolor[rgb]{ 1,  0,  0}{\textbf{0.915}} & \textcolor[rgb]{ 1,  0,  0}{\textbf{0.866}} & 13.399 & 1.031 & 0.949 & 13.428 & 0.997 & 0.948 & \textcolor[rgb]{ .31,  .506,  .741}{\underline{12.670}} & \textcolor[rgb]{ .31,  .506,  .741}{\underline{0.933}} & 0.933 & 13.514 & 1.037 & 0.956 & 12.894 & 0.956 & \textcolor[rgb]{ .31,  .506,  .741}{\underline{0.897}} & 12.995 & 0.970  & \textcolor[rgb]{ .31,  .506,  .741}{\underline{0.910}} \\
    Others & \textcolor[rgb]{ 1,  0,  0}{\textbf{4.668}} & \textcolor[rgb]{ 1,  0,  0}{\textbf{3.126}} & \textcolor[rgb]{ 1,  0,  0}{\textbf{1.020}} & 6.558 & 4.511 & 1.401 & 6.667 & 4.834 & 1.417 & 4.891 & 3.302 & 1.035 & 6.709 & 4.953 & 1.487 & 4.940  & \textcolor[rgb]{ .31,  .506,  .741}{\underline{3.228}} & \textcolor[rgb]{ .31,  .506,  .741}{\underline{1.029}} & \textcolor[rgb]{ .31,  .506,  .741}{\underline{4.805}} & \textcolor[rgb]{ .31,  .506,  .741}{\underline{3.247}} & 1.071 \\
    \bottomrule

    \end{tabular}}%
  \label{tab: Short-term forecasting}%
\end{table*}%

\clearpage
\subsection{Model Generality}
We evaluate the effectiveness of our proposed multi-frequency masking on Transformer-based models and CNN-based models, whose results are shown in Table~\ref{MFM generality}. It is notable that multi-frequency masking consistently improves these forecasting models. Specifically, it achieves average improvements of 6.3\%, 3.7\%, 1.5\% in Autoformer~\citep{autoformer}, TimesNet~\citep{timesnet}, and PatchTST~\citep{patchtst}, respectively. This indicates that multi-frequency Masking can be widely utilized across various time series forecasting models to learn generalized time series representations and improve prediction accuracy.
%%We evaluate 我们提出的Multiple Frequency Masking(MFM) to Transformer-based methods and CNN-based methods in Table 4. It is notable that MFM consistently improves these forecasting methods. 具体的， it分别在Autoforemr, TimesNet, PatchTST achieves averaged %,%,%的提升。这说明MFM能够广泛地使用在各种时序预测模型上去学习泛化的时序特征，提升模型预测的准确率。

\begin{table}[htbp]
  \centering
  \caption{Performance of multi-frequency masking.}
   \resizebox{0.7\linewidth}{!}{
    \begin{tabular}{c|c|c|c|c|c|c|c|c}
    \toprule
    Datasets & \multicolumn{2}{c|}{ETTm1} & \multicolumn{2}{c|}{ETTm2} & \multicolumn{2}{c|}{ETTh1} & \multicolumn{2}{c}{ETTh2} \\
    \midrule
    Metric & MSE   & MAE   & MSE   & MAE   & MSE   & MAE   & MSE   & MAE \\
    \midrule
    \multicolumn{1}{c|}{\multirow{2}[3]{*}{\makecell{Autoformer\\+Multi-frequency Masking}}} & 0.600 & 0.521 & 0.328 & 0.365 & 0.493 & 0.487 & 0.452 & 0.458 \\
\cmidrule{2-9}          & \textbf{0.549} & \textbf{0.488} & \textbf{0.306} & \textbf{0.349} & \textbf{0.474} & \textbf{0.478} & \textbf{0.406} & \textbf{0.425} \\
    \midrule
    \multicolumn{1}{c|}{\multirow{2}[2]{*}{\makecell{TimesNet\\+Multi-frequency Masking}}} & 0.400   & 0.406 & 0.291 & 0.333 & 0.458 & 0.450  & 0.414 & 0.427 \\
    \cmidrule{2-9}          
    & \textbf{0.386} & \textbf{0.398} & \textbf{0.282 } & \textbf{0.324 } & \textbf{0.446} & \textbf{0.438} & \textbf{0.386 } & \textbf{0.403 } \\
    \midrule
    \multicolumn{1}{c|}{\multirow{2}[2]{*}{\makecell{PatchTST\\+Multi-frequency Masking}}} & 0.353 & 0.382 & 0.256 & 0.317 & 0.413 & 0.434 & \textbf{0.331} & 0.381 \\
    \cmidrule{2-9}          
    & \textbf{0.347} & \textbf{0.372} & \textbf{0.252} & \textbf{0.308} & \textbf{0.405} & \textbf{0.424} & 0.337 & \textbf{0.379} \\
    \bottomrule
    \end{tabular}}%
  \label{MFM generality}%
\end{table}%

\subsection{Results Deviation}
\label{appendix_deviation}
We have conducted ROSE three times with different random seeds and have recorded the standard deviations for both the full-shot setting and the 10\% few-shot setting, as illustrated in Table~\ref{deviation}. As the baselines didn't report deviations in the original paper, we only reported the deviations of the PatchTST in the full-shot setting as a comparison. It can be observed that ROSE exhibits stable performance.

\begin{table}[htbp]
  \centering
  \caption{Results deviation.}
   \resizebox{\linewidth}{!}{
    \begin{tabular}{c|c|c|c|c|c|c|c}
    \toprule
    Models & \multicolumn{2}{c|}{ROSE}  & \multicolumn{2}{c|}{ROSE (10\%)} & \multicolumn{2}{c|}{PatchTST} & \multicolumn{1}{p{8.19em}}{confidence interval} \\
    \midrule
    Metric & MSE & MAE & MSE & MAE & MSE & MAE &-  \\
    \midrule
    ETTm1 & 0.342$\pm$0.003& 0.367$\pm$0.002 & 0.349$\pm$0.003& 0.372$\pm$0.002 & 0.349$\pm$ 0.004& 0.383$\pm$0.003 & 99\% \\
    ETTm2 & 0.246$\pm$0.002& 0.303$\pm$0.004 & 0.249$\pm$0.002& 0.308$\pm$0.002 & 0.255$\pm$0.002& 0.314$\pm$0.003 & 99\% \\
    ETTh1 & 0.392$\pm$0.004& 0.413$\pm$0.004 & 0.397$\pm$0.003& 0.419$\pm$0.003 & 0.411$\pm$0.003& 0.432$\pm$0.005 & 99\% \\
    ETTh2 & 0.330$\pm$0.003& 0.374$\pm$0.002 & 0.335$\pm$0.004& 0.380$\pm$0.003 & 0.348$\pm$0.004& 0.390$\pm$0.004 & 99\% \\
    Traffic & 0.391$\pm$0.008& 0.266$\pm$0.005 & 0.418$\pm$0.011& 0.278$\pm$0.006 & 0.404$\pm$0.009& 0.283$\pm$0.002 & 99\% \\
    Weather & 0.217$\pm$0.008& 0.250$\pm$0.007 & 0.224$\pm$0.007& 0.252$\pm$0.009 & 0.223$\pm$0.011& 0.263$\pm$0.014 & 99\% \\
    Electricity & 0.156$\pm$0.007& 0.249$\pm$0.009 & 0.164$\pm$0.004& 0.253$\pm$0.004 & 0.163$\pm$0.009& 0.261$\pm$0.013 & 99\% \\
    \bottomrule
    \end{tabular}}%
  \label{deviation}%
\end{table}%

\newpage
\subsection{Visualization}

\subsubsection{Visualization analysis}
To showcase the benefits of cross-domain pre-training, we performed visualizations in both the zero-shot setting and full-shot setting. 

\begin{figure}[h]
    \centering\vspace{0mm}
    \includegraphics[width=0.82\linewidth]{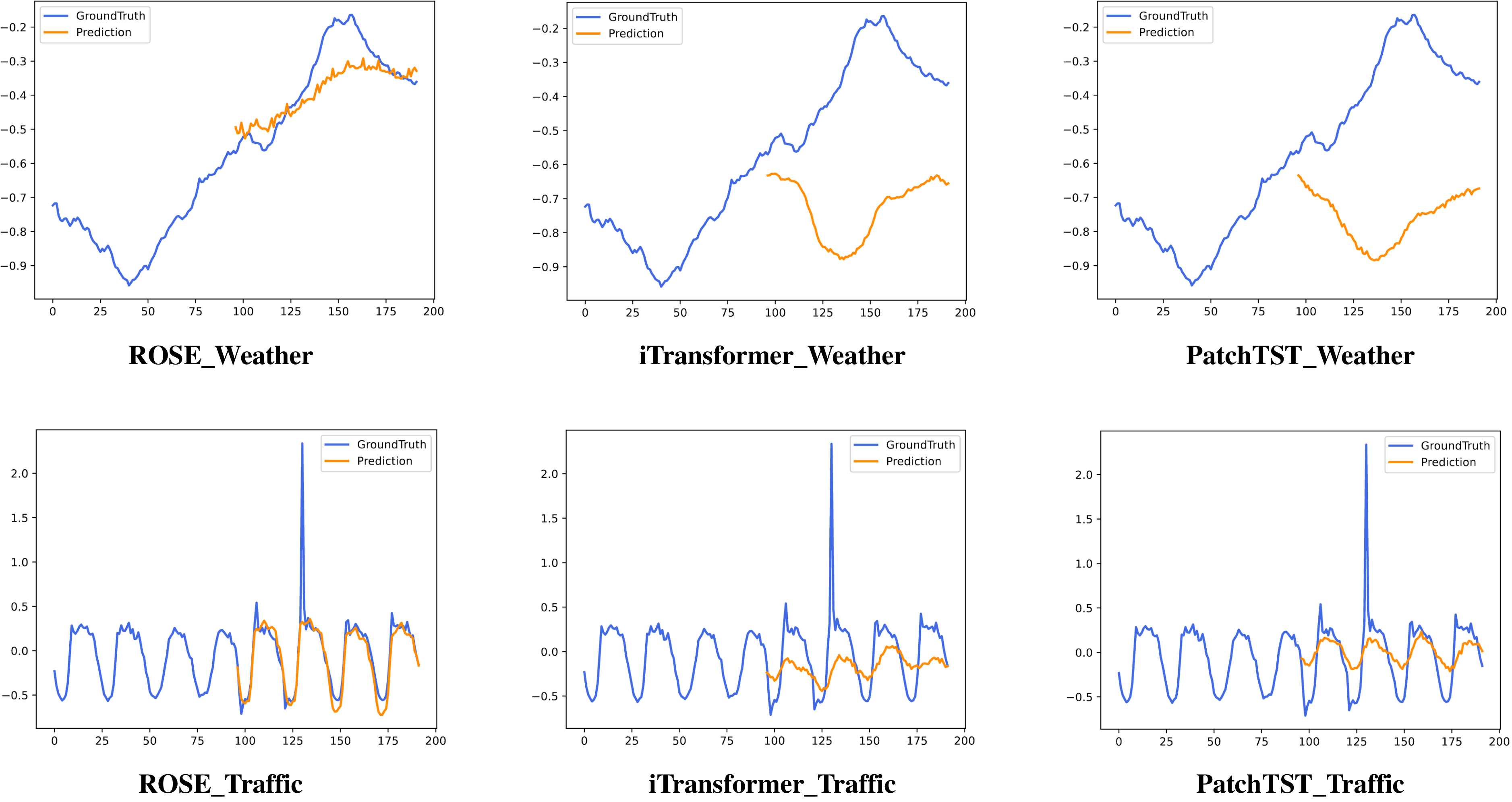}
    \caption{Visualization comparison of ROSE with cross-domain pre-training and other SOTA baselines in the zero-shot setting for three domain datasets.}
    \label{visual_analysis1}
\end{figure}

\textbf{Zero-shot}: We pre-train the baselines iTransformer and PatchTST on the energy domain dataset ETTm1 and test their zero-shot performance on two different domains (weather, traffic) . ROSE, without fine-tuning, is evaluated to the same two test-sets. As shown in the Figure~\ref{visual_analysis1}, we find that the baselines generally perform worse during domain shift due to their poor generalization. However, ROSE excels in scenarios across all domains, which demonstrates the benefits of cross-domain pre-training for improving generalization.

\begin{figure}[h]
    \centering\vspace{0mm}
    \includegraphics[width=0.82\linewidth]{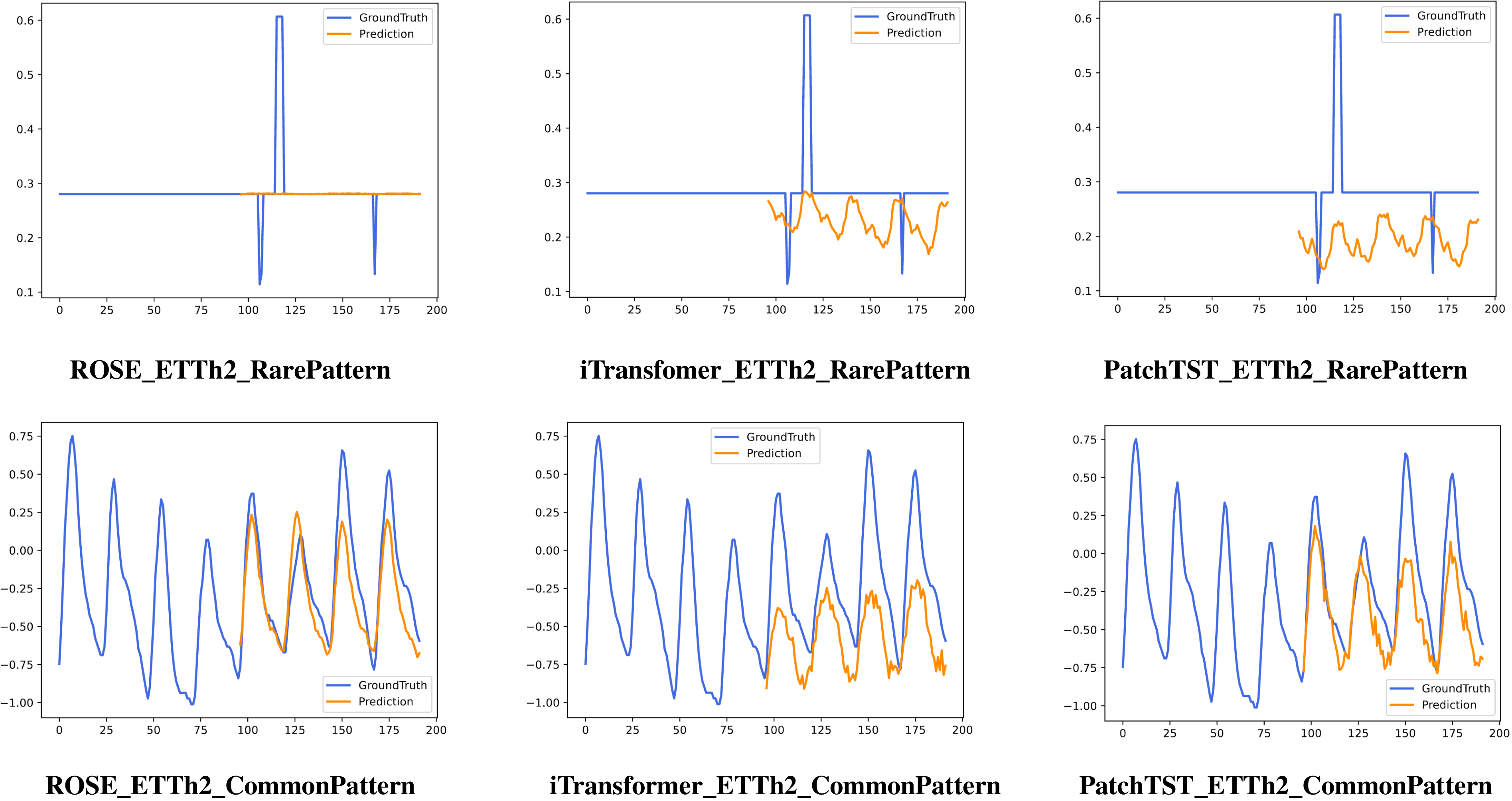}
    \caption{Visualization comparison of ROSE with cross-domain pre-training and other SOTA baselines in the full-shot setting for rare and common patterns.}
    \label{visual_analysis2}
\end{figure}

\textbf{Full-shot}: We train the baselines on the train-set of downstream dataset ETTh2 and fine-tune ROSE on the same train-set. As shown in the Figure~\ref{visual_analysis2}, We find that the baselines is limited by data diversity, leading to poor performance on patterns which rarely appear. However, ROSE excels in these cases, as the cross-domain pre-training allows ROSE to learn diverse temporal patterns, and helps ROSE to predict the patterns which rarely appear in the downstream train-set well.

\subsubsection{Visualization showcase}
\begin{figure}[h]    
    \centerline{\includegraphics[width=0.9\linewidth]{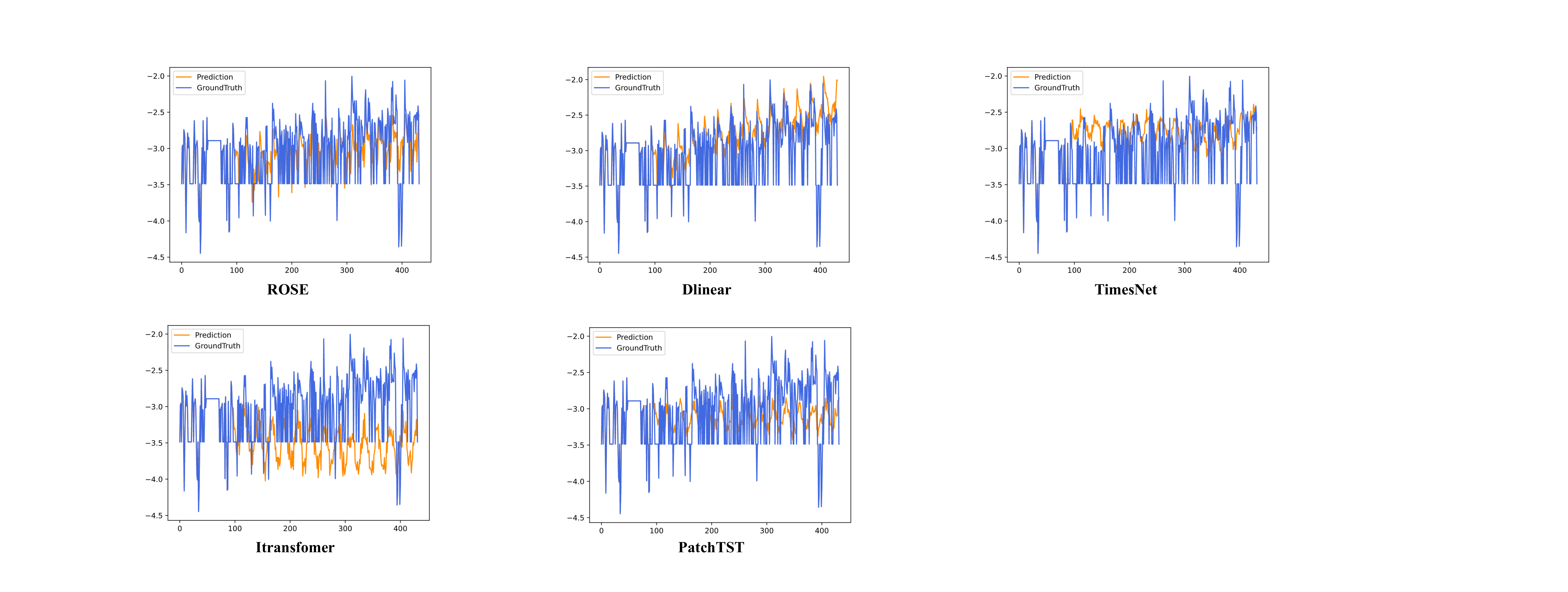}}\vspace{-2mm}
    \caption{Visualization of input-512 and predict-336 forecasting results on the ETTh2 dataset in full-shot setting.}
    \label{fig: appendix_visual1}
\end{figure}

\begin{figure}[h]    
    \centerline{\includegraphics[width=0.9\linewidth]{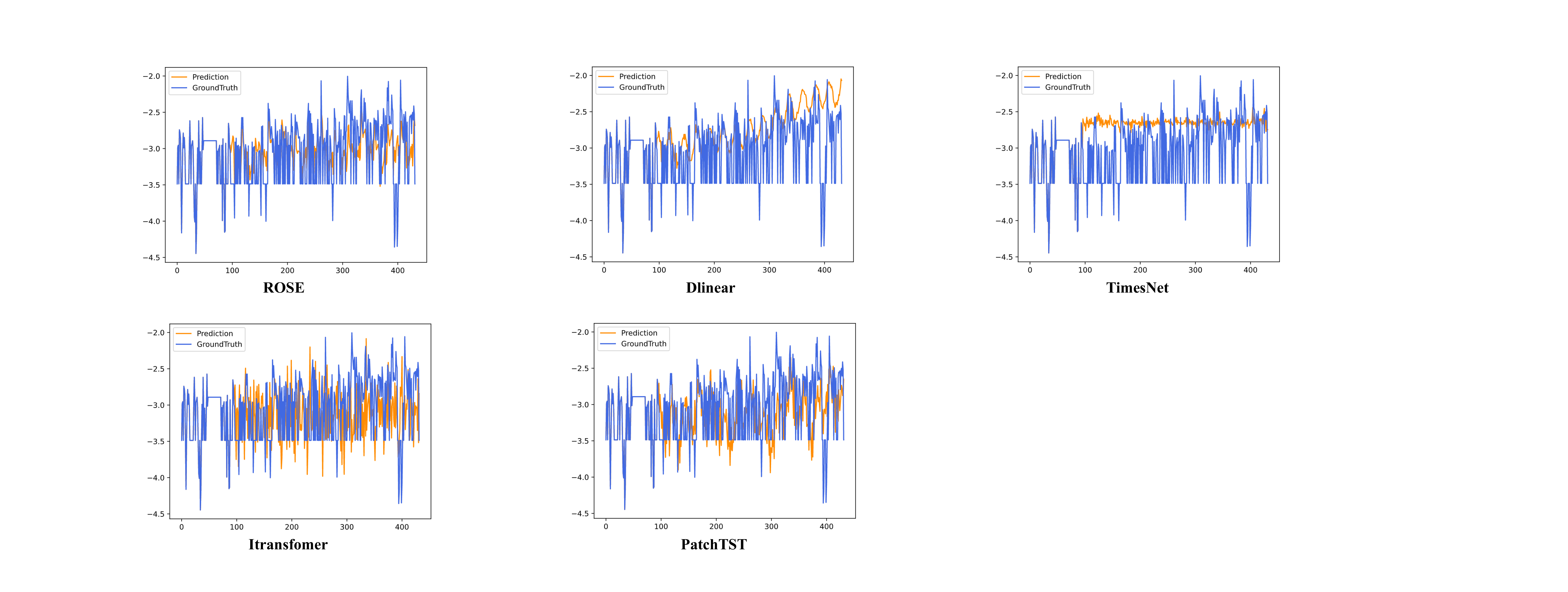}}\vspace{-2mm}
    \caption{Visualization of input-512 and predict-336 forecasting results on the ETTh2 dataset in 10\% few-shot setting.}
    \label{fig: appendix_visual2}
\end{figure}

\begin{figure}[h]    
    \centerline{\includegraphics[width=0.9\linewidth]{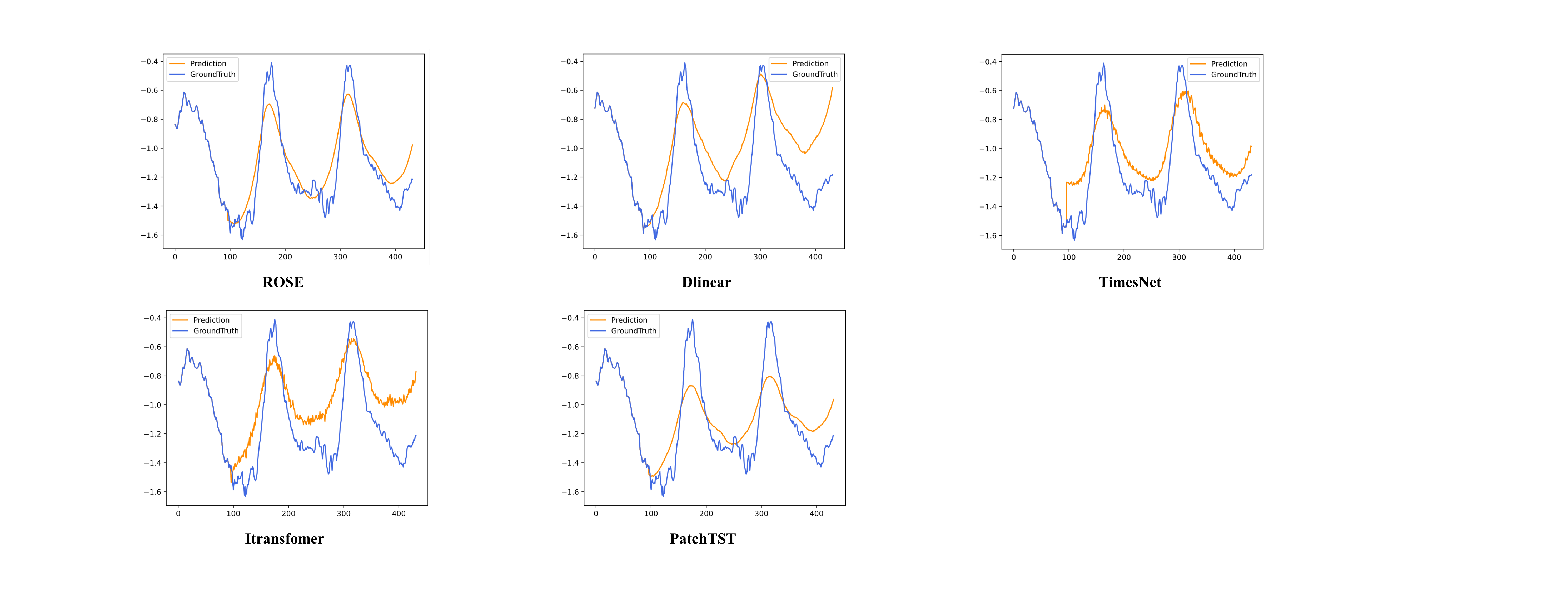}}\vspace{-2mm}
    \caption{Visualization of input-512 and predict-336 forecasting results on the weather dataset in full-shot setting.}
    \label{fig: appendix_visual3}
\end{figure}

\begin{figure}[h]    
    \centerline{\includegraphics[width=0.9\linewidth]{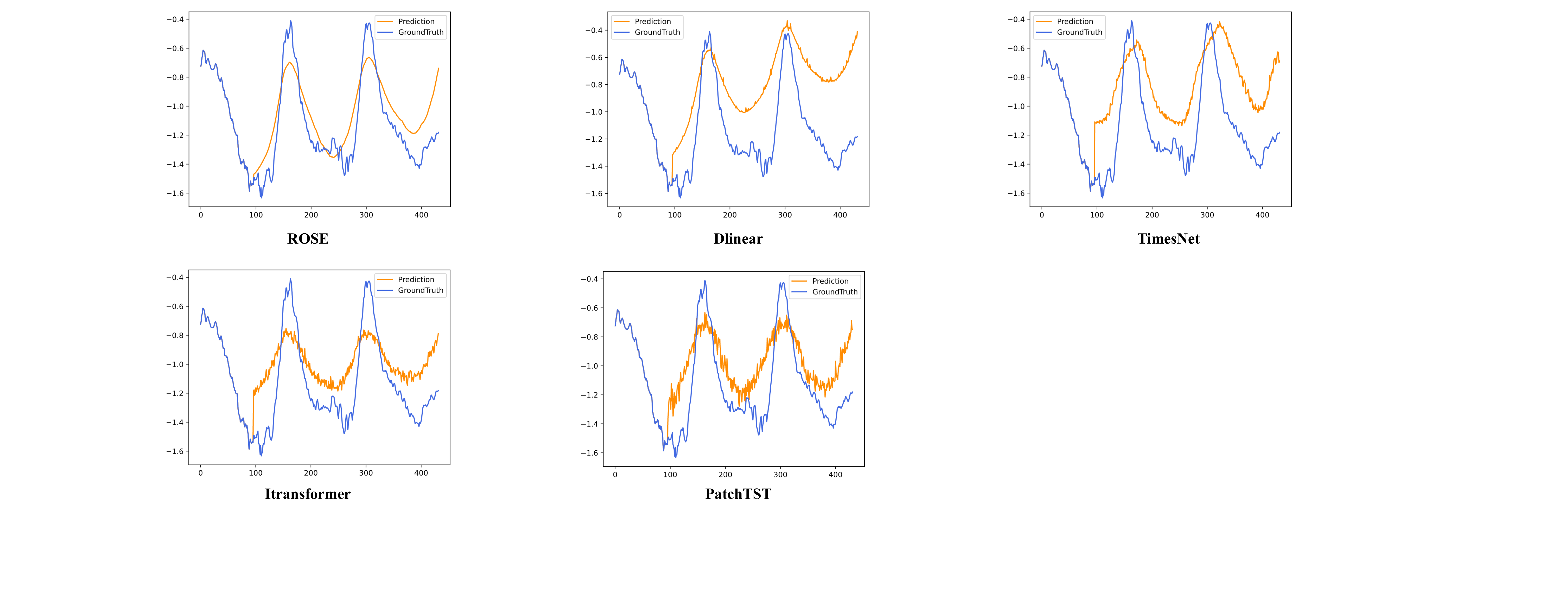}}\vspace{-2mm}
    \caption{Visualization of input-512 and predict-336 forecasting results on the weather dataset in 10\% few-shot setting.}
    \label{fig: appendix_visual4}
\end{figure}

To provide a distinct comparison among different models, we present visualizations of the forecasting results on the ETTh2 dataset and the weather dataset in different settings, as shown in Figures~\ref{fig: appendix_visual1} to Figures~\ref{fig: appendix_visual4}, given by the following models: DLinear~\citep{dlinear}, TimesNet~\citep{timesnet}, iTransfomrer~\citep{itransformer}, and PatchTST~\citep{patchtst}. Among the methods, ROSE demonstrates the most accurate prediction ability.

\newpage

% \clearpage
% \subsection{Limitations}
% \label{limiations}
% Our pre-training datasets still have room for expansion. While ROSE demonstrates strong performance, its effectiveness could be enhanced with a larger and more diverse set of pre-training datasets. By collecting additional real-world scenario data for pre-training, we aim to further improve the robustness and generalizability of ROSE across various domains.

% \subsection{Broader Impacts}
% \label{impacts}
% This paper presents ROSE, a novel general time series forecasting model that demonstrates promising performance. ROSE performs well in few-shot scenarios and shows notable zero-shot capabilities, highlighting its scalability and efficiency. We believe it can be a valuable addition to the pre-training research community. To support future research, we will also release the codebase for time-series pre-training.

% This paper only focuses on the algorithm design. Using all the codes and datasets strictly follows the corresponding licenses~\ref{pretraining_datasets}. There is no potential ethical risk or negative social impact.

\clearpage
\subsection{Full Results}
\subsubsection{Full-shot results}
\label{fullshot}
Table~\ref{tab:fullshot_full_results} shows the full results of ROSE in full-shot setting for four prediction lengths. ROSE shows the advantage over the  specific models and LLM-based models trained with the full training set.
\begin{table*}[htbp]
  \centering
  \caption{Full results in full-shot setting.}
    \resizebox{0.95\linewidth}{!}{
    \begin{tabular}{c|c|c|c|c|c|c|c|c|c|c|c|c|c|c|c}
    \toprule
    \multicolumn{2}{c|}{Models} & \multicolumn{2}{c|}{ReadyTS} & \multicolumn{2}{c|}{ITransformer} & \multicolumn{2}{c|}{PatchTST} & \multicolumn{2}{c|}{Timesnet} & \multicolumn{2}{c|}{Dlinear} & \multicolumn{2}{c|}{GPT4TS} & \multicolumn{2}{c}{$\textbf{S}^2$IP-LLM} \\
    \midrule
    \multicolumn{2}{c|}{Metric} & MSE   & MAE   & MSE   & MAE   & MSE   & MAE   & MSE   & MAE   & MSE   & MAE   & MSE   & MAE   & MSE   & MAE \\
    \midrule
\multirow{5}[10]{*}{ETTh1} & 96    & \textcolor[rgb]{ 1,  0,  0}{\textbf{0.354 }} & \textcolor[rgb]{ 1,  0,  0}{\textbf{0.385 }} & 0.386  & 0.405  & 0.370  & 0.400  & 0.470  & 0.470  & 0.367  & \textcolor[rgb]{ .161,  .447,  .957}{\underline{0.396 }} & 0.376  & 0.397  & \textcolor[rgb]{ .161,  .447,  .957}{\underline{0.366 }} & \textcolor[rgb]{ .161,  .447,  .957}{\underline{0.396 }} \\
\cmidrule{2-16}          & 192   & \textcolor[rgb]{ 1,  0,  0}{\textbf{0.389 }} & \textcolor[rgb]{ 1,  0,  0}{\textbf{0.407 }} & 0.424  & 0.440  & 0.413  & 0.429  & 0.568  & 0.523  & \textcolor[rgb]{ .161,  .447,  .957}{\underline{0.400 }} & \textcolor[rgb]{ .161,  .447,  .957}{\underline{0.417 }} & 0.416  & 0.418  & 0.401  & 0.420  \\
\cmidrule{2-16}          & 336   & \textcolor[rgb]{ 1,  0,  0}{\textbf{0.406 }} & \textcolor[rgb]{ 1,  0,  0}{\textbf{0.422 }} & 0.449  & 0.460  & 0.422  & 0.440  & 0.595  & 0.547  & 0.428  & 0.439  & 0.442  & 0.433  & \textcolor[rgb]{ .161,  .447,  .957}{\underline{0.412 }} & \textcolor[rgb]{ .161,  .447,  .957}{\underline{0.431 }} \\
\cmidrule{2-16}          & 720   & \textcolor[rgb]{ 1,  0,  0}{\textbf{0.413 }} & \textcolor[rgb]{ 1,  0,  0}{\textbf{0.443 }} & 0.495  & 0.487  & 0.447  & 0.468  & 0.694  & 0.591  & 0.468  & 0.491  & 0.477  & \textcolor[rgb]{ .161,  .447,  .957}{\underline{0.456 }} & \textcolor[rgb]{ .161,  .447,  .957}{\underline{0.440 }} & 0.458  \\
\cmidrule{2-16}          & avg   & \textcolor[rgb]{ 1,  0,  0}{\textbf{0.391 }} & \textcolor[rgb]{ 1,  0,  0}{\textbf{0.414 }} & 0.439  & 0.448  & 0.413  & 0.434  & 0.582  & 0.533  & 0.416  & 0.436  & 0.427  & \textcolor[rgb]{ .161,  .447,  .957}{\underline{0.426 }} & \textcolor[rgb]{ .161,  .447,  .957}{\underline{0.406 }} & 0.427  \\
    \midrule
    \multirow{5}[10]{*}{ETTh2} & 96    & \textcolor[rgb]{ 1,  0,  0}{\textbf{0.265 }} & \textcolor[rgb]{ 1,  0,  0}{\textbf{0.320 }} & 0.297  & 0.348  & \textcolor[rgb]{ .161,  .447,  .957}{\underline{0.274 }} & \textcolor[rgb]{ .161,  .447,  .957}{\underline{0.337 }} & 0.351  & 0.399  & 0.302  & 0.368  & 0.285  & 0.342  & 0.278  & 0.340  \\
\cmidrule{2-16}          & 192   & \textcolor[rgb]{ .161,  .447,  .957}{\underline{0.328 }} & \textcolor[rgb]{ 1,  0,  0}{\textbf{0.369 }} & 0.371  & 0.403  & 0.341  & \textcolor[rgb]{ .161,  .447,  .957}{\underline{0.382 }} & 0.394  & 0.429  & 0.404  & 0.433  & 0.354  & 0.389  & \textcolor[rgb]{ 1,  0,  0}{\textbf{0.246 }} & 0.385  \\
\cmidrule{2-16}          & 336   & \textcolor[rgb]{ .161,  .447,  .957}{\underline{0.353 }} & \textcolor[rgb]{ .161,  .447,  .957}{\underline{0.391 }} & 0.404  & 0.428  & \textcolor[rgb]{ 1,  0,  0}{\textbf{0.329 }} & \textcolor[rgb]{ 1,  0,  0}{\textbf{0.384 }} & 0.415  & 0.443  & 0.511  & 0.498  & 0.373  & 0.407  & 0.367  & 0.406  \\
\cmidrule{2-16}          & 720   & \textcolor[rgb]{ 1,  0,  0}{\textbf{0.376 }} & \textcolor[rgb]{ 1,  0,  0}{\textbf{0.417 }} & 0.424  & 0.444  & \textcolor[rgb]{ .161,  .447,  .957}{\underline{0.379 }} & \textcolor[rgb]{ .161,  .447,  .957}{\underline{0.422 }} & 0.477  & 0.481  & 0.815  & 0.640  & 0.406  & 0.441  & 0.400  & 0.436  \\
\cmidrule{2-16}          & avg   & \textcolor[rgb]{ 1,  0,  0}{\textbf{0.331 }} & \textcolor[rgb]{ 1,  0,  0}{\textbf{0.374 }} & 0.374  & 0.406  & \textcolor[rgb]{ 1,  0,  0}{\textbf{0.331 }} & \textcolor[rgb]{ .161,  .447,  .957}{\underline{0.381 }} & 0.409  & 0.438  & 0.508  & 0.485  & 0.354  & 0.394  & \textcolor[rgb]{ .161,  .447,  .957}{\underline{0.347 }} & 0.391  \\
    \midrule
    \multirow{5}[10]{*}{ETTm1} & 96    & \textcolor[rgb]{ 1,  0,  0}{\textbf{0.275 }} & \textcolor[rgb]{ .161,  .447,  .957}{\underline{0.328 }} & 0.300  & 0.353  & 0.293  & 0.346  & 0.405  & 0.421  & 0.303  & 0.346  & 0.292  & \textcolor[rgb]{ 1,  0,  0}{\textbf{0.262 }} & \textcolor[rgb]{ .161,  .447,  .957}{\underline{0.288 }} & 0.346  \\
\cmidrule{2-16}          & 192   & \textcolor[rgb]{ .161,  .447,  .957}{\underline{0.324 }} & \textcolor[rgb]{ .161,  .447,  .957}{\underline{0.358 }} & 0.345  & 0.382  & 0.333  & 0.370  & 0.508  & 0.473  & 0.335  & 0.365  & 0.332  & \textcolor[rgb]{ 1,  0,  0}{\textbf{0.301 }} & \textcolor[rgb]{ 1,  0,  0}{\textbf{0.323 }} & 0.365  \\
\cmidrule{2-16}          & 336   & \textcolor[rgb]{ 1,  0,  0}{\textbf{0.354 }} & \textcolor[rgb]{ .161,  .447,  .957}{\underline{0.377 }} & 0.374  & 0.398  & 0.369  & 0.392  & 0.523  & 0.479  & 0.365  & 0.384  & 0.366  & \textcolor[rgb]{ 1,  0,  0}{\textbf{0.341 }} & \textcolor[rgb]{ .161,  .447,  .957}{\underline{0.359 }} & 0.390  \\
\cmidrule{2-16}          & 720   & \textcolor[rgb]{ .161,  .447,  .957}{\underline{0.411 }} & \textcolor[rgb]{ .161,  .447,  .957}{\underline{0.407 }} & 0.429  & 0.430  & 0.416  & 0.420  & 0.523  & 0.484  & 0.418  & 0.415  & 0.417  & \textcolor[rgb]{ 1,  0,  0}{\textbf{0.401 }} & \textcolor[rgb]{ 1,  0,  0}{\textbf{0.403 }} & 0.418  \\
\cmidrule{2-16}          & avg   & \textcolor[rgb]{ 1,  0,  0}{\textbf{0.341 }} & \textcolor[rgb]{ 1,  0,  0}{\textbf{0.367 }} & 0.362  & 0.391  & 0.353  & 0.382  & 0.490  & 0.464  & 0.356  & \textcolor[rgb]{ .161,  .447,  .957}{\underline{0.378 }} & 0.352  & 0.383  & \textcolor[rgb]{ .161,  .447,  .957}{\underline{0.343 }} & 0.379  \\
    \midrule
    \multirow{5}[10]{*}{ETTm2} & 96    & \textcolor[rgb]{ 1,  0,  0}{\textbf{0.157 }} & \textcolor[rgb]{ 1,  0,  0}{\textbf{0.243 }} & 0.175  & 0.266  & 0.166  & 0.256  & 0.233  & 0.305  & \textcolor[rgb]{ .161,  .447,  .957}{\underline{0.164 }} & \textcolor[rgb]{ .161,  .447,  .957}{\underline{0.255 }} & 0.173  & 0.262  & 0.165  & 0.257  \\
\cmidrule{2-16}          & 192   & \textcolor[rgb]{ 1,  0,  0}{\textbf{0.213 }} & \textcolor[rgb]{ 1,  0,  0}{\textbf{0.283 }} & 0.242  & 0.312  & 0.223  & \textcolor[rgb]{ .161,  .447,  .957}{\underline{0.296 }} & 0.265  & 0.328  & 0.224  & 0.304  & 0.229  & 0.301  & \textcolor[rgb]{ .161,  .447,  .957}{\underline{0.222 }} & 0.299  \\
\cmidrule{2-16}          & 336   & \textcolor[rgb]{ 1,  0,  0}{\textbf{0.266 }} & \textcolor[rgb]{ 1,  0,  0}{\textbf{0.319 }} & 0.282  & 0.340  & \textcolor[rgb]{ .161,  .447,  .957}{\underline{0.274 }} & \textcolor[rgb]{ .161,  .447,  .957}{\underline{0.329 }} & 0.379  & 0.392  & 0.277  & 0.339  & 0.286  & 0.341  & 0.277  & 0.330  \\
\cmidrule{2-16}          & 720   & \textcolor[rgb]{ 1,  0,  0}{\textbf{0.347 }} & \textcolor[rgb]{ 1,  0,  0}{\textbf{0.373 }} & 0.378  & 0.398  & \textcolor[rgb]{ .161,  .447,  .957}{\underline{0.362 }} & \textcolor[rgb]{ .161,  .447,  .957}{\underline{0.385 }} & 0.390  & 0.407  & 0.371  & 0.401  & 0.378  & 0.401  & 0.363  & 0.390  \\
\cmidrule{2-16}          & avg   & \textcolor[rgb]{ 1,  0,  0}{\textbf{0.246 }} & \textcolor[rgb]{ 1,  0,  0}{\textbf{0.305 }} & 0.269  & 0.329  & \textcolor[rgb]{ .161,  .447,  .957}{\underline{0.256 }} & \textcolor[rgb]{ .161,  .447,  .957}{\underline{0.317 }} & 0.317  & 0.358  & 0.259  & 0.325  & 0.266  & 0.326  & 0.257  & 0.319  \\
    \midrule
    \multirow{5}[10]{*}{Weather} & 96    & \textcolor[rgb]{ 1,  0,  0}{\textbf{0.145 }} & \textcolor[rgb]{ 1,  0,  0}{\textbf{0.182 }} & 0.159  & 0.208  & \textcolor[rgb]{ .161,  .447,  .957}{\underline{0.149 }} & 0.198  & 0.193  & 0.244  & 0.170  & 0.230  & 0.162  & 0.212  & \textcolor[rgb]{ 1,  0,  0}{\textbf{0.145 }} & \textcolor[rgb]{ .161,  .447,  .957}{\underline{0.195 }} \\
\cmidrule{2-16}          & 192   & \textcolor[rgb]{ 1,  0,  0}{\textbf{0.183 }} & \textcolor[rgb]{ 1,  0,  0}{\textbf{0.226 }} & 0.200  & 0.248  & 0.194  & 0.241  & 0.320  & 0.329  & 0.212  & 0.267  & 0.204  & 0.248  & \textcolor[rgb]{ .161,  .447,  .957}{\underline{0.190 }} & \textcolor[rgb]{ .161,  .447,  .957}{\underline{0.235 }} \\
\cmidrule{2-16}          & 336   & \textcolor[rgb]{ 1,  0,  0}{\textbf{0.232 }} & \textcolor[rgb]{ 1,  0,  0}{\textbf{0.267 }} & 0.253  & 0.289  & 0.245  & 0.282  & 0.363  & 0.366  & 0.257  & 0.305  & 0.254  & 0.286  & \textcolor[rgb]{ .161,  .447,  .957}{\underline{0.243 }} & \textcolor[rgb]{ .161,  .447,  .957}{\underline{0.280 }} \\
\cmidrule{2-16}          & 720   & \textcolor[rgb]{ 1,  0,  0}{\textbf{0.309 }} & \textcolor[rgb]{ .161,  .447,  .957}{\underline{0.327 }} & 0.321  & 0.338  & 0.314  & 0.334  & 0.440  & 0.404  & 0.318  & 0.356  & 0.326  & 0.337  & \textcolor[rgb]{ .161,  .447,  .957}{\underline{0.312 }} & \textcolor[rgb]{ 1,  0,  0}{\textbf{0.326 }} \\
\cmidrule{2-16}          & avg   & \textcolor[rgb]{ 1,  0,  0}{\textbf{0.217 }} & \textcolor[rgb]{ 1,  0,  0}{\textbf{0.251 }} & 0.233  & 0.271  & 0.226  & 0.264  & 0.329  & 0.336  & 0.239  & 0.289  & 0.237  & 0.270  & \textcolor[rgb]{ .161,  .447,  .957}{\underline{0.222 }} & \textcolor[rgb]{ .161,  .447,  .957}{\underline{0.259 }} \\
    \midrule
    \multirow{5}[10]{*}{Electricity} & 96    & \textcolor[rgb]{ 1,  0,  0}{\textbf{0.125 }} & \textcolor[rgb]{ 1,  0,  0}{\textbf{0.220 }} & 0.138  & 0.237  & \textcolor[rgb]{ .161,  .447,  .957}{\underline{0.129 }} & \textcolor[rgb]{ .161,  .447,  .957}{\underline{0.222 }} & 0.182  & 0.287  & 0.141  & 0.241  & 0.139  & 0.238  & 0.135  & 0.230  \\
\cmidrule{2-16}          & 192   & \textcolor[rgb]{ 1,  0,  0}{\textbf{0.142 }} & \textcolor[rgb]{ 1,  0,  0}{\textbf{0.235 }} & 0.157  & 0.256  & \textcolor[rgb]{ .161,  .447,  .957}{\underline{0.147 }} & \textcolor[rgb]{ .161,  .447,  .957}{\underline{0.240 }} & 0.193  & 0.293  & 0.154  & 0.254  & 0.153  & 0.251  & 0.149  & 0.247  \\
\cmidrule{2-16}          & 336   & \textcolor[rgb]{ 1,  0,  0}{\textbf{0.162 }} & \textcolor[rgb]{ 1,  0,  0}{\textbf{0.252 }} & 0.167  & 0.264  & \textcolor[rgb]{ .161,  .447,  .957}{\underline{0.163 }} & \textcolor[rgb]{ .161,  .447,  .957}{\underline{0.259 }} & 0.196  & 0.298  & 0.168  & 0.271  & 0.169  & 0.266  & 0.167  & 0.266  \\
\cmidrule{2-16}          & 720   & \textcolor[rgb]{ 1,  0,  0}{\textbf{0.191 }} & \textcolor[rgb]{ 1,  0,  0}{\textbf{0.284 }} & \textcolor[rgb]{ .161,  .447,  .957}{\underline{0.194 }} & \textcolor[rgb]{ .161,  .447,  .957}{\underline{0.286 }} & 0.197  & 0.290  & 0.209  & 0.307  & 0.203  & 0.303  & 0.206  & 0.297  & 0.200  & 0.287  \\
\cmidrule{2-16}          & avg   & \textcolor[rgb]{ 1,  0,  0}{\textbf{0.155 }} & \textcolor[rgb]{ 1,  0,  0}{\textbf{0.248 }} & 0.164  & 0.261  & \textcolor[rgb]{ .161,  .447,  .957}{\underline{0.159 }} & \textcolor[rgb]{ .161,  .447,  .957}{\underline{0.253 }} & 0.195  & 0.296  & 0.166  & 0.267  & 0.167  & 0.263  & 0.161  & 0.257  \\
    \midrule
    \multirow{5}[10]{*}{Traffic} & 96    & \textcolor[rgb]{ 1,  0,  0}{\textbf{0.354 }} & \textcolor[rgb]{ .161,  .447,  .957}{\underline{0.252 }} & 0.363  & 0.265  & \textcolor[rgb]{ .161,  .447,  .957}{\underline{0.360 }} & \textcolor[rgb]{ 1,  0,  0}{\textbf{0.249 }} & 0.611  & 0.323  & 0.411  & 0.294  & 0.388  & 0.282  & 0.379  & 0.274  \\
\cmidrule{2-16}          & 192   & \textcolor[rgb]{ 1,  0,  0}{\textbf{0.377 }} & \textcolor[rgb]{ .161,  .447,  .957}{\underline{0.257 }} & 0.385  & 0.273  & \textcolor[rgb]{ .161,  .447,  .957}{\underline{0.379 }} & \textcolor[rgb]{ 1,  0,  0}{\textbf{0.256 }} & 0.609  & 0.327  & 0.421  & 0.298  & 0.407  & 0.290  & 0.397  & 0.282  \\
\cmidrule{2-16}          & 336   & \textcolor[rgb]{ .161,  .447,  .957}{\underline{0.396 }} & \textcolor[rgb]{ 1,  0,  0}{\textbf{0.262 }} & \textcolor[rgb]{ .161,  .447,  .957}{\underline{0.396 }} & 0.277  & \textcolor[rgb]{ 1,  0,  0}{\textbf{0.392 }} & \textcolor[rgb]{ .161,  .447,  .957}{\underline{0.264 }} & 0.616  & 0.335  & 0.431  & 0.304  & 0.412  & 0.294  & 0.407  & 0.289  \\
\cmidrule{2-16}          & 720   & \textcolor[rgb]{ .161,  .447,  .957}{\underline{0.434 }} & \textcolor[rgb]{ 1,  0,  0}{\textbf{0.283 }} & 0.445  & 0.312  & \textcolor[rgb]{ 1,  0,  0}{\textbf{0.432 }} & \textcolor[rgb]{ .161,  .447,  .957}{\underline{0.286 }} & 0.656  & 0.349  & 0.468  & 0.325 & 0.450  & 0.312  & 0.440  & 0.301  \\
\cmidrule{2-16}          & avg   & \textcolor[rgb]{ 1,  0,  0}{\textbf{0.390 }} & \textcolor[rgb]{ 1,  0,  0}{\textbf{0.264 }} & 0.397  & \textcolor[rgb]{ .161,  .447,  .957}{\underline{0.282 }} & \textcolor[rgb]{ .161,  .447,  .957}{\underline{0.391 }} & \textcolor[rgb]{ 1,  0,  0}{\textbf{0.264 }} & 0.623  & 0.333  & 0.433  & 0.305  & 0.414  & 0.294  & 0.405  & 0.286  \\
    
    \bottomrule

    \end{tabular}}%
  \label{tab:fullshot_full_results}%
\end{table*}%

\newpage
\subsubsection{Few-shot results}
\label{fewshot}
Table~\ref{tab:fewshot_few_results} shows the full results of ROSE in 10\% few-shot setting for four prediction lengths. ROSE shows the advantage over the  specific models and LLM-based models trained with the 10\% training set.
% Table generated by Excel2LaTeX from sheet 'fewshot详细'
\begin{table*}[htbp]
  \centering
  \caption{Full results in 10\% few-shot setting}
    \resizebox{0.95\linewidth}{!}{
    \begin{tabular}{c|c|c|c|c|c|c|c|c|c|c|c|c|c|c|c}
    \toprule
    \multicolumn{2}{c|}{Models} & \multicolumn{2}{c|}{ReadyTS} & \multicolumn{2}{c|}{ITransformer} & \multicolumn{2}{c|}{PatchTST} & \multicolumn{2}{c|}{Timesnet} & \multicolumn{2}{c|}{Dlinear} & \multicolumn{2}{c|}{GPT4TS} & \multicolumn{2}{c}{$\textbf{S}^2$IP-LLM} \\
    \midrule
    \multicolumn{2}{c|}{Metric} & MSE   & MAE   & MSE   & MAE   & MSE   & MAE   & MSE   & MAE   & MSE   & MAE   & MSE   & MAE   & MSE   & MAE \\
    \midrule
        \multirow{5}[10]{*}{ETTh1} & 96    & \textcolor[rgb]{ 1,  0,  0}{\textbf{0.367 }} & \textcolor[rgb]{ 1,  0,  0}{\textbf{0.395 }} & \textcolor[rgb]{ .161,  .447,  .957}{\underline{0.442 }} & 0.464  & 0.458  & 0.463  & 0.579  & 0.522  & 1.355  & 0.816  & 0.458  & \textcolor[rgb]{ .161,  .447,  .957}{\underline{0.456 }} & 0.481  & 0.474  \\
\cmidrule{2-16}          & 192   & \textcolor[rgb]{ 1,  0,  0}{\textbf{0.399 }} & \textcolor[rgb]{ 1,  0,  0}{\textbf{0.416 }} & \textcolor[rgb]{ .161,  .447,  .957}{\underline{0.476 }} & \textcolor[rgb]{ .161,  .447,  .957}{\underline{0.475 }} & 0.481  & 0.490  & 0.641  & 0.553  & 1.210  & 0.825  & 0.570  & 0.516  & 0.518  & 0.491  \\
\cmidrule{2-16}          & 336   & \textcolor[rgb]{ 1,  0,  0}{\textbf{0.405 }} & \textcolor[rgb]{ 1,  0,  0}{\textbf{0.423 }} & 0.486  & 0.482  & \textcolor[rgb]{ .161,  .447,  .957}{\underline{0.465 }} & \textcolor[rgb]{ .161,  .447,  .957}{\underline{0.475 }} & 0.721  & 0.582  & 1.487  & 0.914  & 0.608  & 0.535  & 0.664  & 0.570  \\
\cmidrule{2-16}          & 720   & \textcolor[rgb]{ 1,  0,  0}{\textbf{0.416 }} & \textcolor[rgb]{ 1,  0,  0}{\textbf{0.443 }} & 0.509  & 0.506  & \textcolor[rgb]{ .161,  .447,  .957}{\underline{0.478 }} & \textcolor[rgb]{ .161,  .447,  .957}{\underline{0.492 }} & 0.630  & 0.574  & 1.369  & 0.826  & 0.725  & 0.591  & 0.711  & 0.584  \\
\cmidrule{2-16}          & avg   & \textcolor[rgb]{ 1,  0,  0}{\textbf{0.397 }} & \textcolor[rgb]{ 1,  0,  0}{\textbf{0.419 }} & 0.478  & 0.482  & \textcolor[rgb]{ .161,  .447,  .957}{\underline{0.470 }} & \textcolor[rgb]{ .161,  .447,  .957}{\underline{0.480 }} & 0.643  & 0.558  & 1.355  & 0.845  & 0.590  & 0.525  & 0.593  & 0.529  \\
    \midrule
    \multirow{5}[10]{*}{ETTh2} & 96    & \textcolor[rgb]{ 1,  0,  0}{\textbf{0.273 }} & \textcolor[rgb]{ 1,  0,  0}{\textbf{0.332 }} & 0.333  & 0.385  & 0.350  & 0.389  & 0.378  & 0.413  & 1.628  & 0.724  & \textcolor[rgb]{ .161,  .447,  .957}{\underline{0.331 }} & \textcolor[rgb]{ .161,  .447,  .957}{\underline{0.374 }} & 0.354  & 0.400  \\
\cmidrule{2-16}          & 192   & \textcolor[rgb]{ 1,  0,  0}{\textbf{0.334 }} & \textcolor[rgb]{ 1,  0,  0}{\textbf{0.376 }} & 0.402  & 0.428  & 0.416  & 0.426  & 0.463  & 0.460  & 1.388  & 0.713  & 0.402  & \textcolor[rgb]{ .161,  .447,  .957}{\underline{0.411 }} & \textcolor[rgb]{ .161,  .447,  .957}{\underline{0.400 }} & 0.423  \\
\cmidrule{2-16}          & 336   & \textcolor[rgb]{ 1,  0,  0}{\textbf{0.358 }} & \textcolor[rgb]{ 1,  0,  0}{\textbf{0.397 }} & 0.438  & 0.452  & \textcolor[rgb]{ .161,  .447,  .957}{\underline{0.401 }} & \textcolor[rgb]{ .161,  .447,  .957}{\underline{0.429 }} & 0.507  & 0.495  & 1.595  & 0.772  & 0.406  & 0.433  & 0.442  & 0.450  \\
\cmidrule{2-16}          & 720   & \textcolor[rgb]{ 1,  0,  0}{\textbf{0.376 }} & \textcolor[rgb]{ 1,  0,  0}{\textbf{0.417 }} & 0.466  & 0.477  & \textcolor[rgb]{ .161,  .447,  .957}{\underline{0.436 }} & \textcolor[rgb]{ .161,  .447,  .957}{\underline{0.457 }} & 0.516  & 0.501  & 1.664  & 0.857  & 0.449  & 0.464  & 0.480  & 0.486  \\
\cmidrule{2-16}          & avg   & \textcolor[rgb]{ 1,  0,  0}{\textbf{0.335 }} & \textcolor[rgb]{ 1,  0,  0}{\textbf{0.380 }} & 0.410  & 0.436  & 0.401  & 0.425  & 0.466  & 0.467  & 1.569  & 0.766  & \textcolor[rgb]{ .161,  .447,  .957}{\underline{0.397 }} & \textcolor[rgb]{ .161,  .447,  .957}{\underline{0.421 }} & 0.419  & 0.439  \\
    \midrule
    \multirow{5}[10]{*}{ETTm1} & 96    & \textcolor[rgb]{ 1,  0,  0}{\textbf{0.287 }} & \textcolor[rgb]{ 1,  0,  0}{\textbf{0.336 }} & 0.353  & 0.392  & \textcolor[rgb]{ .161,  .447,  .957}{\underline{0.317 }} & \textcolor[rgb]{ .161,  .447,  .957}{\underline{0.363 }} & 0.481  & 0.446  & 0.454  & 0.475  & 0.390  & 0.404  & 0.388  & 0.401  \\
\cmidrule{2-16}          & 192   & \textcolor[rgb]{ 1,  0,  0}{\textbf{0.331 }} & \textcolor[rgb]{ 1,  0,  0}{\textbf{0.362 }} & 0.385  & 0.410  & \textcolor[rgb]{ .161,  .447,  .957}{\underline{0.351 }} & \textcolor[rgb]{ .161,  .447,  .957}{\underline{0.382 }} & 0.621  & 0.491  & 0.575  & 0.548  & 0.429  & 0.423  & 0.422  & 0.421  \\
\cmidrule{2-16}          & 336   & \textcolor[rgb]{ 1,  0,  0}{\textbf{0.362 }} & \textcolor[rgb]{ 1,  0,  0}{\textbf{0.379 }} & 0.422  & 0.432  & \textcolor[rgb]{ .161,  .447,  .957}{\underline{0.376 }} & \textcolor[rgb]{ .161,  .447,  .957}{\underline{0.398 }} & 0.521  & 0.479  & 0.773  & 0.631  & 0.469  & 0.439  & 0.456  & 0.430  \\
\cmidrule{2-16}          & 720   & \textcolor[rgb]{ 1,  0,  0}{\textbf{0.416 }} & \textcolor[rgb]{ 1,  0,  0}{\textbf{0.412 }} & 0.494  & 0.472  & \textcolor[rgb]{ .161,  .447,  .957}{\underline{0.435 }} & \textcolor[rgb]{ .161,  .447,  .957}{\underline{0.430 }} & 0.571  & 0.508  & 0.943  & 0.716  & 0.569  & 0.498  & 0.554  & 0.490  \\
\cmidrule{2-16}          & avg   & \textcolor[rgb]{ 1,  0,  0}{\textbf{0.349 }} & \textcolor[rgb]{ 1,  0,  0}{\textbf{0.372 }} & 0.414  & 0.426  & \textcolor[rgb]{ .161,  .447,  .957}{\underline{0.370 }} & \textcolor[rgb]{ .161,  .447,  .957}{\underline{0.393 }} & 0.549  & 0.481  & 0.686  & 0.593  & 0.464  & 0.441  & 0.455  & 0.435  \\
    \midrule
    \multirow{5}[10]{*}{ETTm2} & 96    & \textcolor[rgb]{ 1,  0,  0}{\textbf{0.159 }} & \textcolor[rgb]{ 1,  0,  0}{\textbf{0.247 }} & 0.183  & 0.279  & \textcolor[rgb]{ .161,  .447,  .957}{\underline{0.170 }} & \textcolor[rgb]{ .161,  .447,  .957}{\underline{0.259 }} & 0.212  & 0.292  & 0.493  & 0.476  & 0.188  & 0.269  & 0.192  & 0.274  \\
\cmidrule{2-16}          & 192   & \textcolor[rgb]{ 1,  0,  0}{\textbf{0.217 }} & \textcolor[rgb]{ 1,  0,  0}{\textbf{0.287 }} & 0.247  & 0.320  & \textcolor[rgb]{ .161,  .447,  .957}{\underline{0.226 }} & \textcolor[rgb]{ .161,  .447,  .957}{\underline{0.297 }} & 0.297  & 0.353  & 0.923  & 0.658  & 0.251  & 0.309  & 0.246  & 0.313  \\
\cmidrule{2-16}          & 336   & \textcolor[rgb]{ 1,  0,  0}{\textbf{0.269 }} & \textcolor[rgb]{ 1,  0,  0}{\textbf{0.322 }} & 0.300  & 0.353  & \textcolor[rgb]{ .161,  .447,  .957}{\underline{0.284 }} & \textcolor[rgb]{ .161,  .447,  .957}{\underline{0.333 }} & 0.328  & 0.364  & 1.407  & 0.822  & 0.307  & 0.346  & 0.301  & 0.340  \\
\cmidrule{2-16}          & 720   & \textcolor[rgb]{ 1,  0,  0}{\textbf{0.357 }} & \textcolor[rgb]{ 1,  0,  0}{\textbf{0.377 }} & 0.385  & 0.408  & \textcolor[rgb]{ .161,  .447,  .957}{\underline{0.363 }} & \textcolor[rgb]{ .161,  .447,  .957}{\underline{0.382 }} & 0.456  & 0.440  & 1.626  & 0.905  & 0.426  & 0.417  & 0.400  & 0.403  \\
\cmidrule{2-16}          & avg   & \textcolor[rgb]{ 1,  0,  0}{\textbf{0.250 }} & \textcolor[rgb]{ 1,  0,  0}{\textbf{0.308 }} & 0.279  & 0.340  & \textcolor[rgb]{ .161,  .447,  .957}{\underline{0.261 }} & \textcolor[rgb]{ .161,  .447,  .957}{\underline{0.318 }} & 0.323  & 0.362  & 1.112  & 0.715  & 0.293  & 0.335  & 0.284  & 0.332  \\
    \midrule
    \multirow{5}[10]{*}{Weather} & 96    & \textcolor[rgb]{ 1,  0,  0}{\textbf{0.145 }} & \textcolor[rgb]{ 1,  0,  0}{\textbf{0.184 }} & 0.189  & 0.229  & 0.166  & 0.217  & 0.199  & 0.248  & 0.230  & 0.318  & 0.163  & 0.215  & \textcolor[rgb]{ .161,  .447,  .957}{\underline{0.159 }} & \textcolor[rgb]{ .161,  .447,  .957}{\underline{0.210 }} \\
\cmidrule{2-16}          & 192   & \textcolor[rgb]{ 1,  0,  0}{\textbf{0.190 }} & \textcolor[rgb]{ 1,  0,  0}{\textbf{0.227 }} & 0.239  & 0.269  & 0.211  & 0.257  & 0.249  & 0.285  & 0.357  & 0.425  & 0.210  & 0.254  & \textcolor[rgb]{ .161,  .447,  .957}{\underline{0.200 }} & \textcolor[rgb]{ .161,  .447,  .957}{\underline{0.251 }} \\
\cmidrule{2-16}          & 336   & \textcolor[rgb]{ 1,  0,  0}{\textbf{0.245 }} & \textcolor[rgb]{ 1,  0,  0}{\textbf{0.269 }} & 0.294  & 0.308  & 0.261  & 0.296  & 0.297  & 0.316  & 0.464  & 0.493  & \textcolor[rgb]{ .161,  .447,  .957}{\underline{0.256 }} & \textcolor[rgb]{ .161,  .447,  .957}{\underline{0.292 }} & 0.257  & 0.293  \\
\cmidrule{2-16}          & 720   & \textcolor[rgb]{ 1,  0,  0}{\textbf{0.317 }} & \textcolor[rgb]{ 1,  0,  0}{\textbf{0.328 }} & 0.366  & 0.356  & 0.328  & 0.342  & 0.367  & 0.361  & 0.515  & 0.532  & 0.321  & 0.339  & \textcolor[rgb]{ .161,  .447,  .957}{\underline{0.317 }} & \textcolor[rgb]{ .161,  .447,  .957}{\underline{0.335 }} \\
\cmidrule{2-16}          & avg   & \textcolor[rgb]{ 1,  0,  0}{\textbf{0.224 }} & \textcolor[rgb]{ 1,  0,  0}{\textbf{0.252 }} & 0.272  & 0.291  & 0.242  & 0.278  & 0.278  & 0.303  & 0.391  & 0.442  & 0.238  & 0.275  & \textcolor[rgb]{ .161,  .447,  .957}{\underline{0.233 }} & \textcolor[rgb]{ .161,  .447,  .957}{\underline{0.272 }} \\
    \midrule
    \multirow{5}[10]{*}{Electricity} & 96    & \textcolor[rgb]{ 1,  0,  0}{\textbf{0.135 }} & \textcolor[rgb]{ 1,  0,  0}{\textbf{0.226 }} & 0.184  & 0.276  & 0.161  & 0.256  & 0.279  & 0.359  & 0.227  & 0.334  & \textcolor[rgb]{ .161,  .447,  .957}{\underline{0.139 }} & \textcolor[rgb]{ .161,  .447,  .957}{\underline{0.237 }} & 0.143  & 0.243  \\
\cmidrule{2-16}          & 192   & \textcolor[rgb]{ 1,  0,  0}{\textbf{0.150 }} & \textcolor[rgb]{ 1,  0,  0}{\textbf{0.240 }} & 0.192  & 0.284  & 0.163  & 0.257  & 0.282  & 0.363  & 0.265  & 0.366  & \textcolor[rgb]{ .161,  .447,  .957}{\underline{0.156 }} & \textcolor[rgb]{ .161,  .447,  .957}{\underline{0.252 }} & 0.159  & 0.258  \\
\cmidrule{2-16}          & 336   & \textcolor[rgb]{ 1,  0,  0}{\textbf{0.166 }} & \textcolor[rgb]{ 1,  0,  0}{\textbf{0.258 }} & 0.216  & 0.308  & 0.173  & \textcolor[rgb]{ .161,  .447,  .957}{\underline{0.266 }} & 0.289  & 0.367  & 0.339  & 0.417  & 0.175  & 0.270  & \textcolor[rgb]{ .161,  .447,  .957}{\underline{0.170 }} & 0.269  \\
\cmidrule{2-16}          & 720   & \textcolor[rgb]{ 1,  0,  0}{\textbf{0.205 }} & \textcolor[rgb]{ 1,  0,  0}{\textbf{0.290 }} & 0.265  & 0.347  & \textcolor[rgb]{ .161,  .447,  .957}{\underline{0.221 }} & \textcolor[rgb]{ .161,  .447,  .957}{\underline{0.313 }} & 0.333  & 0.399  & 0.482  & 0.478  & 0.233  & 0.317  & 0.230  & 0.315  \\
\cmidrule{2-16}          & avg   & \textcolor[rgb]{ 1,  0,  0}{\textbf{0.164 }} & \textcolor[rgb]{ 1,  0,  0}{\textbf{0.253 }} & 0.214  & 0.304  & 0.180  & 0.273  & 0.296  & 0.372  & 0.328  & 0.399  & 0.176  & \textcolor[rgb]{ .161,  .447,  .957}{\underline{0.269 }} & \textcolor[rgb]{ .161,  .447,  .957}{\underline{0.175 }} & 0.271  \\
    \midrule
    \multirow{5}[9]{*}{Traffic} & 96    & \textcolor[rgb]{ 1,  0,  0}{\textbf{0.398 }} & \textcolor[rgb]{ 1,  0,  0}{\textbf{0.270 }} & 0.458  & 0.314  & 0.421  & 0.299  & 0.705  & 0.386  & 0.616  & 0.385  & 0.414  & 0.297  & \textcolor[rgb]{ .161,  .447,  .957}{\underline{0.403 }} & \textcolor[rgb]{ .161,  .447,  .957}{\underline{0.293 }} \\
\cmidrule{2-16}          & 192   & \textcolor[rgb]{ 1,  0,  0}{\textbf{0.405 }} & \textcolor[rgb]{ 1,  0,  0}{\textbf{0.270 }} & 0.473  & 0.319  & 0.439  & 0.313  & 0.710  & 0.393  & 0.710  & 0.480  & 0.426  & 0.301  & \textcolor[rgb]{ .161,  .447,  .957}{\underline{0.412 }} & \textcolor[rgb]{ .161,  .447,  .957}{\underline{0.295 }} \\
\cmidrule{2-16}          & 336   & \textcolor[rgb]{ 1,  0,  0}{\textbf{0.417 }} & \textcolor[rgb]{ 1,  0,  0}{\textbf{0.277 }} & 0.491  & 0.329  & 0.448  & 0.318  & 0.863  & 0.456  & 0.723  & 0.481  & 0.434  & \textcolor[rgb]{ .161,  .447,  .957}{\underline{0.303 }} & \textcolor[rgb]{ .161,  .447,  .957}{\underline{0.427 }} & 0.316  \\
\cmidrule{2-16}          & 720   & \textcolor[rgb]{ 1,  0,  0}{\textbf{0.452 }} & \textcolor[rgb]{ 1,  0,  0}{\textbf{0.294 }} & 0.536  & 0.361  & 0.478  & \textcolor[rgb]{ .161,  .447,  .957}{\underline{0.320 }} & 0.928  & 0.485  & 0.673  & 0.436  & 0.487  & 0.337  & \textcolor[rgb]{ .161,  .447,  .957}{\underline{0.469 }} & 0.325  \\
\cmidrule{2-16}          & avg   & \textcolor[rgb]{ 1,  0,  0}{\textbf{0.418 }} & \textcolor[rgb]{ 1,  0,  0}{\textbf{0.278 }} & 0.490  & 0.331  & 0.447  & 0.312  & 0.801  & 0.430  & 0.680  & 0.446  & 0.440  & 0.310  & \textcolor[rgb]{ .161,  .447,  .957}{\underline{0.427 }} & \textcolor[rgb]{ .161,  .447,  .957}{\underline{0.307 }} \\
    
        \bottomrule
    \end{tabular}}%
  \label{tab:fewshot_few_results}%
\end{table*}%

\newpage
\subsubsection{Ablation study results}
\label{Appendix_ablation}
\textbf{Novelty of decomposed frequency learning.}
% Frequency masking is not a new concept, but past approaches primarily focus on utilizing frequency masking as a way of data augmentation that aims at enhancing the diversity of input data~~\citep{fraug, freqmask2}. In contrast, the key innovation of our work is not only proposing a Multi-frequency masking method, but also the combination of frequency masking with reconstruction task as a novel pre-training framework that learns a universal and unified feature representation by separating the semantic and cross-period information of time series at different frequencies. Supported by Multi-frequency masking, the reconstruction task makes ROSE capture diversified features.
Frequency masking is not a new concept, but past approaches randomly mask frequencies of a single time series once~~\citep{fraug, freqmask2}, which show limited forecasting effectiveness due to the lack of common pattern learning from heterogeneous time series that come from various domains. While the multi-frequency masking we proposed randomly mask either high-frequency or low-frequency components of a time series multiple times as the key to enable learning of common time series patterns, such as trends and various long and short term fluctuations.
Moreover, different from utilizing frequency masking as a way of data augmentation to enhance the diversity of input data~~\citep{fraug, freqmask2}, we combine multi-frequency masking with reconstruction task as a novel pre-training framework, that learns a universal and unified feature representation by comprehending the data from various frequency perspectives, thereby enabling it to learn generalized representations.

\textbf{Difference between frequency-domain masking and time-domain noise addition.} Multi-frequency masking and reconstruction are not equivalent to the pre-training methods of adding noise and denoising (Noise). Due to the sparsity of time series, the process of adding noise and denoising may potentially disrupt the information of original time series~\citep{simmtm}. In contrast, multi-frequency masking not only preserves the series from such disruption but also helps the model understand
temporal patterns from a multi-frequency perspective, thereby helping the model to learn general features better.

\textbf{Other pre-training tasks.} Based on the two points above, we conduct experiments to compare two other pre-training tasks: 1) using frequency-domain augmentation only for data expansion without reconstruction task (\textit{Aug}); 2) replacing multi-frequency masking and reconstruction task with adding time-domian noise and denoise task (\textit{Noise}).As shown in Table~\ref{Appendix_ablation_tables}, we find that ROSE is significantly more effective than \textit{Aug} and \textit{Noise}, which demonstrates the effectiveness of multi-frequency masking and reconstruction task in learning generalized features.

\begin{table*}[htbp]
  \centering
  \caption{Full results of ablation study}
    \resizebox{0.85\linewidth}{!}{
    \begin{tabular}{c|c|c|c|c|c|c|c|c|c|c}
    \toprule
    \multicolumn{2}{c|}{\multirow{2}[4]{*}{Design}} & \multirow{2}[4]{*}{Pred\_len} & \multicolumn{2}{c|}{ETTm1} & \multicolumn{2}{c|}{ETTm2} & \multicolumn{2}{c|}{ETTh1} & \multicolumn{2}{c}{ETTh2} \\
\cmidrule{4-11}    \multicolumn{2}{c|}{} &       & MSE   & MAE   & MSE   & MAE   & MSE   & MAE   & MSE   & MAE \\
    \midrule
    \multicolumn{2}{c|}{\multirow{5}[10]{*}{ReadyTS}} & 96    & \textbf{0.287 } & \textbf{0.336 } & \textbf{0.159 } & \textbf{0.247 } & \textbf{0.367 } & \textbf{0.395 } & \textbf{0.273 } & \textbf{0.332} \\
\cmidrule{3-11}    \multicolumn{2}{c|}{} & 192   & \textbf{0.331 } & \textbf{0.362 } & \textbf{0.217 } & \textbf{0.287 } & \textbf{0.399 } & \textbf{0.416 } & \textbf{0.334 } & \textbf{0.376} \\
\cmidrule{3-11}    \multicolumn{2}{c|}{} & 336   & 0.362  & \textbf{0.379 } & \textbf{0.269 } & \textbf{0.322 } & \textbf{0.405 } & \textbf{0.423 } & \textbf{0.358 } & \textbf{0.397} \\
\cmidrule{3-11}    \multicolumn{2}{c|}{} & 720   & \textbf{0.416 } & \textbf{0.412 } & \textbf{0.357 } & \textbf{0.377 } & \textbf{0.416 } & \textbf{0.443 } & \textbf{0.376 } & \textbf{0.417} \\
\cmidrule{3-11}    \multicolumn{2}{c|}{} & avg   & \textbf{0.349 } & \textbf{0.372 } & \textbf{0.250 } & \textbf{0.308 } & \textbf{0.397 } & \textbf{0.419 } & \textbf{0.335 } & \textbf{0.380} \\
    \midrule
    \multicolumn{1}{c|}{\multirow{15}[30]{*}{\makecell[c]{Replace \\ Multi-Frequency Masking}}} & \multirow{5}[10]{*}{Random Frequency Masking} & 96    & 0.330  & 0.370  & 0.170  & 0.262  & 0.391  & 0.399  & 0.303  & 0.359  \\
\cmidrule{3-11}    \multicolumn{1}{c|}{} &       & 192   & 0.352  & 0.392  & 0.232  & 0.298  & 0.406  & 0.430  & 0.356  & 0.395  \\
\cmidrule{3-11}    \multicolumn{1}{c|}{} &       & 336   & 0.390  & 0.389  & 0.276  & 0.342  & 0.411  & 0.432  & 0.417  & 0.428  \\
\cmidrule{3-11}    \multicolumn{1}{c|}{} &       & 720   & 0.452  & 0.438  & 0.366  & 0.392  & 0.432  & 0.447  & 0.420  & 0.439  \\
\cmidrule{3-11}    \multicolumn{1}{c|}{} &       & avg   & 0.381 & 0.397 & 0.261 & 0.324 & 0.410 & 0.427 & 0.374 & \multicolumn{1}{c}{0.405} \\
\cmidrule{2-11}          & \multirow{5}[10]{*}{Multi-Patch Masking} & 96    & 0.302  & 0.348  & 0.168  & 0.257  & 0.377  & 0.408  & 0.282  & 0.343  \\
\cmidrule{3-11}          &       & 192   & 0.336  & 0.367  & 0.228  & 0.297  & 0.404  & 0.423  & 0.343  & 0.379  \\
\cmidrule{3-11}          &       & 336   & 0.364  & 0.385  & 0.277  & 0.328  & 0.405  & 0.420  & 0.374  & 0.403  \\
\cmidrule{3-11}          &       & 720   & 0.423  & 0.416  & 0.364  & 0.381  & 0.431  & 0.455  & 0.396  & 0.430  \\
\cmidrule{3-11}          &       & avg   & 0.356 & 0.379 & 0.259 & 0.316 & 0.404 & 0.426 & 0.349 & 0.389 \\
\cmidrule{2-11}          & \multirow{5}[10]{*}{Patch Masking} & 96    & 0.318  & 0.366  & 0.168  & 0.259  & 0.388  & 0.412  & 0.303  & 0.359  \\
\cmidrule{3-11}          &       & 192   & 0.355  & 0.388  & 0.228  & 0.298  & 0.402  & 0.422  & 0.370  & 0.399  \\
\cmidrule{3-11}          &       & 336   & 0.388  & 0.406  & 0.279  & 0.331  & 0.411  & 0.435  & 0.413  & 0.428  \\
\cmidrule{3-11}          &       & 720   & 0.450  & 0.438  & 0.370  & 0.388  & 0.431  & 0.459  & 0.413  & 0.443  \\
\cmidrule{3-11}          &       & avg   & 0.378 & 0.400   & 0.261 & 0.319 & 0.408 & 0.432 & 0.375 & 0.407 \\
    \midrule
    \multicolumn{1}{c|}{\multirow{10}[19]{*}{\makecell{Other \\ pre-training tasks}}} & \multirow{5}[10]{*}{\textit{Aug}} & 96    & 0.304  & 0.357  & 0.178  & 0.266  & 0.376  & 0.405  & 0.281  & 0.348  \\
\cmidrule{3-11}    \multicolumn{1}{c|}{} &       & 192   & 0.343  & 0.379  & 0.254  & 0.318  & 0.409  & 0.429  & 0.345  & 0.389  \\
\cmidrule{3-11}    \multicolumn{1}{c|}{} &       & 336   & 0.373  & 0.400  & 0.299  & 0.354  & 0.435  & 0.453  & 0.382  & 0.417  \\
\cmidrule{3-11}    \multicolumn{1}{c|}{} &       & 720   & 0.444  & 0.432  & 0.387  & 0.408  & 0.452  & 0.471  & 0.434  & 0.452  \\
\cmidrule{3-11}    \multicolumn{1}{c|}{} &       & avg   & 0.366  & 0.392  & 0.279  & 0.336  & 0.418  & 0.439  & 0.360  & 0.401  \\
\cmidrule{2-11}    \multicolumn{1}{c|}{} & \multirow{5}[9]{*}{\textit{Noise}} & 96    & 0.303  & 0.355  & 0.172  & 0.261  & 0.370  & 0.405  & 0.280  & 0.341  \\
\cmidrule{3-11}    \multicolumn{1}{c|}{} &       & 192   & 0.342  & 0.376  & 0.221  & 0.292  & 0.403  & 0.427  & 0.350  & 0.384  \\
\cmidrule{3-11}    \multicolumn{1}{c|}{} &       & 336   & 0.368  & 0.393  & 0.272  & 0.325  & 0.420  & 0.439  & 0.385  & 0.410  \\
\cmidrule{3-11}    \multicolumn{1}{c|}{} &       & 720   & 0.423  & 0.422  & 0.367  & 0.386  & 0.442  & 0.462  & 0.403  & 0.432  \\
\cmidrule{3-11}    \multicolumn{1}{c|}{} &       & avg   & 0.359  & 0.387  & 0.258  & 0.316  & 0.409  & 0.433  & 0.355  & 0.392  \\
    \midrule
    \multirow{15}[30]{*}{w/o} & \multirow{5}[10]{*}{ TS-Register} & 96    & 0.297  & 0.345  & 0.164  & 0.252  & 0.379  & 0.399  & 0.276  & 0.336  \\
\cmidrule{3-11}          &       & 192   & 0.334  & 0.367  & 0.221  & 0.290  & 0.419  & 0.420  & 0.350  & 0.380  \\
\cmidrule{3-11}          &       & 336   & \textbf{0.360 } & 0.384  & 0.275  & 0.325  & 0.438  & 0.442  & 0.393  & 0.411  \\
\cmidrule{3-11}          &       & 720   & 0.424  & 0.416  & 0.364  & 0.379  & 0.435  & 0.448  & 0.400  & 0.432  \\
\cmidrule{3-11}          &       & avg   & 0.354 & 0.378 & 0.256 & 0.312 & 0.418 & 0.427 & 0.355 & 0.390 \\
\cmidrule{2-11}          & \multirow{5}[10]{*}{Prediction Task} & 96    & 0.301  & 0.348  & 0.166  & 0.255  & 0.380  & 0.407  & 0.295  & 0.359  \\
\cmidrule{3-11}          &       & 192   & 0.343  & 0.374  & 0.221  & 0.291  & 0.410  & 0.426  & 0.372  & 0.406  \\
\cmidrule{3-11}          &       & 336   & 0.374  & 0.393  & 0.275  & 0.327  & 0.440  & 0.443  & 0.403  & 0.429  \\
\cmidrule{3-11}          &       & 720   & 0.424  & 0.420  & 0.366  & 0.384  & 0.458  & 0.476  & 0.418  & 0.446  \\
\cmidrule{3-11}          &       & avg   & 0.360  & 0.384 & 0.257 & 0.314 & 0.422 & 0.438 & 0.372 & 0.410 \\
\cmidrule{2-11}          & \multirow{5}[10]{*}{Reconstruction  Task} & 96    & 0.329  & 0.371  & 0.175  & 0.265  & 0.374  & 0.399  & 0.296  & 0.355  \\
\cmidrule{3-11}          &       & 192   & 0.363  & 0.391  & 0.233  & 0.304  & 0.407  & 0.422  & 0.354  & 0.389  \\
\cmidrule{3-11}          &       & 336   & 0.394  & 0.407  & 0.287  & 0.340  & 0.437  & 0.440  & 0.385  & 0.413  \\
\cmidrule{3-11}          &       & 720   & 0.461  & 0.442  & 0.379  & 0.396  & 0.430  & 0.453  & 0.408  & 0.438  \\
\cmidrule{3-11}          &       & avg   & 0.387 & 0.403 & 0.269 & 0.327 & 0.412 & 0.428 & 0.361 & 0.399 \\
    \midrule
    \multicolumn{2}{c|}{\multirow{5}[8]{*}{From Scratch}} & 96    & 0.301  & 0.357  & 0.171  & 0.260  & 0.419  & 0.439  & 0.315  & 0.389  \\
\cmidrule{3-11}    \multicolumn{2}{c|}{} & 192   & 0.358  & 0.385  & 0.223  & 0.294  & 0.438  & 0.457  & 0.366  & 0.405  \\
\cmidrule{3-11}    \multicolumn{2}{c|}{} & 336   & 0.390  & 0.396  & 0.282  & 0.336  & 0.484  & 0.484  & 0.424  & 0.435  \\
\cmidrule{3-11}    \multicolumn{2}{c|}{} & 720   & 0.436  & 0.427  & 0.366  & 0.380  & 0.540  & 0.538  & 0.495  & 0.473  \\
\cmidrule{3-11}    \multicolumn{2}{c|}{} & avg   & 0.371  & 0.391  & 0.261  & 0.318  & 0.470  & 0.480  & 0.400  & 0.425  \\

    \bottomrule
    \end{tabular}}%
  \label{Appendix_ablation_tables}%
\end{table*}%

% \newpage
\subsubsection{Zero-shot results}
Table~\ref{tab:zero_shot} shows the full results of ROSE and other foundation models in zero-shot setting for four prediction lengths. ROSE exhibits competitive performance.
% Table generated by Excel2LaTeX from sheet 'Sheet2'
\begin{table*}[htbp]
  \centering
  \caption{Full results in zero-shot setting.}
  \resizebox{0.95\linewidth}{!}{
    \begin{tabular}{c|c|c|c|c|c|c|c|c|c|c|c|c|c}
    \toprule
    \multicolumn{2}{c|}{Models} & \multicolumn{2}{c|}{ROSE\_512} & \multicolumn{2}{c|}{Timer} & \multicolumn{2}{c|}{MOIRAI} & \multicolumn{2}{c|}{Chronos} & \multicolumn{2}{c|}{TimesFM} & \multicolumn{2}{c}{Moment} \\
    \midrule
    \multicolumn{2}{c|}{Metric} & MSE   & MAE   & MSE   & MAE   & MSE   & MAE   & MSE   & MAE   & MSE   & MAE   & MSE   & MAE \\
    \midrule
    \multirow{5}[10]{*}{ETTh1} & 96    & \textcolor[rgb]{ 1,  0,  0}{\textbf{0.382 }} & 0.408  & 0.414  & 0.439  & \textcolor[rgb]{ .161,  .447,  .957}{\underline{0.405 }} & \textcolor[rgb]{ 1,  0,  0}{\textbf{0.397 }} & 0.494  & 0.409  & 0.432  & \textcolor[rgb]{ .161,  .447,  .957}{\underline{0.405 }} & 0.706  & 0.561  \\
\cmidrule{2-14}          & 192   & \textcolor[rgb]{ 1,  0,  0}{\textbf{0.400 }} & \textcolor[rgb]{ 1,  0,  0}{\textbf{0.420 }} & \textcolor[rgb]{ .161,  .447,  .957}{\underline{0.440 }} & 0.455  & 0.458  & \textcolor[rgb]{ .161,  .447,  .957}{\underline{0.428 }} & 0.561  & 0.443  & 0.492  & 0.438  & 0.716  & 0.579  \\
\cmidrule{2-14}          & 336   & \textcolor[rgb]{ 1,  0,  0}{\textbf{0.404 }} & \textcolor[rgb]{ 1,  0,  0}{\textbf{0.426 }} & \textcolor[rgb]{ .161,  .447,  .957}{\underline{0.455 }} & 0.463  & 0.509  & \textcolor[rgb]{ .161,  .447,  .957}{\underline{0.454 }} & 0.580  & 0.460  & 0.519  & 0.458  & 0.705  & 0.583  \\
\cmidrule{2-14}          & 720   & \textcolor[rgb]{ 1,  0,  0}{\textbf{0.420 }} & \textcolor[rgb]{ 1,  0,  0}{\textbf{0.447 }} & \textcolor[rgb]{ .161,  .447,  .957}{\underline{0.496 }} & 0.496  & 0.529  & 0.494  & 0.605  & 0.495  & 0.512  & \textcolor[rgb]{ .161,  .447,  .957}{\underline{0.477 }} & 0.705  & 0.597  \\
\cmidrule{2-14}          & avg   & \textcolor[rgb]{ 1,  0,  0}{\textbf{0.401 }} & \textcolor[rgb]{ 1,  0,  0}{\textbf{0.425 }} & \textcolor[rgb]{ .161,  .447,  .957}{\underline{0.451 }} & 0.463  & 0.475  & \textcolor[rgb]{ .161,  .447,  .957}{\underline{0.443 }} & 0.560  & 0.452  & 0.489  & 0.444  & 0.708  & 0.580  \\
    \midrule
    \multirow{5}[10]{*}{ETTh2} & 96    & \textcolor[rgb]{ 1,  0,  0}{\textbf{0.298 }} & 0.362  & 0.305  & 0.355  & \textcolor[rgb]{ .161,  .447,  .957}{\underline{0.303 }} & \textcolor[rgb]{ 1,  0,  0}{\textbf{0.338 }} & 0.306  & \textcolor[rgb]{ 1,  0,  0}{\textbf{0.338 }} & 0.311  & \textcolor[rgb]{ .161,  .447,  .957}{\underline{0.345 }} & 0.373  & 0.416  \\
\cmidrule{2-14}          & 192   & \textcolor[rgb]{ 1,  0,  0}{\textbf{0.336 }} & \textcolor[rgb]{ .161,  .447,  .957}{\underline{0.385 }} & \textcolor[rgb]{ .161,  .447,  .957}{\underline{0.365 }} & 0.406  & 0.369  & \textcolor[rgb]{ 1,  0,  0}{\textbf{0.384 }} & 0.396  & 0.394  & 0.401  & 0.397  & 0.384  & 0.422  \\
\cmidrule{2-14}          & 336   & \textcolor[rgb]{ 1,  0,  0}{\textbf{0.353 }} & \textcolor[rgb]{ 1,  0,  0}{\textbf{0.399 }} & \textcolor[rgb]{ .161,  .447,  .957}{\underline{0.378 }} & 0.413  & 0.397  & \textcolor[rgb]{ .161,  .447,  .957}{\underline{0.410 }} & 0.423  & 0.417  & 0.436  & 0.430  & 0.386  & 0.426  \\
\cmidrule{2-14}          & 720   & \textcolor[rgb]{ 1,  0,  0}{\textbf{0.395 }} & \textcolor[rgb]{ 1,  0,  0}{\textbf{0.432 }} & \textcolor[rgb]{ .161,  .447,  .957}{\underline{0.414 }} & 0.457  & 0.447  & 0.450  & 0.442  & \textcolor[rgb]{ .161,  .447,  .957}{\underline{0.439 }} & 0.437  & 0.450  & 0.425  & 0.454  \\
\cmidrule{2-14}          & avg   & \textcolor[rgb]{ 1,  0,  0}{\textbf{0.346 }} & \textcolor[rgb]{ 1,  0,  0}{\textbf{0.394 }} & \textcolor[rgb]{ .161,  .447,  .957}{\underline{0.366 }} & 0.408  & 0.379  & \textcolor[rgb]{ .161,  .447,  .957}{\underline{0.396 }} & 0.392  & 0.397  & 0.396  & 0.405  & 0.392  & 0.430  \\
    \midrule
    \multirow{5}[10]{*}{ETTm1} & 96    & 0.512  & 0.460  & \textcolor[rgb]{ .161,  .447,  .957}{\underline{0.440 }} & \textcolor[rgb]{ .161,  .447,  .957}{\underline{0.422 }} & 0.660  & 0.476  & 0.514  & 0.443  & \textcolor[rgb]{ 1,  0,  0}{\textbf{0.366 }} & \textcolor[rgb]{ 1,  0,  0}{\textbf{0.374 }} & 0.679  & 0.544  \\
\cmidrule{2-14}          & 192   & 0.512  & 0.462  & \textcolor[rgb]{ .161,  .447,  .957}{\underline{0.505 }} & \textcolor[rgb]{ .161,  .447,  .957}{\underline{0.458 }} & 0.707  & 0.500  & 0.608  & 0.475  & \textcolor[rgb]{ 1,  0,  0}{\textbf{0.413 }} & \textcolor[rgb]{ 1,  0,  0}{\textbf{0.401 }} & 0.690  & 0.550  \\
\cmidrule{2-14}          & 336   & \textcolor[rgb]{ .161,  .447,  .957}{\underline{0.523 }} & \textcolor[rgb]{ .161,  .447,  .957}{\underline{0.470 }} & 0.570  & 0.490  & 0.730  & 0.515  & 0.690  & 0.507  & \textcolor[rgb]{ 1,  0,  0}{\textbf{0.445 }} & \textcolor[rgb]{ 1,  0,  0}{\textbf{0.429 }} & 0.701  & 0.557  \\
\cmidrule{2-14}          & 720   & \textcolor[rgb]{ .161,  .447,  .957}{\underline{0.552 }} & \textcolor[rgb]{ .161,  .447,  .957}{\underline{0.490 }} & 0.659  & 0.534  & 0.758  & 0.536  & 0.733  & 0.555  & \textcolor[rgb]{ 1,  0,  0}{\textbf{0.513 }} & \textcolor[rgb]{ 1,  0,  0}{\textbf{0.470 }} & 0.719  & 0.569  \\
\cmidrule{2-14}          & avg   & \textcolor[rgb]{ .161,  .447,  .957}{\underline{0.525 }} & \textcolor[rgb]{ .161,  .447,  .957}{\underline{0.471 }} & 0.544  & 0.476  & 0.714  & 0.507  & 0.636  & 0.495  & \textcolor[rgb]{ 1,  0,  0}{\textbf{0.434 }} & \textcolor[rgb]{ 1,  0,  0}{\textbf{0.419 }} & 0.697  & 0.555  \\
    \midrule
    \multirow{5}[10]{*}{ETTm2} & 96    & 0.224  & 0.309  & 0.203  & 0.285  & 0.216  & \textcolor[rgb]{ .161,  .447,  .957}{\underline{0.282 }} & \textcolor[rgb]{ .161,  .447,  .957}{\underline{0.202 }} & 0.293  & \textcolor[rgb]{ 1,  0,  0}{\textbf{0.189 }} & \textcolor[rgb]{ 1,  0,  0}{\textbf{0.257 }} & 0.230  & 0.308  \\
\cmidrule{2-14}          & 192   & \textcolor[rgb]{ .161,  .447,  .957}{\underline{0.266 }} & 0.333  & \textcolor[rgb]{ 1,  0,  0}{\textbf{0.265 }} & \textcolor[rgb]{ .161,  .447,  .957}{\underline{0.327 }} & 0.294  & 0.330  & 0.286  & 0.348  & 0.277  & \textcolor[rgb]{ 1,  0,  0}{\textbf{0.325 }} & 0.285  & 0.338  \\
\cmidrule{2-14}          & 336   & \textcolor[rgb]{ 1,  0,  0}{\textbf{0.310 }} & \textcolor[rgb]{ 1,  0,  0}{\textbf{0.358 }} & \textcolor[rgb]{ .161,  .447,  .957}{\underline{0.319 }} & \textcolor[rgb]{ .161,  .447,  .957}{\underline{0.361 }} & 0.368  & 0.373  & 0.355  & 0.386  & 0.350  & 0.381  & 0.339  & 0.369  \\
\cmidrule{2-14}          & 720   & \textcolor[rgb]{ 1,  0,  0}{\textbf{0.395 }} & \textcolor[rgb]{ 1,  0,  0}{\textbf{0.407 }} & \textcolor[rgb]{ .161,  .447,  .957}{\underline{0.405 }} & \textcolor[rgb]{ .161,  .447,  .957}{\underline{0.410 }} & 0.494  & 0.439  & 0.409  & 0.425  & 0.464  & 0.448  & 0.423  & 0.424  \\
\cmidrule{2-14}          & avg   & \textcolor[rgb]{ 1,  0,  0}{\textbf{0.299 }} & \textcolor[rgb]{ 1,  0,  0}{\textbf{0.352 }} & 0.360  & 0.386  & 0.343  & 0.356  & \textcolor[rgb]{ .161,  .447,  .957}{\underline{0.313 }} & 0.363  & 0.320  & \textcolor[rgb]{ .161,  .447,  .957}{\underline{0.353 }} & 0.319  & 0.360  \\
    \midrule
    \multirow{5}[10]{*}{Weather} & 96    & \textcolor[rgb]{ 1,  0,  0}{\textbf{0.200 }} & 0.260  & 0.190  & \textcolor[rgb]{ 1,  0,  0}{\textbf{0.236 }} & \textcolor[rgb]{ .161,  .447,  .957}{\underline{0.188 }} & 0.250  & 0.209  & \textcolor[rgb]{ .161,  .447,  .957}{\underline{0.244 }} & -     & -     & 0.216  & 0.271  \\
\cmidrule{2-14}          & 192   & \textcolor[rgb]{ .161,  .447,  .957}{\underline{0.239 }} & \textcolor[rgb]{ .161,  .447,  .957}{\underline{0.288 }} & 0.261  & 0.293  & \textcolor[rgb]{ 1,  0,  0}{\textbf{0.237 }} & \textcolor[rgb]{ 1,  0,  0}{\textbf{0.284 }} & 0.254  & \textcolor[rgb]{ .161,  .447,  .957}{\underline{0.288 }} & -     & -     & 0.264  & 0.306  \\
\cmidrule{2-14}          & 336   & \textcolor[rgb]{ 1,  0,  0}{\textbf{0.279 }} & \textcolor[rgb]{ 1,  0,  0}{\textbf{0.315 }} & 0.332  & 0.340  & \textcolor[rgb]{ .161,  .447,  .957}{\underline{0.282 }} & \textcolor[rgb]{ .161,  .447,  .957}{\underline{0.323 }} & 0.301  & 0.332  & -     & -     & 0.313  & 0.336  \\
\cmidrule{2-14}          & 720   & \textcolor[rgb]{ 1,  0,  0}{\textbf{0.340 }} & \textcolor[rgb]{ .161,  .447,  .957}{\underline{0.357 }} & 0.385  & 0.381  & \textcolor[rgb]{ .161,  .447,  .957}{\underline{0.359 }} & \textcolor[rgb]{ 1,  0,  0}{\textbf{0.345 }} & 0.388  & 0.374  & -     & -     & 0.369  & 0.380  \\
\cmidrule{2-14}          & avg   & \textcolor[rgb]{ 1,  0,  0}{\textbf{0.265 }} & \textcolor[rgb]{ .161,  .447,  .957}{\underline{0.305 }} & 0.292  & 0.312  & \textcolor[rgb]{ .161,  .447,  .957}{\underline{0.267 }} & \textcolor[rgb]{ 1,  0,  0}{\textbf{0.300 }} & 0.288  & 0.310  & -     & -     & 0.291  & 0.323  \\
    \midrule
    \multirow{5}[10]{*}{Electricity} & 96    & \textcolor[rgb]{ .161,  .447,  .957}{\underline{0.209 }} & 0.307  & 0.210  & 0.312  & 0.212  & \textcolor[rgb]{ .161,  .447,  .957}{\underline{0.301 }} & \textcolor[rgb]{ 1,  0,  0}{\textbf{0.194 }} & \textcolor[rgb]{ 1,  0,  0}{\textbf{0.266 }} & -     & -     & 0.844  & 0.761  \\
\cmidrule{2-14}          & 192   & \textcolor[rgb]{ .161,  .447,  .957}{\underline{0.219 }} & \textcolor[rgb]{ .161,  .447,  .957}{\underline{0.315 }} & 0.239  & 0.337  & 0.225  & 0.320  & \textcolor[rgb]{ 1,  0,  0}{\textbf{0.218 }} & \textcolor[rgb]{ 1,  0,  0}{\textbf{0.289 }} & -     & -     & 0.850  & 0.762  \\
\cmidrule{2-14}          & 336   & \textcolor[rgb]{ .161,  .447,  .957}{\underline{0.236 }} & \textcolor[rgb]{ .161,  .447,  .957}{\underline{0.330 }} & 0.284  & 0.372  & 0.245  & 0.333  & \textcolor[rgb]{ .161,  .447,  .957}{\underline{0.244 }} & \textcolor[rgb]{ 1,  0,  0}{\textbf{0.321 }} & -     & -     & 0.862  & 0.766  \\
\cmidrule{2-14}          & 720   & \textcolor[rgb]{ 1,  0,  0}{\textbf{0.273 }} & \textcolor[rgb]{ 1,  0,  0}{\textbf{0.328 }} & 0.456  & 0.479  & \textcolor[rgb]{ .161,  .447,  .957}{\underline{0.282 }} & \textcolor[rgb]{ .161,  .447,  .957}{\underline{0.358 }} & 0.324  & 0.371  & -     & -     & 0.888  & 0.774  \\
\cmidrule{2-14}          & avg   & \textcolor[rgb]{ 1,  0,  0}{\textbf{0.234 }} & \textcolor[rgb]{ .161,  .447,  .957}{\underline{0.320 }} & 0.297  & 0.375  & \textcolor[rgb]{ .161,  .447,  .957}{\underline{0.241 }} & 0.328  & 0.245  & \textcolor[rgb]{ 1,  0,  0}{\textbf{0.312 }} & -     & -     & 0.861  & 0.766  \\
    \midrule
    \multirow{5}[9]{*}{Traffic} & 96    & 0.572  & 0.407  & \textcolor[rgb]{ 1,  0,  0}{\textbf{0.526 }} & \textcolor[rgb]{ 1,  0,  0}{\textbf{0.368 }} & -     & -     & \textcolor[rgb]{ .161,  .447,  .957}{\underline{0.562 }} & \textcolor[rgb]{ .161,  .447,  .957}{\underline{0.378 }} & -     & -     & 1.390  & 0.800  \\
\cmidrule{2-14}          & 192   & \textcolor[rgb]{ .161,  .447,  .957}{\underline{0.575 }} & \textcolor[rgb]{ .161,  .447,  .957}{\underline{0.406 }} & \textcolor[rgb]{ 1,  0,  0}{\textbf{0.561 }} & \textcolor[rgb]{ 1,  0,  0}{\textbf{0.385 }} & -     & -     & 0.579  & 0.412  & -     & -     & 1.403  & 0.802  \\
\cmidrule{2-14}          & 336   & \textcolor[rgb]{ 1,  0,  0}{\textbf{0.588 }} & \textcolor[rgb]{ 1,  0,  0}{\textbf{0.411 }} & 0.614  & \textcolor[rgb]{ .161,  .447,  .957}{\underline{0.412 }} & -     & -     & \textcolor[rgb]{ .161,  .447,  .957}{\underline{0.594 }} & 0.420  & -     & -     & 1.415  & 0.804  \\
\cmidrule{2-14}          & 720   & \textcolor[rgb]{ 1,  0,  0}{\textbf{0.618 }} & \textcolor[rgb]{ 1,  0,  0}{\textbf{0.422 }} & 0.749  & \textcolor[rgb]{ .161,  .447,  .957}{\underline{0.464 }} & -     & -     & \textcolor[rgb]{ .161,  .447,  .957}{\underline{0.723 }} & 0.472  & -     & -     & 1.437  & 0.808  \\
\cmidrule{2-14}          & avg   & \textcolor[rgb]{ 1,  0,  0}{\textbf{0.588 }} & \textcolor[rgb]{ .161,  .447,  .957}{\underline{0.412 }} & \underline{0.613 } & \textcolor[rgb]{ 1,  0,  0}{\textbf{0.407 }} & -     & -     & \textcolor[rgb]{ .161,  .447,  .957}{\underline{0.615 }} & 0.421  & -     & -     & 1.411  & 0.804  \\
    
    \bottomrule
    \end{tabular}}%
  \label{tab:zero_shot}%
\end{table*}%

% You can have as much text here as you want. The main body must be at most $8$ pages long.
% For the final version, one more page can be added.
% If you want, you can use an appendix like this one.  

% The $\mathtt{\backslash onecolumn}$ command above can be kept in place if you prefer a one-column appendix, or can be removed if you prefer a two-column appendix.  Apart from this possible change, the style (font size, spacing, margins, page numbering, etc.) should be kept the same as the main body.
%%%%%%%%%%%%%%%%%%%%%%%%%%%%%%%%%%%%%%%%%%%%%%%%%%%%%%%%%%%%%%%%%%%%%%%%%%%%%%%
%%%%%%%%%%%%%%%%%%%%%%%%%%%%%%%%%%%%%%%%%%%%%%%%%%%%%%%%%%%%%%%%%%%%%%%%%%%%%%%

\end{document}